\PassOptionsToPackage{table}{xcolor}
\RequirePackage{rotating}
\documentclass{article}
\usepackage{arxiv}

\usepackage[utf8]{inputenc} 
\usepackage[T1]{fontenc}    
\usepackage{hyperref}       
\usepackage{url}            
\usepackage{booktabs}       
\usepackage{amsfonts}       
\usepackage{nicefrac}       
\usepackage{microtype}      
\usepackage{pdflscape}

\usepackage{bm}
\usepackage{multirow}
\usepackage{rotating}
\usepackage{pifont}
\usepackage{graphicx}
\usepackage{doi}
\usepackage{amsmath}
\usepackage{amssymb}
\usepackage{multirow}
\usepackage{xcolor}
\usepackage{cleveref}

\usepackage[backend=biber, style=acmnumeric,doi=false,url=false,defernumbers=true,backref=true]{biblatex}

\defbibfilter{onlymain}{%
 segment=0
}

\defbibfilter{onlyapx}{%
not segment=0
}

\AtEveryBibitem{\clearfield{editor}}
\AtEveryBibitem{\clearfield{editors}}
\AtEveryBibitem{\clearfield{publisher}}

\DeclareNameAlias{author}{family-given}
\newcommand{\citet}{\textcite}

\makeatletter
\let\blx@rerun@biber\relax
\makeatother

\newcommand{\Nat}{\mathbb{N}}
\newcommand{\Real}{\mathbb{R}}
\newcommand{\R}{\bm{R}}
\newcommand{\Rp}{\bm{R'}}
\newcommand{\Out}{\bm{O}}
\newcommand{\Outp}{\bm{O'}}
\newcommand{\Sm}{\bm{S}}
\newcommand{\Smp}{\bm{S'}}

\newcommand{\E}{\mathbb{E}}
\newcommand{\transp}{\mathsf{T}}
\newcommand{\tr}{\operatorname{tr}}

\newcommand{\bbone}{\text{\usefont{U}{bbold}{m}{n}1}}
\MakeRobust{\bbone}

\newcommand{\ols}[1]{\mskip.5\thinmuskip\overline{\mskip-.5\thinmuskip {#1} \mskip-.5\thinmuskip}\mskip.5\thinmuskip}

\newcommand\myeqref[1]{%
    (\textup{Eq.~\ref{#1}})%
}

\newcommand{\subheader}{\textbf}

\newcommand{\cmark}{\ding{51}}
\newcommand{\xmark}{\ding{55}} %

\definecolor{Gray}{gray}{0.9}

\title{Similarity of Neural Network Models:\\A Survey of Functional and Representational Measures}

\date{} 					

\author{ 
    Max~Klabunde \\
	University of Passau\\
	\texttt{max.klabunde@uni-passau.de} \\
 \And
    Tobias~Schumacher \\
	University of Mannheim, \\
	RWTH Aachen University \\
   \texttt{tobias.schumacher@uni-mannheim.de} \\
 \And
    Markus~Strohmaier \\
      University of Mannheim,\\
      GESIS - Leibniz Institute for the Social Sciences, and\\
      Complexity Science Hub Vienna\\
      \texttt{markus.strohmaier@uni-mannheim.de}
 \And
    Florian~Lemmerich \\
	University of Passau\\
    \texttt{florian.lemmerich@uni-passau.de} \\
}

\renewcommand{\headeright}{}

\renewcommand{\shorttitle}{Similarity of Neural Networks: A Survey of Functional and Representational Measures}

\hypersetup{
pdftitle={Similarity of Neural Networks: A Survey of Functional and Representational Measures},
pdfsubject={cs.LG},
pdfauthor={Max~Klabunde, Tobias~Schumacher, Markus~Strohmaier, Florian~Lemmerich},
pdfkeywords={neural networks, representational similarity, functional similarity, survey},
}
\begin{document}
\maketitle


\begin{abstract}
Measuring similarity of neural networks to understand and improve their behavior has become an issue of great importance and research interest.
In this survey, we provide a comprehensive overview of two complementary perspectives of measuring neural network similarity: (i) representational similarity, which considers how \emph{activations} of intermediate layers differ, and (ii) functional similarity, which considers how models differ in their \emph{outputs}.
In addition to providing detailed descriptions of existing measures, we summarize and discuss results on the properties of and relationships between these measures, and point to open research problems.
We hope our work lays a foundation for more systematic research on the properties and applicability of similarity measures for neural network models.
\end{abstract}

\begin{figure*}[t]
    \centering
    \includegraphics[width=0.75\textwidth]{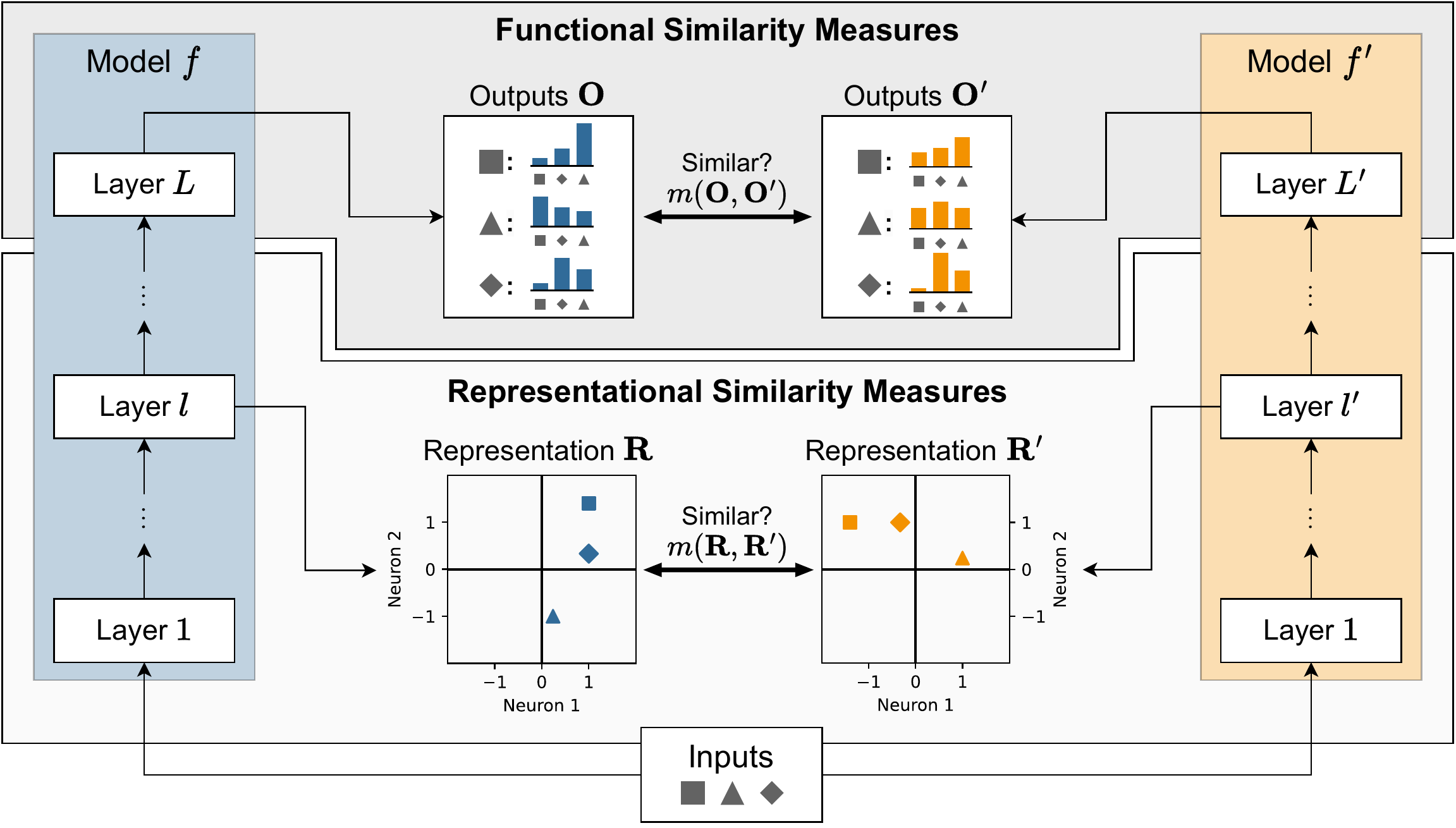}
    \caption{A conceptual overview of representational and functional similarity.
    We compare a pair of neural network models $f, f'$.
    Functional similarity measures mainly consider the outputs $\Out, \Outp$ of the compared models, whereas representational similarity measures consider their intermediate representations $\R, \Rp$.
    All models get the same inputs.
    Specifically in classification tasks, outputs have clear and universal semantics, so that they can be compared in a straightforward manner.
    In contrast, the geometry of the representations requires more care when measuring their similarity. In the illustration above, for instance, rotating $\R$ by 90 degrees would yield an alignment of representations after which they would appear much more similar.
    Combined, representational and functional measures cover all layers of the models.
    }
    \label{fig:fig1}
\end{figure*}

\section{Introduction}\label{sec:introduction}
Measures to quantify similarity of neural network models have been widely applied in the literature, usually to understand and improve deep learning systems. Examples include research on learning dynamics \cite{morcos_insights_2018, mehrer_beware_2018}, effects of width and depth \cite{nguyen_wide_2021}, differences between supervised and unsupervised models \cite{gwilliam_beyond_2022}, robustness \cite{jones_if_2022,nanda_measuring_2022}, effects of data and model updates \cite{milani_fard_launch_2016, liu_model_2022, klabunde_prediction_2023, mehrer_individual_2020}, evaluating knowledge distillation \cite{stanton_does_2022}, designing ensembles \cite{zhang_efficient_2021}, language representation \cite{kudugunta_investigating_2019, hamilton_cultural_2016, hamilton_diachronic_2018}, and generalizability \cite{mccoy_berts_2020, lee_diversify_2023, pagliardini_agree_2023}.

However, understanding and measuring similarity of neural networks is a complex problem, as there are multiple perspectives on how such models can be similar.
In this work, we specifically focus on two key perspectives: \emph{representational} and \emph{functional measures of similarity} (see Figure \ref{fig:fig1}).
Representational similarity measures assess how activations of intermediate layers differ, whereas functional similarity measures compare the outputs of neural networks with respect to their task.
Both perspectives only provide a partial view on neural network similarity.
Seemingly similar representations can still yield different outputs, and conversely, similar outputs can result from different representations.
In that sense, combining these two complementary perspectives provides a more comprehensive approach to analyze similarity between neural networks at all layers.

Given the broad range of research on neural network similarity, numerous representational and functional similarity measures have been proposed and applied, often with lines of research being disconnected from each other.
With this work, we provide a comprehensive overview of these two groups of similarity measures that gives a unified perspective on the existing literature and can inform and guide both researchers and practitioners interested in understanding and comparing neural network models.

Measures for representational or functional similarity have been covered in prior work to some extent.
Regarding representational similarity, measures for matrix correlation have been reviewed by \cite{ramsay_matrix_1984, yang_survey_2021}.
Existing surveys, however, lack coverage of more recent measures or do not consider the context of deep learning.
A recent survey by \citet{rauker_toward_2022} reviews methods to interpret inner workings of neural networks, but discusses representational similarity measures only briefly.
\citet{sucholutsky_getting_2023} complement this survey by discussing representational similarity with a focus on bringing together the communities of machine learning, neuroscience, and cognitive science, which have all been working independently on comparing representations.
Functional similarity measures have been surveyed in the context of ensemble learning \cite{kuncheva_measures_2003, brown_diversity_2005}, inter-rater agreement \cite{banerjee_beyond_1999, gisev_interrater_2013,tinsley_interrater_1975}, model fingerprinting \cite{sun_deep_2023}, and image and text generation scenarios \cite{borji_pros_2018, celikyilmaz_evaluation_2021}, which each focus on application scenarios with objectives different to our survey.
We specifically focus on multi-class classification contexts for functional similarity measures.

To the best of our knowledge, our survey represents the first comprehensive review of representational and functional similarity measures for neural network models.
This survey makes the following contributions:
\begin{enumerate}
    \item \textbf{Systematic and comprehensive overview:}
    We formally define the problem of measuring representational and functional similarity in neural networks---the latter in the context of classification---and provide a systematic and comprehensive overview of existing measures.
    \item \textbf{Unified terminology:} We provide detailed definitions, explanations, and categorizations for each measure in a unified manner, facilitating the understanding of commonalities and differences between measures.
    \item \textbf{Analysis of practical properties and applicability:} We discuss the practical properties of existing measures, such as robustness to noise or confounding issues, and connections between existing measures to guide researchers and practitioners in applying these measures.
    \item \textbf{Open research challenges:} We highlight unresolved issues of similarity measures and point out research gaps that can be addressed in the future to improve our understanding of neural networks in general.
\end{enumerate}

While we focus on measures for representational and functional similarity due to their prevalence and general applicability, we acknowledge various other approaches to comparing neural networks.
In particular, the measures covered in our survey differ from methods typically used to assess and optimize similarity during model training. We discuss these and other approaches in \Cref{sec:alt_similarity}.

\section{Similarity of Neural Network Models}\label{sec:modelsim}
We consider the problem of comparing  neural networks, which we assume to have the form
\begin{equation}
f = f^{(L)} \circ f^{(L-1)} \circ \dots \circ f^{(1)},
\end{equation}
with each function $f^{(l)}:\Real^{D^{(l-1)}} \longrightarrow \Real^{D^{(l)}}$ denoting a single layer of $D:=D^{(l)}$ neurons, and a total number of $L\in\Nat$ layers.
These networks operate on a set of $N$ given inputs $\{\bm{X}_i\}_{i=1}^N$,
which we typically assume to be vectors in $\Real^p$, $p\in\mathbb{N}$, although these can also be higher-dimensional structures as occurring in image or video data.
We collect these inputs in a matrix $\bm{X}\in\Real^{N\times p}$ so that the $i$-th row $\bm{X}_i$ corresponds to the $i$-th input.
To further simplify notation, we also denote individual inputs $\bm{X}_i$ as \emph{instances} $i\in\{1,\dots,N\}$.
We generally do not make any assumption about the number of features $p$, the depth of the network $L$, the width or activation function of any layer $f^{(l)}$, or the training objective.

Similarity of neural network models is then quantified by \emph{similarity measures} $m$.
For simplicity, we also consider measures that quantify \emph{distance} between models as similarity measures, since these concepts are generally equivalent.
In our survey, we specifically consider two kinds of similarity, namely \emph{representational similarity} and \emph{functional similarity}.
Representational similarity measures consider how the inner activations of neural network models differ, whereas functional similarity measures compare the output behavior of neural networks with respect to a given (classification) task.
Combined, these two notions allow for nuanced insights into similarity of neural network models \cite{summers_nondeterminism_2021,klabunde_prediction_2023,gwilliam_beyond_2022}.

In the following, we give more thorough definitions of representational and functional similarity.
For the rest of this paper, we introduce notations for commonly used variables only once.
In \Cref{ap:notations}, we provide an overview of notation and several definitions of variables and functions that are used in this paper.

\subsection{Representational Similarity}
Representational similarity measures compare neural networks by measuring similarity between activations of a fixed set of inputs at any pair of layers. 
Given such inputs $\bm{X}$, we define the representation of model $f$ at layer $l$ as a matrix
\begin{equation}
    \R := \R^{(l)} = \left(f^{(l)} \circ f^{(l-1)} \circ \dots \circ f^{(1)}\right)(\bm{X})\in\Real^{N\times D}.
\end{equation}
The activations of instance $i$ then correspond to the $i$-th row $\R_i = \left( f^{(l)} \circ \dots \circ f^{(1)}\right) \left(\bm{X}_i\right)\in\Real^D$, which we denote as \emph{instance representation}.
The activations of single neurons over all instances correspond to the columns of $\R$, and we denote the $j$-th column of $\R$ as $\R_{\--,j}$.
Like the inputs, we also consider the instance representations $\R_i$ to be vectors even though in practice, e.g., in convolutional neural networks, these activations can also be matrices. In such a case, these representations can be flattened (see \Cref{ap:preprocessing}).

Representational similarity measures are typically defined as mappings $m: \Real^{N\times D}\times \Real^{N\times D'} \longrightarrow \Real$ that assign a similarity score $m(\R,\Rp)$ to a pair of representations $\R,\Rp$, which are derived from different models $f, f'$, but use the same inputs $\bm{X}$.
While we assume here that representations stem from different models, representational similarity measures can also be used to compare representations of different layers of the same model.
Without loss of generality, we assume that $D\leq D'$, though some measures require that $D=D'$. In such cases, preprocessing techniques can be applied (see \Cref{ap:preprocessing}).
We note that this definition is limited to comparisons of \emph{pairs} of representations, which is the standard setting in literature. In practice, one may also be interested in measuring similarity of \emph{groups} of representations.
The most direct way to obtain such measures of similarity for groups of representations from the standard pairwise measures is to aggregate the pairwise similarity scores, e.g., by averaging the similarities of all pairs of representations.

Typical issues when measuring similarity of representations are that the measures have to identify when a pair of representations is equivalent, and that some measures may require preprocessing of the representations.
In the following sections, we discuss these issues and related concepts in more detail.

\subheader{Equivalence of Representations.}
Even if two representation matrices $\R, \Rp\in\Real^{N\times D}$ are not identical on an element-per-element basis, one may still consider them to be equivalent, i.e., perfectly similar.
An intuitive example for such a case would be when representations only differ in their sign, i.e., $\R= -\Rp$, or when representations can be rotated onto another.
Such notions of equivalence can be formalized in terms of bijective mappings (transformations) $\varphi:\Real^{N \times D}\longrightarrow \Real^{N \times D}$ that yield $\varphi(\R) = \Rp$.
What kind of transformations constitute equivalence between representations may vary depending on the context at hand.
For instance, equivalence up to rotation does not make sense if some feature dimensions are already aligned with fixed axes, as is the case in interpretable word embeddings where axes may represent scales between polar opposites like ``bright" and ``dark" \cite{mathew_polar_2020}.
Thus, we define equivalence of representations in terms of groups of transformations $\mathcal{T}:=\mathcal{T}(N,D)$, and call two representations $\R, \Rp$ equivalent with respect to a group $\mathcal{T}$, written as $\R \sim_\mathcal{T} \Rp$, if there is a $\varphi\in\mathcal{T}$ such that $\varphi(\R) = \Rp$.

In practice, it is crucial to determine under which groups of transformations representations should be considered equivalent, as equivalent representations should be indistinguishable for the chosen similarity measure.
Conversely, representations that are not equivalent have to be distinguishable for a similarity measure.
In formal terms, this means that a measure has to be \emph{invariant} to exactly those groups of transformations that the underlying representations are equivalent under.
We call a representational similarity measure $m$ invariant to a group of transformations $\mathcal{T}$, if for all $\R\in\Real^{N\times D}$, $\Rp\in\Real^{N\times D'}$ and all $\varphi \in \mathcal{T}(N,D)$, $\varphi' \in \mathcal{T}(N,D')$ it holds that $m(\R,\Rp) = m(\varphi(\R),\varphi'(\Rp))$.
Thus, if a measure $m$ is invariant to $\mathcal{T}$, it directly follows that $m(\R,\R) = m(\R, \Rp)$ if $\R\sim_\mathcal{T}\Rp$.
This implies that a measure can only distinguish representations that are not equivalent under the groups of transformations it is invariant to.
Using this notion of invariance, and assuming that representations have identical dimensionality, we can also analyze whether measures satisfy the criteria of a distance metric
---in the context of representational similarity, these criteria are typically relaxed to only require $m(\R,\Rp)=0$ if and only if $\R\sim_{\mathcal{T}} \Rp$ for a group of transformations $\mathcal{T}$ that $m$ is invariant to \cite[Apx. A.2]{williams_generalized_2022}. 

In the literature \cite{kornblith_similarity_2019,williams_generalized_2022,raghu_svcca_2017,li_convergent_2016}, there are six main groups of transformations under which representations are considered equivalent, and that representational similarity measures are often designed to be invariant to:

\begin{itemize}
    \item \textbf{Permutations (PT).}
    A similarity measure $m$ is invariant to permutations if swapping columns of the representation matrices $\R$, that is, reordering neurons, does not affect the resulting similarity score. Letting $S_D$ denote the set of all permutations on $\{1,\dots,D\}$, and for $\pi\in S_D$, $\bm{P}_\pi = (p_{i,j})\in\Real^{D\times D}$ denote the permutation matrix where $p_{i,j}=1$ if $\pi(i)=j$ and $p_{i,j}=0$ otherwise, the group of all permutation transformations is given by
    \begin{equation}
     \mathcal{T}_\text{PT} = \{\R\mapsto \R \bm{P}_\pi: \pi\in S_D\}.
    \end{equation}
    Permutations neither affect Euclidean distances nor angles between instance representations.

    \item \textbf{Orthogonal Transformations (OT).}
    As noted in an earlier example, one might intuitively consider two representations equivalent if they can be rotated onto each other.
    Next to rotations, the group of orthogonal transformations also includes permutations and reflections.
    Letting $\operatorname{O}(D) := \{\bm{Q} \in\Real^{D\times D}, \bm{Q}^{\transp} \bm{Q} = \bm{I}_D\}$ denote the orthogonal group, the set of these transformations is given by
    \begin{equation}
    \mathcal{T}_\text{OT} = \{\R\mapsto \R \bm{Q}: \bm{Q} \in\operatorname{O}(D)\}.
    \end{equation}
    These transformations preserve both Euclidean distances and angles between instance representations.

    \item \textbf{Isotropic Scaling (IS).} Scaling all elements of a representation $\R$ identically (isotropic scaling) does not change the angles between instance representations $\R_i$.
    The set of all isotropic scaling transformations is defined as
    \begin{equation}
     \mathcal{T}_\text{IS} = \{\R\mapsto a\cdot \R: a\in\Real_+\}.
    \end{equation}
    Isotropic scaling of representations will also rescale the Euclidean distance between instance representations by the same scaling factor $a$.

    \item \textbf{Invertible Linear Transformations (ILT).}
    The group of invertible linear transformations, which is defined as
    \begin{equation}
     \mathcal{T}_\text{ILT} = \{\R\mapsto \R\bm{A}: \bm{A}\in\operatorname{GL}(D,\Real)\},
    \end{equation}
    with $\operatorname{GL}(D,\Real)$ denoting the general linear group of all invertible matrices $\bm{A}\in\Real^{D\times D}$, forms a broader group of transformations.
    It includes both orthogonal transformations and rescalings.
    Both angles and Euclidean distances between instance representations are generally not preserved.

    \item \textbf{Translations (TR).} If the angles between instance representations $\R_i$ are not of concern, one might argue that two representations are equivalent if they can be mapped onto each other by adding a constant vector.
    In that regard, a measure $m$ is invariant to translations if is invariant to the set of all mappings
    \begin{equation}
     \mathcal{T}_\text{TR} = \{\R\mapsto \R + \bm{1}_N\bm{b}^{\transp}: \bm{b}\in\Real^D\},
    \end{equation}
    where $\bm{1}_N$ is a vector of $N$ ones. 
    Translations preserve Euclidean distances between instance representations.

    \item \textbf{Affine Transformations (AT).}
    The most general group of transformations that is typically considered for representations is given by the set of affine transformations
    \begin{equation}
     \mathcal{T}_\text{AT} = \{\R\mapsto \R\bm{A} + \bm{1}_N\bm{b}^\transp: \bm{A}\in\operatorname{GL}(D,\Real),\, \bm{b}\in\Real^D\}.
    \end{equation}
    This group of transformations in particular also includes rescaling, translations, orthogonal transformations, and invertible linear transformations.
    Therefore, affine transformations in general do neither preserve angles nor Euclidean distances between instance representations.
\end{itemize}

\begin{figure}[t]
    \centering
    \includegraphics[width=0.9\linewidth]{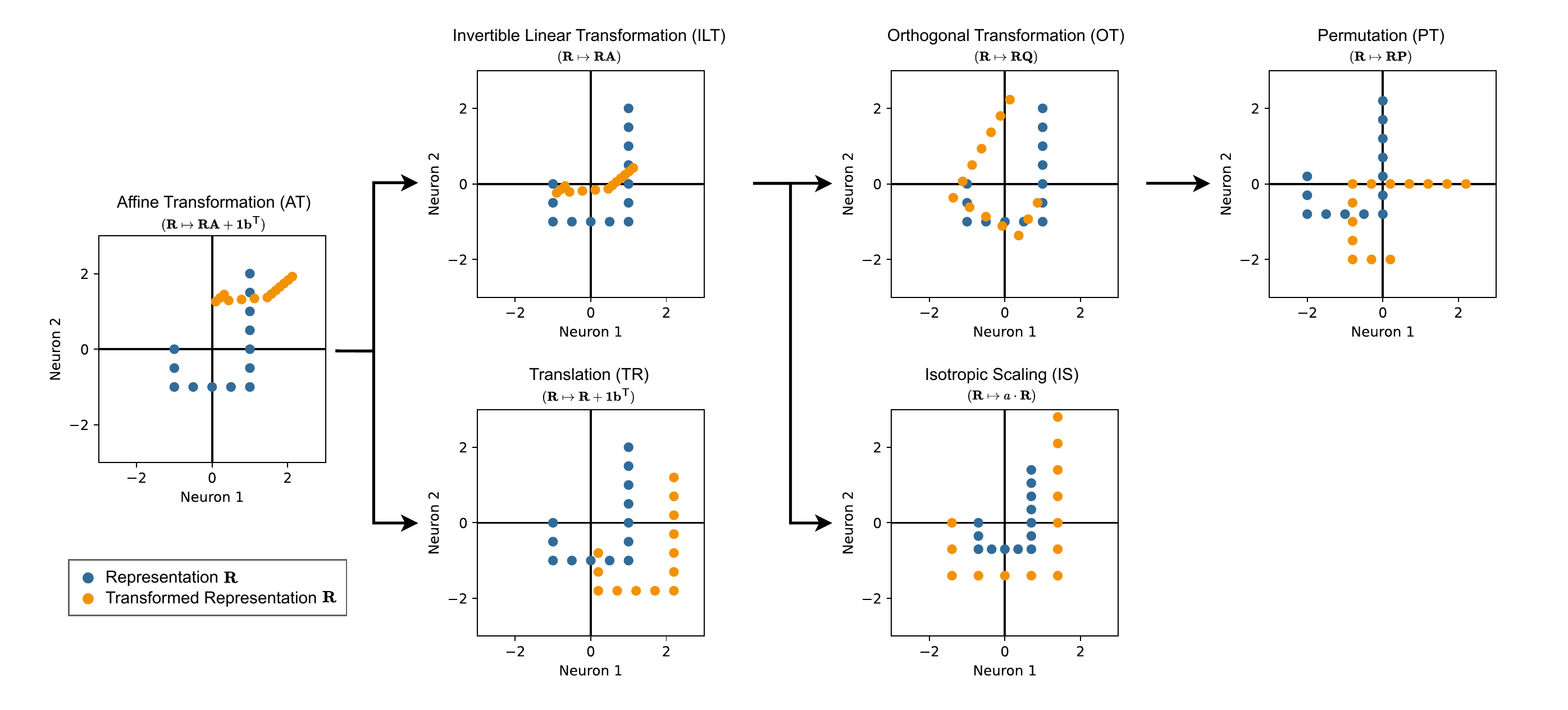}
    \caption{Illustration of representations considered equivalent under different invariances.
    The invariances form a hierarchy:
    Arrows describe implication, with the left invariance being more general.
    For AT and ILT, the same linear transformation is applied.
    AT further translates representations by the same vector that is used in TR.
    In OT, the representations are rotated (120°) and reflected over the 15° axis.
    In PT, axes are swapped.
    IS applies a scaling factor of 2.
    See \Cref{ap:fig2_transforms} for exact parameter values.
    }
    \label{fig:invariances}
\end{figure}
\noindent We depict the hierarchy of these groups in Figure \ref{fig:invariances}.
\Cref{tab:categorization} also shows the invariances of all representational similarity measures covered in this survey with respect to these groups.
This list of groups of transformations is, however, not exhaustive, and both for practical and theoretical reasons, various other groups may be considered \cite{godfrey_symmetries_2022}.
In practice, neurons are typically assumed to be indexed arbitrarily, so most representational similarity measures are invariant to permutations~\cite{kornblith_similarity_2019,khosla_soft_2024,li_convergent_2016}.
\citet{raghu_svcca_2017} further argued for invariance to invertible linear transformations, as any such transformation could be reverted by a directly following linear layer without altering overall network behavior.
However, \citet{kornblith_similarity_2019} criticized that such invariance leads to unintuitive similarity behavior when $D>N$, such as all representations with full rank being equivalent, and that training of neural networks is not invariant to linear transformations.
They argued that orthogonal transformations capture practical differences in representations better.

\subheader{Preprocessing of Representations.}\label{sec:reppreproc}
Many representational similarity measures assume certain properties of the representations $\R, \Rp$ that, in practice, are not always given.
For instance, it is often assumed that representations are mean-centered in the columns \cite{kornblith_similarity_2019,williams_generalized_2022,morcos_insights_2018}, or that they have the same dimensionality.
In these cases, the representations need to be preprocessed.
There are three kinds of preprocessing that may have to be applied, namely normalization, adjusting dimensionality, and flattening of representations.
We discuss these problems in Appendix \ref{ap:preprocessing}.

\subsection{Functional Similarity Measures}\label{subsec:funcsim_intro}
Functional similarity measures compare neural networks by measuring similarity of their output behavior \cite{csiszarik_similarity_2021}.
Given a set of inputs  $\bm{X}$ and a neural network $f$ that is trained for a classification task on $C$ classes, we let
\begin{equation}
  \bm{O} := f(\bm{X}) \in \mathbb{R}^{N \times C}
\end{equation}
denote the matrix of its outputs.
Each row $\Out_i=f(\bm{X}_i)\in\Real^C$ corresponds to the output for input $\bm{X}_i$.
In the context of this survey, we assume that this vector-based output corresponds to \emph{soft predictions}, where each element  $\Out_{i,c}$ denotes the probabilities or decision scores of class $c$ for input $\bm{X}_i$.
From these soft predictions, we can compute the \emph{hard predictions} for a given multiclass classification task via $\hat{c} = \arg\max_c \Out_{i,c}$, where $\hat{c}$ denotes the predicted class for input $\bm{X}_i$.

Then, similar to representational similarity measures, functional similarity measures are defined as mappings $m: \Real^{N\times C}\times \Real^{N\times C} \longrightarrow \Real$ that assign a similarity score $m(\Out,\Outp)$ to a pair of outputs $\Out,\Outp$, which are derived from the same inputs $\bm{X}$.
For the compared outputs $\Out, \Outp \in \Real^{N\times C}$, it is assumed that they are aligned in the sense that the columns $\Out_{\--, c}, \Outp_{\--, c}$ correspond to probability/decision scores of the same class $c$.

Due to this alignment and the fixed semantics of outputs, analyzing functional similarity generally does not require consideration of preprocessing or invariances.
For the same reason, representational similarity measures are unsuitable for comparison of outputs---the previous assumptions do not hold for representations.
Thus, for instance, all permutation invariant measures would consider two outputs that assign 100\% probability to different classes equivalent.

Moreover, many functional similarity measures require only black-box access to a model, relying solely on knowledge about inputs and outputs.
However, functional similarity measures may include additional information aside from the raw outputs $\Out_i$.
For instance, a set of ground-truth labels $\bm{y}\in\Real^N$ is often given, which is typically used by a quality function $q$ that quantifies how well the output matches the ground-truth.
Another kind of additional information are task-based gradients, which, however, require white-box access to the model.
Finally, in the context of functional similarity, it is more common that measures compare multiple models at once without relying on pairwise comparisons.

\subsection{Relationship Between Representational and Functional Similarity}\label{subsec:relation_func_rep}
The notions of representational and functional similarity complement each other (see Fig. \ref{fig:fig1}), and applying both representational and functional similarity measures allows for a more holistic view of neural network similarity \cite[e.g.,][]{summers_nondeterminism_2021,klabunde_prediction_2023,gwilliam_beyond_2022}.
To properly interpret potentially conflicting similarity scores stemming from these two perspectives, it is crucial to understand their relationship.

When functional similarity measures indicate dissimilarity, representations must be dissimilar at some layer, assuming that differences in the final classification layer cannot fully explain the functional difference.
The opposite is not true: two functionally similar models may use dissimilar representations.
Even more, if a functional similarity measure indicates high similarity on a given input set, this does not imply that the compared models are functionally similar in general: high similarity may be the due to easy-to-classify inputs, and out-of-distribution inputs, which tend to amplify functional differences, could yield lower similarity in the corresponding outputs.
Similarly, a representational measure indicating high similarity might not generally indicate high functional or representational similarity between models either, as the invariance of a measure might not fit to the given representations.

In conclusion, one generally cannot expect functional and representational measures to correlate, and their scores require contextualization. Only if there is significant functional dissimilarity between two models, there also should be a representational measure indicating significant dissimilarity.
Since functional outputs and their similarity measures have a clear and intuitive semantic, this relation can also be used to validate representational similarity measures \cite{ding_grounding_2021}.

\section{Representational Similarity Measures}
\label{sec:repsim_methods}

\begin{figure}
    \centering
    \includegraphics[width=\textwidth]{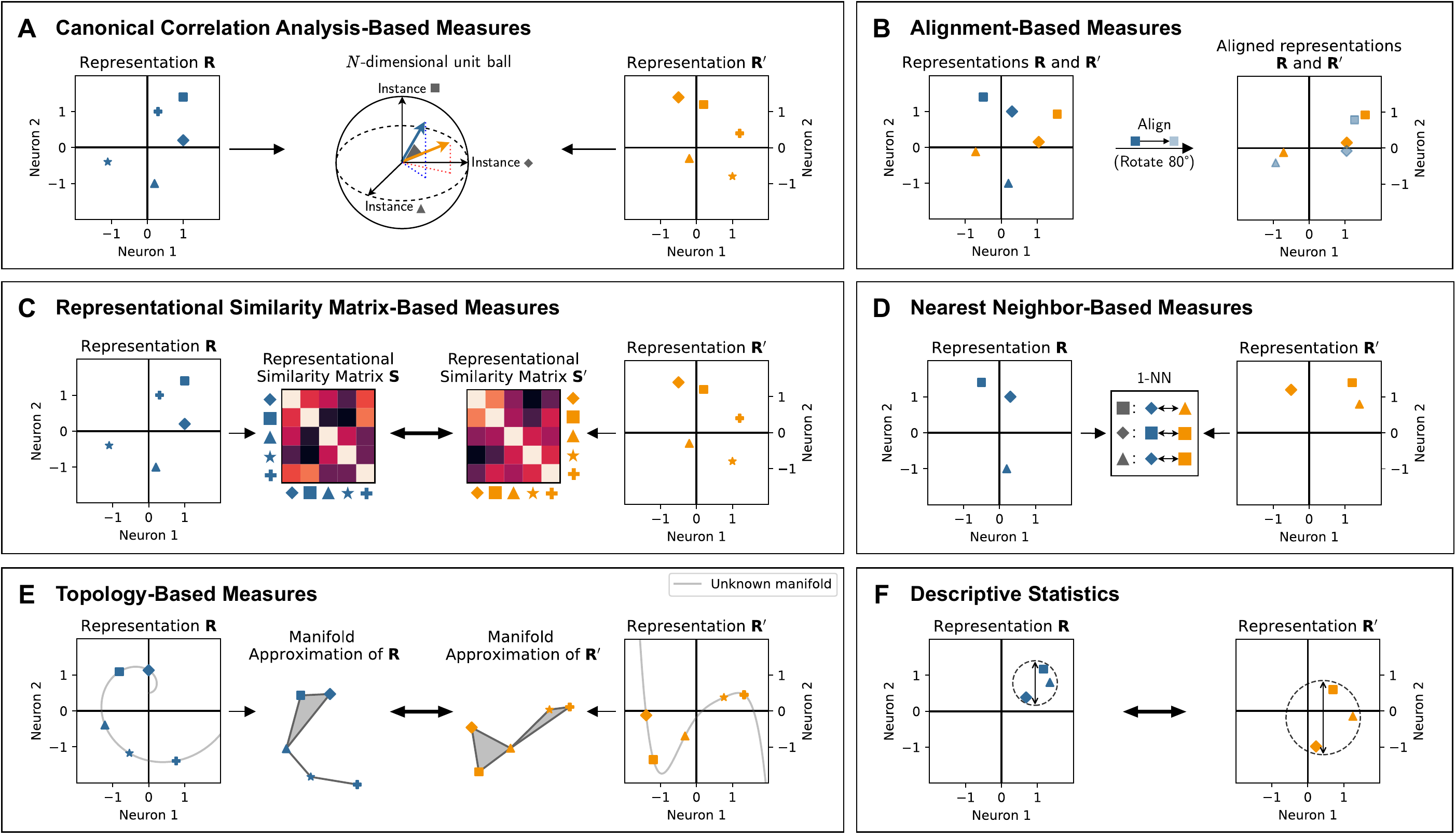}
    \caption{Types of representational similarity measures, illustrated with 2-dimensional representations.
    \textbf{A:} Representations of $N$ instances are projected onto the $N$-dimensional unit ball, and similarity is then quantified based on their angle (their correlation). The illustration of the unit ball is not to scale, and only the first three dimensions are shown.
    \textbf{B:} Representations are aligned with each other, and similarity is computed after alignment.
    \textbf{C:} Similarity is based on comparing matrices of pairwise similarities within representations.
    \textbf{D:} Representations are compared based on similarity of their $k$ nearest neighbors, here $k=1$.
    \textbf{E:} Manifolds of the representations are approximated and compared.
    \textbf{F:} Statistics are computed individually for each representation (here: spread of instance representations) and then compared.}
    \label{fig:repsim_illustr}
\end{figure}

We now review existing representational similarity measures, categorized by their underlying approach to measuring similarity. The categories are illustrated in \Cref{fig:repsim_illustr}.
An overview of all reviewed representational similarity measures can be found in Table \ref{tab:categorization}.

\begin{table*}[ht]

    \resizebox{\linewidth}{!}{%
    \rowcolors{2}{white}{gray!25}
	\begin{tabular}{llcccccccccc} \toprule
									                                                        &                                  		                                                                    & \multicolumn{6}{c}{\cellcolor{white}Invariances} 																&             		&        	    &           &                           \\
		\cline{3-8}
	Type                                                                                	&          Measure                                                                                          & PT 			& OT 	  			&\cellcolor{white} IS   & ILT 			& TR 				& AT 	        & Preprocessing		& $D \neq D'$ 	& Metric	& Similarity $\uparrow$ 	\\
		\midrule
\cellcolor{white}		       	                                                            & Mean Canonical Correlation \cite{raghu_svcca_2017}                                                        & \cmark 		& \cmark 				& \cmark 			& \cmark 		& \cmark 			& \cmark 	    & CC	        	& \cmark      	& \xmark 	& \cmark        			\\
\cellcolor{white}					                                                        & Mean Squared Canonical Correlation \cite{kornblith_similarity_2019,yanai_unification_1974}     & \cmark 		& \cmark 				& \cmark 			& \cmark 		& \cmark 			& \cmark 	    & CC	        	& \cmark      	& \xmark 	& \cmark         			\\
\cellcolor{white}					                                                        & Singular Vector Canonical Correlation Analysis (SVCCA) \cite{raghu_svcca_2017}                            & \cmark 		& \cmark 				& \cmark 			& \xmark 		& \cmark			& \xmark 	    & CC	        	& \cmark      	& \xmark 	& \cmark         			\\
\cellcolor{white} \multirow{-4}{*}{\shortstack{Canonical\\ Correlation\\ Analysis}}         & Projection-Weighted Canonical Correlation Analysis (PWCCA) \cite{morcos_insights_2018}                    & \xmark 		& \xmark 				& \cmark 			& \xmark 		& \cmark			& \xmark 	    & CC	        	& \cmark      	& \xmark 	& \cmark         			\\
		\midrule
\cellcolor{white}				                                                            & Orthogonal Procrustes \cite{ding_grounding_2021,williams_generalized_2022}                                & \cmark 		& \cmark$^*$			& \xmark 			& \xmark 		& \xmark			& \xmark 	    & \xmark	        & \xmark      	& \cmark  	& \xmark        			\\
\cellcolor{white}				                                                            & Angular Shape Metric \cite{williams_generalized_2022} 			                                        & \cmark 		& \cmark$^*$			& \cmark 			& \xmark 		& \xmark			& \xmark 	    & MN		        & \xmark      	& \cmark  	& \xmark        			\\
\cellcolor{white}									                                        & Partial Whitening Shape Metric \cite{williams_generalized_2022}                                           & \cmark 	    & \cmark        		& \cmark$^\dagger$	& \cmark$^\dagger$& \cmark          & \cmark$^\dagger$ & \xmark 	    & \xmark   		& \cmark     & \xmark       			\\
\cellcolor{white}									                                        & Soft Matching Distance \cite{khosla_soft_2024}                                                      & \cmark 	    & \xmark        		& \cmark         	& \xmark        & \cmark            & \xmark & CC, MN 	    & \cmark   		& \cmark     & \xmark       			\\
\cellcolor{white}                                                                           & Linear Regression \cite{li_convergent_2016,kornblith_similarity_2019}                                     & \cmark 		& \cmark 				& \cmark 			& \xmark 		& \cmark			& \xmark 	    & CC	        & \cmark       	& \xmark     & \xmark       			\\
\cellcolor{white}									                                        & Aligned Cosine Similarity \cite{hamilton_diachronic_2018}                                                 & \cmark 		& \cmark 				& \cmark 			& \xmark 		& \xmark			& \xmark 	    & \xmark	        & \xmark       	& \xmark     & \cmark       			\\
\cellcolor{white}									                                        & Correlation Match \cite{li_convergent_2016}                                                               & \cmark		& \xmark 				& \cmark 			& \xmark 		& \cmark 			& \xmark 	    & CC		        & \xmark       	& \xmark     & \cmark       			\\
\cellcolor{white} 									            & Maximum Matching Similarity \cite{wang_towards_2018}                                                                    & \cmark 		& \cmark 				& \cmark 			& \cmark 		& \xmark			& \xmark 	    & \xmark	        & \xmark       	& \xmark     & \cmark       			\\
\cellcolor{white} \multirow{-9}{*}{Alignment} 									            & ContraSim \cite{rahamim_contrasim_2024}                                                                    & \cmark$^\dagger$ 		& \cmark$^\dagger$ 				& \cmark$^\dagger$ 			& \cmark$^\dagger$ 		& \cmark$^\dagger$			& \cmark$^\dagger$ 	    & \xmark	        & \cmark       	& \xmark     & \cmark       			\\
		\midrule
\cellcolor{white}		 		                                                            & Norm of Representational Similarity Matrix Difference \cite{shahbazi_using_2021,yin_dimensionality_2018}               & \cmark$^\ddagger$ & \cmark$^\ddagger$ & \xmark$^\ddagger$ & \xmark 		& \xmark$^\ddagger$	& \xmark 	    & \xmark	        & \cmark      	& \cmark$^\ddagger$     & \xmark  					\\
\cellcolor{white}									                                        & Representational Similarity Analysis (RSA) \cite{kriegeskorte_representational_2008}                      & \cmark$^\ddagger$ & \xmark$^\ddagger$ & \cmark$^\ddagger$ & \xmark 		& \cmark$^\ddagger$ & \xmark 	    & \xmark	        & \cmark      	& \xmark     & \cmark$^\ddagger$       	\\
\cellcolor{white}									                                        & Centered Kernel Alignment (CKA) \cite{kornblith_similarity_2019}                                          & \cmark 		& \cmark 				& \cmark 			& \xmark 		& \cmark			& \xmark 	    & CC		        & \cmark      	& \xmark     & \cmark         			\\
\cellcolor{white}									                                        & Distance Correlation (dCor) \cite{szekely_measuring_2007}                                                 & \cmark 		& \cmark 				& \xmark 			& \xmark 		& \cmark 			& \xmark 	    & \xmark	        & \cmark      	& \xmark     & \cmark        			\\
\cellcolor{white}                                                                           & Normalized Bures Similarity (NBS) \cite{tang_similarity_2020}                                             & \cmark 		& \cmark 				& \cmark 			& \xmark 		& \xmark			& \xmark 	    & \xmark	        & \cmark      	& \xmark     & \cmark        			\\
\cellcolor{white} 																			& Eigenspace Overlap Score (EOS) \cite{may_downstream_2020}                                                 & \cmark 		& \cmark 				& \cmark 			& \cmark        & \xmark			& \xmark 	    & \xmark            & \cmark      	& \xmark     & \cmark        			\\
\cellcolor{white}                                                                           & Unified Linear Probing (GULP) \cite{boix-adsera_gulp_2022}                                                & \cmark 		& \cmark 				& \cmark 			& \xmark$^\dagger$ & \cmark			& \xmark$^\dagger$ & CC, RN 	    & \cmark      	& \cmark     & \xmark        			\\
\cellcolor{white} & Riemmanian Distance \cite{shahbazi_using_2021}                                                            & \cmark 		& \cmark 				& \xmark 			& \xmark 		& \xmark			& \xmark 	    & \xmark	        & \cmark      	& \cmark     & \xmark        			\\
\cellcolor{white} \multirow{-9}{*}{\shortstack{Represen-\\tational\\ Similarity\\ Matrix}}	& Relational Knowledge Loss \cite{park_relational_2019}                                                            & \cmark 		& \cmark				& \cmark$^\dagger$ 			& \xmark 		& \cmark		& \xmark	    & \xmark	        & \cmark      	& \xmark    &  \xmark        			\\
\midrule
\cellcolor{white}			                                                                & $k$-NN Jaccard Similarity \cite{schumacher_effects_2021, wang_towards_2020,hryniowski_inter-layer_2020,gwilliam_beyond_2022}& \cmark 		& \cmark 				& \cmark 			& \xmark 		& \xmark$^\ddagger$	& \xmark 	    & \xmark	        & \cmark       	& \xmark     & \cmark         			\\
\cellcolor{white}									                                        & Second-Order Cosine Similarity \cite{hamilton_cultural_2016}                                              & \cmark 		& \cmark 				& \cmark 			& \xmark 		& \xmark			& \xmark 	    & \xmark	        & \cmark       	& \xmark     & \cmark         			\\
\cellcolor{white}									                                        & Rank Similarity \cite{wang_towards_2020}                                                                  & \cmark 		& \cmark 				& \cmark 			& \xmark 		& \xmark$^\ddagger$	& \xmark 	    & \xmark	        & \cmark       	& \xmark     & \cmark         			\\
\cellcolor{white} \multirow{-4}{*}{Neighbors}								                & Joint Rank and Jaccard Similarity \cite{wang_towards_2020}                                                & \cmark 		& \cmark 				& \cmark 			& \xmark 		& \xmark$^\ddagger$	& \xmark 	    & \xmark	        & \cmark       	& \xmark     & \cmark         			\\
\midrule
\cellcolor{white}     																		& Geometry Score (GS) \cite{khrulkov_geometry_2018}                                                         & \cmark 		& \cmark 				& \cmark 			& \xmark 		& \cmark			& \xmark 	    & \xmark	        & \cmark      	& \xmark     & \xmark        		\\
\cellcolor{white} 							                                                & Multi-Scale Intrinsic Distance (IMD) \cite{tsitsulin_shape_2020}                                          & \cmark 		& \cmark 				& \cmark 			& \xmark 		& \cmark			& \xmark 	    & \xmark	        & \cmark      	& \xmark     & \xmark           		\\
\cellcolor{white} \multirow{-3}{*}{Topology}                                                & Representation Topology Divergence (RTD) \cite{barannikov_representation_2022}                            & \cmark 		& \cmark 				& \cmark 			& \xmark 		& \cmark			& \xmark 	    & \xmark	        & \cmark      	& \xmark     & \xmark        			\\
\midrule
\cellcolor{white} 							                                                & Intrinsic Dimension \cite{camastra_intrinsic_2016}                                                        & \cmark 		& \cmark 				& \cmark 			& \cmark 		& \cmark			& \cmark 	    & \xmark	        & \cmark      	& \xmark     & $^\mathsection$        		\\
\cellcolor{white}			                                                                & Magnitude \cite{wang_towards_2020}                                                                        & \cmark 		& \cmark 				& \xmark 			& \xmark 		& \xmark			& \xmark 	    & \xmark	        & \cmark      	& \xmark     & $^\mathsection$         		\\
\cellcolor{white}									                                        & Concentricity \cite{wang_towards_2020}                                                                    & \cmark 		& \cmark 				& \cmark 			& \xmark 		& \xmark			& \xmark 	    & \xmark	        & \cmark      	& \xmark     & $^\mathsection$         		\\
\cellcolor{white}									                                        & Uniformity \cite{wang_understanding_2020}                                                                & \cmark 		& \cmark 				& \xmark 			& \xmark 		& \cmark			& \xmark 	    & \xmark	        & \cmark      	& \xmark     & $^\mathsection$         		\\
\cellcolor{white} 							         	        							& Tolerance \cite{wang_understanding_2021}                                                                  & \cmark 		& \cmark 				& \cmark 			& \xmark 		& \xmark			& \xmark 	    & RN	        	& \cmark      	& \xmark     & $^\mathsection$         		\\
\cellcolor{white}									                                        & Instance-Graph Modularity \cite{saini_subspace_2021,lu_understanding_2022}                                                         & \cmark 		& \cmark 				& \cmark$^\dagger$ 			& \xmark 		& \xmark			& \xmark 	    & \xmark	        & \cmark      	& \xmark     & $^\mathsection$        		\\
\cellcolor{white} \multirow{-7}{*}{Statistic}							                    & Neuron-Graph Modularity \cite{lange_clustering_2022}                                                      & \cmark 		& \xmark 				& \xmark 			& \xmark 		& \xmark			& \xmark 	    & \xmark	        & \cmark      	& \xmark     & $^\mathsection$        		\\
        \bottomrule
	\end{tabular}
    }
	{\tiny \raggedright $^*$: Subgroups possible.\quad $^\dagger$: Varies based on hyperparameters.\quad $^\ddagger$: Similarity function dependent.\quad $^\mathsection$: Depends on comparison.\\
		\par}

	\caption{
		Overview of representational similarity measures.
        The invariances are permutation (PT), orthogonal transformation (OT), isotropic scaling (IS), invertible linear transformation (ILT), translation (TR), and affine transformation (AT).
        We report invariances based on default hyperparameters and preprocessing as proposed by the authors.
        These may vary if different parameters or similarity functions are applied.
        Three kinds of preprocessing are commonly used: centering columns (CC), normalizing the matrix norm (MN), or normalizing row norms (RN).
		The column $D \neq D'$ indicates whether a measure requires the compared representations to have identical dimensionality.
		\emph{Metric} indicates whether a similarity measure satisfies the criteria of a distance metric when representations have equal dimensionality. 
        \emph{Similarity} $\uparrow$ indicates whether increasing scores imply increasing similarity of models.
	}
	\label{tab:categorization}
\end{table*}

\subsection{Canonical Correlation Analysis-Based Measures}\label{sec:cca}
\emph{Canonical Correlation Analysis} (CCA) \cite{hotelling_relations_1936} is a classical method to compare two sets of values of random variables.
CCA finds weights $\bm{w_{\R}} \in \mathbb{R}^{D}, \bm{w_{\Rp}} \in \mathbb{R}^{D'}$ for the columns in the representations, such that the linear combinations $\R \bm{w_{\R}}$ and $\Rp \bm{w_{\Rp}} \in \mathbb{R}^{N}$ have maximal correlation.
Geometrically, the vectors $\bm{w_{\R}}, \bm{w_{\Rp}}$ are projected to the unit ball in $\Real^N$ via their representation matrices, such that their angle is minimal.
Assuming mean-centered representations, the first \emph{canonical correlation} $\rho$ is defined as
\begin{equation}\label{eq:cancor1}
    \rho := \rho(\R, \Rp) := \max_{\bm{w_{\R}},\bm{w_{\Rp}}} \tfrac{\langle \R \bm{w_{\R}}, \Rp \bm{w_{\Rp}} \rangle}{\|\R\bm{w_{\R}}\| \cdot \|\Rp\bm{w_{\Rp}}\|}.
\end{equation}

One can find additional canonical correlations $\rho_i$, that are uncorrelated and thus orthogonally projected to the previous ones.
This yields a system of $D$ canonical correlations $\rho_i$ defined as
\begin{equation}
    \rho_i := \max_{\bm{w_{\R}}^{(i)},\bm{w_{\Rp}}^{(i)}} \tfrac{\langle \R \bm{w_{\R}}^{(i)}, \Rp \bm{w_{\Rp}}^{(i)} \rangle}{\|\R\bm{w_{\R}}^{(i)}\| \cdot \|\Rp\bm{w_{\Rp}}^{(i)}\|} \quad
    \mathrm{s.t.}~ \R\bm{w_{\R}}^{(j)} \bot \R\bm{w_{\R}}^{(i)},~~\Rp\bm{w_{\Rp}}^{(j)} \bot \Rp\bm{w_{\Rp}}^{(i)}~~ \forall j < i,
\end{equation}
where $\bot$ means orthogonality.
If the representations are (nearly) collinear, regularized \emph{Ridge CCA} \cite{vinod_canonical_1976} can be used.

A single similarity score $m(\R, \Rp)$ is then computed by aggregating the canonical correlations $\rho_i$.
Standard aggregation choices used to quantify neural network similarity are the mean canonical correlation $m_{\operatorname{CCA}}$ \cite{raghu_svcca_2017,kornblith_similarity_2019,hayne2024does} and the mean squared canonical correlation $m_{\operatorname{CCA}^2}$ \cite{kornblith_similarity_2019,hayne2024does}, also called \emph{Yanai's generalized coefficient of determination} \cite{yanai_unification_1974}:
\begin{equation}\label{eq:mean_cca}
    m_{\operatorname{CCA}}(\R,\Rp) = \tfrac{1}{D} \textstyle\sum_{i=1}^D \rho_i, \qquad m_{\operatorname{CCA}^2}(\R,\Rp) = \tfrac{1}{D} \textstyle\sum_{i=1}^D \rho_i^2.
\end{equation}
CCA is invariant to affine transformations \cite{morcos_insights_2018}. If the representations $\R,\Rp$ are equivalent, it holds that $\rho_i=1$ for all $i\in\{1,\dots, D\}$ and thus $m_{\operatorname{CCA}}(\R,\Rp) = 1$ and  $m_{\operatorname{CCA}^2}(\R,\Rp) = 1$.

Other prominent aggregation schemes, though not applied for representational similarity, include the sum of the squared canonical correlations (also known as Pillai's trace \cite{pillai_new_1955}),
Wilk's lambda statistic \cite{wilks_certain_1932},
and the Lawley-Hotelling trace \cite{lawley_generalization_1938,hotelling_most_1935}.
Several more aggregation methods can be applied, and there are numerous variants of CCA measures, including non-linear and multi-view ones---overviews on such variants are provided in the recent survey by \citet{yang_survey_2021} or the tutorial by \citet{uurtio_tutorial_2017}.
In this work, however, we only consider those CCA-based measures that have been used to measure representational similarity of neural networks.

\subheader{Singular Value CCA.}
\citet{raghu_svcca_2017} argued that representations are noisy and that this noise should be removed before conducting CCA on the representations $\R,\Rp$.
Thus, they proposed the \emph{Singular Value CCA (SVCCA)} approach, in which denoised representations are obtained by performing PCA on the representations.
The number $k$ of principal components that are kept is selected such that a fixed relative amount $t$ of the variance in the data, usually 99 percent, is explained.
Afterward, they use standard CCA on the denoised representations.
Thus, letting $\widetilde{\bm{\R}},\widetilde{\R}'$ denote the denoised representations, the average canonical correlation is used as the final similarity measure:
\begin{equation}\label{eq:svcca}
    m_{\operatorname{SVCCA}}(\R, \Rp) = m_{\operatorname{CCA}}(\widetilde{\bm{\R}},\widetilde{\R}').
\end{equation}
Practically, the representations are also mean-centered before the PCA denoising.
Unlike CCA, SVCCA is only invariant to orthogonal transformations, isotropic scaling and translation.
SVCCA is bounded in the interval $[0,1]$, with a score of one indicating perfectly similar representations.

To compute SVCCA efficiently for CNNs with many features, \citet{raghu_svcca_2017} applied a Discrete Fourier Transform on each channel, yielding block-diagonal matrices for CCA computation, which eliminates unneeded operations.

\subheader{Projection Weighted CCA.}\label{sssec:pwcca}
\citet{morcos_insights_2018} proposed Projection Weighted CCA (PWCCA) as an alternative to SVCCA.
They argued that a representational similarity measure should weigh the individual canonical correlations $\rho_i$ by their importance, i.e., the similarity of the \emph{canonical variables} $\bm{Rw_{\R}}^{(i)}$ with the raw representation $\R$.

For that purpose, given mean-centered representations, they defined a weighting coefficient
$\widetilde{\alpha}_i = \textstyle\sum_{j=1}^D |\langle \bm{Rw_{\R}}^{(i)}, \R_{\--,j} \rangle |
$
for every canonical correlation $\rho_i$ that models its importance.
These coefficients are then normalized to weights $\alpha_i = \widetilde{\alpha}_i / \textstyle\sum_j \widetilde{\alpha}_j$, yielding the final representational similarity measure
\begin{equation}\label{eq:pwcca}
    m_{\operatorname{PWCCA}}(\R,\Rp) = \textstyle\sum_{i=1}^{D} \alpha_i \rho_i.
\end{equation}
This measure is asymmetric, since the weights $\alpha_i$ are only computed based on $\R$. Further, it is invariant to isotropic scaling and translation.
PWCCA is bounded in the interval $[0,1]$, with a value of one indicating equivalent representations.

\subsection{Alignment-Based Measures}
The next group of measures stipulates that a pair of representations $\R,\Rp$ can be compared directly once the corresponding representation spaces have been aligned to each other.
Alignment is usually realized by finding an optimal transformation $\varphi\in\mathcal{T}$ that minimizes a difference of the form $\|\varphi(\R) - \Rp \|$.
The exact group of transformations $\mathcal{T}$ used for alignment also directly determines and usually corresponds to the group of transformations that the corresponding measure will be invariant to.
Such direct alignment is only possible if the number of neurons in both representations are equal.
Thus, we assume throughout the next section that $D = D'$, unless otherwise mentioned.
We now discuss existing measures from this category.

\subheader{Orthogonal Procrustes.}\label{sssec:orthoproc}
The orthogonal Procrustes problem is a classical problem of finding the best orthogonal transformation to align two matrices in terms of minimizing the Frobenius norm \myeqref{eq:frobenius_norm} of the difference.
Solving the problem leads to the similarity measure
\begin{equation}\label{eq:procrustes}
    m_{\operatorname{Ortho-Proc}}(\R,\Rp) = \min_{\bm{Q} \in \operatorname{O}(D)} \|\R \bm{Q} - \Rp\|_F = (\|\R\|_F^2 + \|\Rp\|_F^2 - 2 \| \R^\transp\Rp\|_*)^{\frac{1}{2}} ,
\end{equation}
where $\|\cdot\|_*$ denotes the nuclear norm \myeqref{eq:nuclear_norm} of a matrix \cite{schonemann_generalized_1966}.
The second formulation can also be used if $D\neq D'$ \cite{khosla_soft_2024}.
\citet{ding_grounding_2021} used the square of $m_{\operatorname{Ortho-Proc}}$ as a similarity score.
By design, this measure is invariant to orthogonal transformations, and \citet{williams_generalized_2022} showed that this measure satisfies the properties of a distance metric. This also holds when one optimizes Equation \ref{eq:procrustes} over any subgroup $\operatorname{G}(D)\subset \operatorname{O}(D)$.
Notably, considering the subgroup of permutation matrices yields the \emph{Permutation Procrustes} measure \cite{williams_generalized_2022}, also known as \emph{one-to-one matching distance}~\cite{khosla_soft_2024}.

A similar optimization was proposed by \citet{godfrey_symmetries_2022} in their $G_{\operatorname{ReLU}}$-Procrustes measure, which is designed to be invariant to $\operatorname{G}_{\operatorname{ReLU}}$ transformations, a special set of linear transformations (see \Cref{ap:intertwiner}).
\citet{williams_generalized_2022} further proposed a variant that is invariant to spatial shifts in convolutional layers.

\subheader{Generalized Shape Metrics.} \label{subsubsec:generalized_shape_metrics}
\citet{williams_generalized_2022} applied theory of statistical shape analysis on the problem of measuring representational similarity.
In that context, they also defined novel similarity measures.
For representations with unit Frobenius norm~\myeqref{eq:frobenius_norm} and any subgroup $\operatorname{G}(D)\subseteq \operatorname{O}(D)$, they introduced the \emph{Angular Shape Metric}
\begin{equation}\label{eq:angshapemetric}
    m_{\theta}(\R,\Rp) = \min_{\bm{Q} \in \operatorname{G}(D)} \operatorname{arccos} \langle \R \bm{Q}, \Rp\rangle_F,
\end{equation}
which is invariant to transformations from $\operatorname{G}(D)$.
To obtain a more general measure that is not restricted to representations preprocessed to unit norm, they apply the partial whitening function $\phi_\alpha(\R) = \bm{H}_N\R (\alpha \bm{I}_D + (1 - \alpha)(\R^\transp \bm{H}_N \R)^{-1/2})$, where $\alpha\in[0,1]$ and $\bm{H}_N= \bm{I}_N-\tfrac{1}{N}\bm{1}_N\bm{1}_N^\transp$ denotes a centering matrix.
This yields the \emph{Partial Whitening Shape Metric} 
\begin{equation}\label{eq:partialwhiteninghapemetric}
    m_{\theta,\alpha}(\R, \Rp) = \min_{\bm{Q}\in \operatorname{O}(D)} \arccos \tfrac{\langle \phi_\alpha(\R) \bm{Q}, \phi_\alpha(\Rp)\rangle_F}{\|\phi_\alpha(\R)\|_F \|\phi_\alpha(\Rp)\|_F}.
\end{equation}
For all $\alpha>0$, this metric is invariant to orthogonal transformations and translations.
For $\alpha=1$, it is further invariant to isotropic scaling, for $\alpha=0$ it is even invariant to affine transformations.
\citet{williams_generalized_2022} showed that this metric is also related to (regularized) canonical correlations.
Both shape metrics are bounded in the interval $[0,\pi]$ and satisfy the properties of a distance metric.

\citet{duong_representational_2022} generalized the metrics from \citet{williams_generalized_2022} to stochastic neural networks such as variational autoencoders \cite{kingma_auto-encoding_2014}, which map to distributions of representations instead of deterministic representations.
\citet{ostrow2023beyond} further extended this metric to measure similarity of dynamical systems, such as RNNs.

\subheader{Soft Matching Distance.}
\citet{khosla_soft_2024} generalized the Permutation Procrustes measure to settings in which the number of neurons in the representations $\R, \Rp$ differ, i.e., $D \not= D'$.
This was done by interpreting the problem of matching neurons as a transportation problem with possible solutions in the transportation polytope $\operatorname{TP}(D,D')$ \cite{de_combinatorics_2013}.
Thus,  assuming the representations are centered and scaled to unit norm,
they defined the \emph{soft matching distance} as
\begin{equation}
    m_{\operatorname{SoftMatch}}(\R, \Rp) = \textstyle\sqrt{\min_{\bm{P} \in \operatorname{TP}(D,D')} \textstyle\sum_{ij} \bm{P}_{ij}\|\R_{\--,i} - \Rp_{\--,j}\|_2^2}.
\end{equation}
This measure is a special case of the 2-Wasserstein distance, and thus a metric \cite{khosla_soft_2024}.
Further, it is invariant to permutations, translations and scaling.
\citet{khosla_soft_2024} also proposed a variant that is related to Correlation Match \myeqref{eq:corr_match}.

\subheader{Linear Regression.}
An approach similar to Procrustes, but not restricted to orthogonal transformations, is based on predicting one representation from the other with a linear transformation \cite{li_convergent_2016,kornblith_similarity_2019}.
Then, the R-squared score of the optimal fit can be used to measure similarity \cite{kornblith_similarity_2019}.
Assuming mean-centered representations, this yields the measure
\begin{equation}
    m_{R^2}(\R, \Rp) = 1 - \tfrac{\min_{\bm{W}\in \Real^{D \times D}} \|\Rp - \R \bm{W}\|_F^2}{\|\Rp\|_F^2} = \tfrac{\big\|\big(\Rp(\Rp^\transp\Rp)^{-1/2}\big)^\transp \R\big\|_F^2}{\|\R\|_F^2}.
\end{equation}
This asymmetric measure is invariant to orthogonal transformation and isotropic scaling.
A value of one indicates maximal similarity, lower values indicate lower similarity.
This measure has no lower bound.

\citet{li_convergent_2016} added a L1 penalty to the optimization to encourage a sparse mapping between neurons.
\citet{bau_identifying_2019} matched the full representation of one model to a single neuron of another by linear regression.

\subheader{Aligned Cosine Similarity.}
This measure was used to quantify similarity of instance representations, such as embeddings of individual words over time \cite{hamilton_diachronic_2018}.
Its idea is to first align the representations by the orthogonal Procrustes transformation, and then to use cosine similarity \myeqref{eq:cos-sim} to measure similarity between the aligned representations.
Letting $\bm{Q}^*$ denote the solution to the Procrustes problem \myeqref{eq:procrustes}, the similarity of two instance representations is given by
$\operatorname{cos-sim}\left((\R \bm{Q}^*)_i, \Rp_i\right).$
Overall similarity can then be analyzed by comparing the overall distribution of similarity scores, or aggregating them by, for instance, taking their mean value \cite{schumacher_effects_2021}.
The latter option yields a similarity measure
\begin{equation}\label{eq:aligned_cossim}
    m_{\operatorname{Aligned-Cossim}}(\R, \Rp) = \tfrac{1}{N}\textstyle\sum_{i=1}^N \operatorname{cos-sim}\left((\R \bm{Q}^*)_i, \Rp_i\right),
\end{equation}
which is bounded in the interval $[-1,1]$, with $m_{\operatorname{Aligned-Cossim}}(\R, \Rp) = 1$ indicating perfect similarity.
It is invariant to orthogonal transformations and isotropic scaling.

\subheader{Correlation Match.}
\citet{li_convergent_2016} measured representational similarity by creating a correlation matrix between the neuron activations of two representations that are assumed to be mean-centered.
They then matched each neuron $\R_{-,j}$ to the neuron $\Rp_{-,k}$ that it correlated the strongest with.
\citet{wu_similarity_2020} applied strict one-to-one matching, \citet{li_convergent_2016} further used a relaxed version, in which one neuron can correspond to multiple other ones.
Letting $\bm{M}$ denote the matrix that matches the neurons, which is a permutation matrix in strict one-to-one matching,
the average correlation between the matched neurons is given by
\begin{equation}\label{eq:corr_match}
    m_{\text{Corr-Match}}(\R, \Rp) = \tfrac{1}{D}\textstyle\sum_{j=1}^D \tfrac{\langle \R_{-,j}, (\Rp \bm{M})_{-,j} \rangle}{\|\R_{-,j}\|_2 \|(\Rp \bm{M})_{-,j}\|_2},
\end{equation}
This measure is invariant to permutations, isotropic scaling, and translations.
A value of one indicates equivalent representations, a value of zero uncorrelated ones.

\subheader{Maximum Matching Similarity.}
In contrast to the previous measures, \emph{Maximum Matching Similarity} \cite{wang_towards_2018} aligns representations only implicitly and can compare representations of different dimension by testing whether neuron activations of one representation, i.e., columns of the representation matrix, (approximately) lie in a subspace spanned from neuron activations of the other representation.
Every neuron, of which the activation vector can be approximated by such a subspace, is then considered part of a match between the representations.
Following this intuition, the main idea of the measure proposed by \citet{wang_towards_2018} is to find the maximal set of neurons in each representation that can be matched with the other subspace.
Formally, for an index subset $\mathcal{J} \subseteq\{1,\dots,D\}$, let $\R_{\--,\mathcal{J}} = \{\R_{\--,j}, j\in\mathcal{J}\}$ denote the set of corresponding neuron activation vectors.
Then a pair $(\mathcal{J}, \mathcal{J}')$ forms an $\varepsilon$-approximate match, $\varepsilon\in(0,1]$, on the representations $\R, \Rp$ if for all $j\in\mathcal{J}, j'\in\mathcal{J}'$ it holds that
\begin{align}
    \min_{\bm{r}\in\operatorname{span} (\R_{\--,\mathcal{J}})}\| \Rp_{-,j'} - \bm{r} \| \leq \varepsilon\cdot \| \Rp_{\--,j'} \| \quad\text{and}\quad \min_{\bm{r'}\in\operatorname{span} (\Rp_{\--,\mathcal{J}'})}\| \R_{\--,j} - \bm{r'} \| \leq \varepsilon\cdot \| \R_{\--,j} \|.
\end{align}
A pair $(\mathcal{J}_{\max}, \mathcal{J}_{\max}')$ is considered a maximum match, if for all $\varepsilon$-matches $(\mathcal{J}, \mathcal{J}')$ it holds that $\mathcal{J}\subseteq\mathcal{J}_{\max}$ and $\mathcal{J}'\subseteq\mathcal{J}_{\max}'$.
\citet{wang_towards_2018} showed that the maximum match is unique and provided algorithms to determine it.
Based on the maximum match, the \emph{maximum matching similarity} is defined as
\begin{equation}
    m_{\operatorname{maximum-match}}^\varepsilon(\R, \Rp) = \tfrac{|\mathcal{J}_{\max}| + |\mathcal{J}_{\max}'|}{D+D'}.
\end{equation}
This measure is invariant to invertible linear transformation, since such transformations do not alter the subspaces.
It is bounded in the interval $[0,1]$, with a similarity score of 1 indicating maximum similarity.

\subheader{ContraSim.}
Inspired by ideas from contrastive learning, \citet{rahamim_contrasim_2024} proposed a measure that implicitly aligns representations by applying a neural encoder to map them into a joint embedding space.
The encoder was trained using explicitly selected pairs of instances as positive and negative samples, for which the resulting embeddings should and should not be similar, respectively.
In the joint embedding space, similarity is then modeled by the angle between representations, and thus, letting $\operatorname{enc}$ denote the trained encoder network, \emph{ContraSim} is defined as
\begin{equation}\label{eq:contrasim}
    m_{\operatorname{ContraSim}}(\R, \Rp) = \tfrac{1}{N} \textstyle\sum_{i=1}^N \operatorname{cos-sim}\left(\operatorname{enc}(\R_i), \operatorname{enc}(\Rp_i)\right).
\end{equation}
If $\R, \Rp$ have different dimensionality, two different encoders are trained together.
The invariances of the measure are determined via the training examples for the encoder.
The measure is bounded in the interval $[-1,1]$, with a similarity score of 1 indicating maximum similarity.

\subsection{Representational Similarity Matrix-Based Measures}\label{sec:rsm}

A common approach to avoid alignment issues in direct comparisons of representations is to use \emph{representational similarity matrices} (RSMs).
Intuitively, an RSM describes the similarity of the representation of each instance $i$ to all other instances in a given representation $\R$.
The RSMs of two representations $\R,\Rp$ can then be used to quantify representational similarity in terms of the difference between these RSMs.
Formally, given an instance-wise similarity function $s: \Real^D \times \Real^D \longrightarrow\Real$, the RSM $\Sm \in \Real^{N \times N}$ of a representation $\R$ can be defined in terms of its elements via
\begin{equation}\label{eq:rsm}
    \Sm_{i,j}:=s(\R_i, \R_j).
\end{equation}
Each row $\Sm_i$ then corresponds to the similarity between the representations of instance $i$ and the representations of all other inputs, including itself.
RSMs can be computed with a variety of similarity functions $s$ such as cosine similarity \cite{chen_graph-based_2021} or kernel functions \cite{kornblith_similarity_2019}---like before, we do not differentiate between the equivalent concepts of similarity and distance functions.
Naturally, the choice of the underlying similarity function $s$ impacts the kind of transformations that the representational similarity measures $m$ will be invariant to:
if the RSM is unchanged by a transformation, then the representational similarity will not change either.
In Appendix \ref{sec:simfct} we give an overview of commonly used similarity functions, along with the invariances they induce on the RSMs.
After selecting a suitable similarity function $s$, two RSMs $\Sm, \Smp$ are compared.
In the following, we review existing measures that use this approach.

\subheader{Norm of RSM Difference.}
A direct approach to compare RSMs is to apply some matrix norm $\|\cdot\|$ to the difference between RSMs to obtain a measure
\begin{equation}\label{eq:rsm_norm}
    m_{\operatorname{Norm}}(\R, \Rp) = \|\Sm - \Smp\|.
\end{equation}
which assigns a score of zero to equivalent representations, and higher scores to dissimilar representations.
To compute the RSMs, \citet{shahbazi_using_2021} and \citet{yin_dimensionality_2018} used the linear kernel.
In that case, this measure is invariant to orthogonal transformations and satisfies the properties of a distance metric for representations of equal dimensionality.

\subheader{Representational Similarity Analysis.}
\citet{kriegeskorte_representational_2008} proposed \emph{Representational Similarity Analysis} (RSA) in neuroscience.
RSA is a general framework that utilizes RSMs to compare sets of measurements, such as neural representations.
In the first step of this framework, RSMs are computed with respect to an inner similarity function $s_\text{in}$.
Since the RSMs are symmetric, their lower triangles can then be vectorized in a next step to vectors $\mathsf{v}(\Sm)\in\Real^{N(N-1)/2}$.
Finally, these vectors are compared by an outer similarity function $s_\text{out}$:
\begin{equation}\label{eq:rsa}
    m_{\operatorname{RSA}}(\R, \Rp) = s_\text{out}(\mathsf{v}(\Sm), \mathsf{v}(\Smp)).
\end{equation}
This framework can be instantiated with various choices for the similarity functions $s_\text{in}$ and $s_\text{out}$.
This choice however affects the kind of transformations that RSA is invariant to, and further determines the range and interpretation of this measure.
\citet{kriegeskorte_representational_2008} used Pearson correlation \myeqref{eq:pearson} as inner similarity function $s_\text{in}$ to compute the RSMs, and Spearman correlation as outer similarity function $s_\text{out}$, since these correlation measures induce invariance to scaling and translations.
\citet{kriegeskorte_representational_2008} further suggested functions such as Euclidean or Mahalanobis distance.

\subheader{Centered Kernel Alignment.}\label{sssec:cka}
\citet{kornblith_similarity_2019} proposed \emph{Centered Kernel Alignment} (CKA) \cite{cortes_algorithms_2012, cristianini_kernel-target_2001} to measure representational similarity.
CKA uses kernel functions on mean-centered representations to compute the RSMs, which are then compared via the Hilbert-Schmidt Independence Criterion (HSIC) \cite{gretton_measuring_2005}.
Given two RSMs $\Sm, \Smp$, the HSIC can be computed via $\operatorname{HSIC}(\Sm, \Smp) = \tfrac{1}{(N-1)^2} \operatorname{tr}(\Sm\bm{H}_N\Smp\bm{H}_N)$, where $\bm{H}_N = \bm{I}_N - \tfrac{1}{N} \bm{1}_N\bm{1}_N^\transp$ denotes a centering matrix. 
Recent work \cite{murphy_correcting_2024} highlights the importance of using the debiased HSIC estimator of \citet{song_feature_2012}, especially when $N<D$.
Then, a normalization of the HSIC yields the CKA measure:
\begin{equation}\label{eq:cka}
    m_{\operatorname{CKA}}(\R,\Rp) = \tfrac{\operatorname{HSIC}(\Sm, \Smp)}{\sqrt{\operatorname{HSIC}(\Sm, \Sm) \mathrm{HSIC}(\Smp, \Smp)}}.
\end{equation}
CKA is bounded in the interval $[0,1]$, with $m_{\operatorname{CKA}}(\R,\Rp)=1$ indicating equivalent representations.
\citet{kornblith_similarity_2019} computed the RSMs from the linear kernel and tested the RBF kernel without reporting large differences in results.
\citet{saini_subspace_2021} used so-called affinity matrices, which result from sparse subspace clustering \cite{elhamifar_sparse_2013} of the representations, instead of RSMs.
The standard linear version is invariant to orthogonal transformations and isotropic scaling.

CKA with linear kernel is equivalent to the RV coefficient, a statistical measure to compare data matrices \cite{robert_unifying_1976, kornblith_similarity_2019}.
It can also be seen as a variant of PWCCA \myeqref{eq:pwcca} with an alternative weighting scheme, with the advantage that it does not require a matrix decomposition to be computed \cite{kornblith_similarity_2019}.
Further, \citet{godfrey_symmetries_2022} proposed the $G_{\operatorname{ReLU}}$-CKA variant that is specific to models that use ReLU activations, and invariant to $G_{\operatorname{ReLU}}$ transformations \myeqref{eq:intertwiner_transformers}.

\subheader{Distance Correlation.}
\emph{Distance Correlation} (dCor) \cite{szekely_measuring_2007} is a non-linear correlation measure that tests dependence of two vector-valued random variables $X$ and $Y$ with finite mean.
In the context of our survey, we consider the instance representations as samples of such random variables.
To determine the distance correlation of two representation matrices $\R, \Rp$, one first computes the RSMs $\Sm, \Smp$ using Euclidean distance as similarity function $s$.
Next, the RSMs are mean-centered in both rows and columns, which yields $\widetilde{\Sm}, \widetilde{\Sm}'$.
Then the squared sample distance covariance of the RSMs $\Sm,\Smp$ can be computed via $\operatorname{dCov}^2(\Sm,\Smp)=\tfrac{1}{N^2}\textstyle\sum_{i=1}^N\textstyle\sum_{j=1}^N \widetilde{\Sm}_{i, j} \widetilde{\Sm}'_{i,j}.$
Finally, the squared distance correlation is defined as
\begin{equation}
    m_{\operatorname{dCor}}^2(\R, \Rp)=\tfrac{\operatorname{dCov}^2(\Sm, \Smp)}{\sqrt{\operatorname{dCov}^2(\Sm, \Sm) \operatorname{dCov}^2(\Smp, \Smp)}}.
\end{equation}
A distance correlation of zero indicates statistical independence between the representations $\R$ and $\Rp$.
Due to the usage of Euclidean distance as similarity function $s$, dCor is invariant to orthogonal transformations and translations.

\citet{lin_geometric_2022} considered a variant called \emph{Adaptive Geo-Topological Independence Criterion} (AGTIC) \cite{lin_adaptive_2020}, which rescales the values in the RSMs with respect to an upper and lower threshold to eliminate noise.

\subheader{Normalized Bures Similarity.}
This measure was inspired by the Bures distance, which has its roots in quantum information theory \cite{bures_extension_1969} and satisfies the properties of a distance metric on the space of positive semi-definite matrices \cite{bhatia_bures-wasserstein_2017}.
As \citet{tang_similarity_2020} used the linear kernel to compute the RSMs $\Sm, \Smp$, these matrices are positive semi-definite.
Hence, these matrices also have a unique square root.
Therefore, they could define the \emph{Normalized Bures Similarity} as
\begin{equation}\label{eq:nbs}
    m_{\text{NBS}}(\R, \Rp)= \tfrac{\operatorname{tr}(\Sm^{1/2}\Smp \Sm^{1/2})^{1/2}}{\sqrt{\operatorname{tr}(\Sm)\operatorname{tr}(\Smp)}}.
\end{equation}
This measure is bounded in the interval $[0,1]$, with $m_{\text{NBS}}(\R, \Rp) = 1$ indicating perfect similarity.
Due to use of the linear kernel, it is invariant to orthogonal transformations, and further invariant to isotropic scaling due to the normalization.

One can show that NBS is equivalent---up to an arc cosine---to the angular shape metric \myeqref{eq:angshapemetric}, and that the unnormalized Bures distance is equal to the orthogonal Procrustes measure \cite{harvey_duality_2024}.

\subheader{Eigenspace Overlap Score.}
\citet{may_downstream_2020} proposed the \emph{Eigenspace Overlap Score} (EOS) as a criterion to select compressed word embeddings with best downstream performance.
EOS compares RSMs by comparing the spaces spanned from their eigenvectors.
Assuming full-rank representations $\R,\Rp$, they compute the RSMs $\Sm$, $\Smp$ using the linear kernel \myeqref{eq:linkernel}.
Letting $\bm{U}\in\Real^{N\times D},\bm{U'}\in \Real^{N\times D'}$ denote the matrices of eigenvectors that correspond to the non-zero eigenvalues of $\Sm,\Smp$, respectively, the measure is defined as
\begin{equation}\label{eq:eos}
    m_{\text{EOS}}(\R,\Rp)=\tfrac{1}{\max(D, D')}\|\bm{U}^\transp \bm{U'}\|_F^2.
\end{equation}
EOS indicates minimal similarity with a value of zero when the spans of $\bm{U}$ and $\bm{U'}$ are orthogonal, and maximal similarity with a value of one when the spans are identical.
This measure is invariant to invertible linear transformations.

EOS is related to the expected difference in generalization error of two linear models that are each trained on one of the representations \cite{may_downstream_2020}, similar to the following measure.

\subheader{Unified Linear Probing (GULP).}\label{sssec:gulp}
\emph{GULP} quantifies similarity by measuring how differently linear regression models that use either the representation $\R$ or the representation $\Rp$ \cite{boix-adsera_gulp_2022} can generalize.
This is done by considering all regression functions $\eta$ on the original instances $\bm{X}$ that are bounded so that $\|\eta\|_{L^2} \leq 1$, and trying to replicate these relations via ridge regression on the representations $\R$ and $\Rp$.
The similarity measure is then defined as the supremum of the expected discrepancy in the predictions of ridge regression models trained to approximate $\eta$ with $\R$ or $\Rp$ as inputs, taken over all regression functions $\eta$.

Practically, \citet{boix-adsera_gulp_2022} proved that there is a closed form expression to estimate this value in terms of the covariance matrices of the representations, where it is assumed that the representations are mean-centered in the columns, and that their rows have unit norm.
Letting the RSMs $\Sm=\frac{1}{N}\R^\transp \R$ denote the matrix of covariance within a representation, $\bm{S_{\R,\Rp}}=\frac{1}{N}\R^\transp \Rp$ the cross-covariance matrix, and $\Sm^{-\lambda}= (\Sm + \lambda\bm{I}_D)^{-1}$ the inverse of a regularized covariance matrix, the GULP measure can be computed as
\begin{equation}\label{eq:gulp}
    m_{\operatorname{GULP}}^\lambda(\R,\Rp) = \Big(\tr(\Sm^{-\lambda}\Sm \Sm^{-\lambda}\Sm)
    + \tr(\Smp^{-\lambda}\Smp \Smp^{-\lambda}\Smp)
    - 2 \tr(\Sm^{-\lambda} \Sm_{\R,\Rp} \Smp^{-\lambda} \Sm_{\R,\Rp}^\transp)\Big)^{1/2}.
\end{equation}
The hyperparameter $\lambda\geq 0$ corresponds to the regularization weight of the ridge regression models over the representations.
For all $\lambda\geq 0$, GULP is unbounded, satisfies the properties of a distance metric, and is invariant to orthogonal transformations, scaling, and translations.
For $\lambda = 0$, GULP is invariant to affine transformations, and can further be expressed as a linear transformation of the mean-squared CCA measure \myeqref{eq:mean_cca}.

\emph{Transferred Discrepancy} (TD) \cite{feng_transferred_2020} used an approach similar to GULP by measuring the discrepancy of linear classifiers, instead of linear regression models.
TD is also related to the mean-squared CCA measure \myeqref{eq:mean_cca}.

\subheader{Riemannian Distance.}
This measure considers the special geometry of symmetric positive definite (SPD) matrices, which lie on a Riemannian manifold \cite{bhatia_positive_2007}.
Every inner product defined on a Riemannian manifold induces a distance metric that considers the special curvature of these structures.
On the manifold of SPD matrices,
\begin{equation}\label{eq:riemann}
    m_{\operatorname{Riemann}}(\R, \Rp) = \sqrt{\textstyle\sum_{i=1}^N \log^2(\lambda_i)},
\end{equation}
denotes such a metric, where $\lambda_i$ is the $i$-th eigenvalue of $\Sm^{-1}\Smp$.
\citet{shahbazi_using_2021} proposed this measure using RSMs defined as $\Sm=\R\R^\transp/D$.
This matrix however can only be positive definite if $D>N$, which limits applicability of this measure.
This measure is invariant to orthogonal transformations.
Equivalence is indicated by a value of zero, and larger values indicate dissimilarity.

\subheader{Relational Knowledge Loss.}
A common approach to transfer knowledge in the context of knowledge distillation is to train the student model to mimic the relations between the teacher's instance representations \cite{gou2021knowledge}.
This is done by minimizing the total element-wise difference of RSMs with respect to a loss function $l: \Real \times \Real \longrightarrow \Real_+$:
\begin{equation}
    m_{\operatorname{RK}}(\R, \Rp)= \textstyle\sum_{i,j = 1}^N l\big(\Sm_{i,j}, \Smp_{i,j} \big).
\end{equation}
While we defined RSMs via pairwise similarities, this approach has notably been generalized to higher-dimensional RSMs.
For example, \citet{park_relational_2019} considered three-dimensional RSMs $\Sm\in\Real^{N \times N\times N}$, where each entry $\Sm_{i,j,k}$ corresponds to the cosine of the angle enclosed by the vectors $v_{i,j} = \R_i - \R_j$ and $v_{k,j} = \R_k-\R_j$, i.e., $\Sm_{i,j,k} = \operatorname{cos-sim}(v_{i,j}, v_{j,k})$.
Similarities are then aggregated over all instance triples.
This measure instantiation is invariant to orthogonal transformations,  translations, and scaling.
Equivalence is indicated by a value of zero, larger values indicate dissimilarity.

\subsection{Neighborhood-Based Measures}
The measures in this section compare the nearest neighbors of instances in the representation space.
More precisely, each of these measures determine the $k$ nearest neighbors of each instance representation $\R_i$ in the full representation matrix $\R$ with respect to a given similarity function $s$. In that context, the neighborhood size $k$ is a parameter that has to be chosen for the application at hand.
Letting $\Sm$ denote the RSM of representation $\R$, and w.l.o.g. assuming that higher values indicate more similar representations, we formally define the set of the $k$ nearest neighbors of the instance representation $\R_i$ as the set $\mathcal{N}^k_{\R}(i)  \subset \{j: 1\leq j \leq N, j\not=i\}$ with $|\mathcal{N}^k_{\R}(i)| = k$ for which it holds that $\Sm_{i,j} > \Sm_{i,l}$ for all $j\in \mathcal{N}^k_{\R}(i), l\not\in \mathcal{N}^k_{\R}(i)\cup\{i\}$.
Once the nearest neighbors sets are determined, they are either compared directly, or one further considers distances of the representation $\R_i$ to its nearest neighbors.
For each of these measures, we then obtain a vector of instance-wise neighborhood similarities $\big(v_{\operatorname{NN-sim}}^k(\R,\Rp)_i\big)_{i\in\{1,\dots,N\}}$, which are averaged over all instances to obtain similarity measures for the full representations $\R,\Rp$:
\begin{equation}
    m_{\operatorname{NN-sim}}^k(\R,\Rp) = \tfrac{1}{N} \textstyle\sum_{i=1}^N v_{\operatorname{NN-sim}}^k(\R,\Rp)_i.
\end{equation}
However, the instance-wise similarities and their distribution could also be inspected more closely to obtain additional insights \cite{kolling_pointwise_2023}.
For brevity, in all the measures that we introduce in the following, we only give a description of how the instance-wise similarities are computed.
Similar to RSM-based measures, the choice of the similarity function $s$ determines which transformations these measures are invariant to.
By default, and in line with the literature, we assume the use of cosine similarity~\myeqref{eq:cos-sim}, which leads to invariance to orthogonal transformations and isotropic scaling.

\subheader{$k$-NN Jaccard Similarity.}
This measure, also named \emph{Nearest Neighbor Graph Similarity} \cite{gwilliam_beyond_2022} and \emph{Nearest Neighbor Topological Similarity} \cite{hryniowski_inter-layer_2020}, considers how many of the $k$ nearest neighbors each instance has in common over a given pair of representations.
The instance-wise neighborhood similarities are computed in terms of the Jaccard similarities of the neighborhood sets $\mathcal{N}^k_{\R}(i),\mathcal{N}^k_{\Rp}(i)$:
\begin{equation} \label{eq:jaccard}
\big(\bm{v}_\text{Jac}^k\big(\R, \Rp\big)\big)_i :=
\tfrac{|\mathcal{N}^k_{\R}(i)\cap \mathcal{N}^k_{\Rp}(i)|}{|\mathcal{N}^k_{\R}(i)\cup \mathcal{N}^k_{\Rp}(i)|}.
\end{equation}
Jaccard similarity is bounded in the interval $[0,1]$, with a value of one indicating identical neighborhoods.
Aside from the commonly used cosine similarity \cite{schumacher_effects_2021, wang_towards_2020}, Euclidean distance was also used as similarity function \cite{hryniowski_inter-layer_2020}.

\subheader{Second-Order Cosine Similarity.}
This measure was proposed by \citet{hamilton_cultural_2016} to analyze changes in word embeddings over time.
For each instance $i$, it first computes the union of nearest neighbors as an ordered set $\{j_1,\dots, j_{K(i)}\}:=\mathcal{N}^k_{\R}(i)\cup \mathcal{N}^k_{\Rp}(i)$ in $\R$ and $\Rp$ in terms of cosine similarity \myeqref{eq:cos-sim}.
Then the cosine similarities to these neighbors are compared between the two representations.
Utilizing the cosine similarity RSMs $\Sm, \Sm'$ of the representations $\R, \Rp$, the instance-wise second-order cosine similarities can then be defined as follows:
\begin{equation}\label{eq:2ndcos}
\big(\bm{v}_\text{2nd-cos}^k\big(\R, \Rp\big)\big)_i:=  \quad \operatorname{cos-sim}\big(\big(\Sm_{i,j_1},\dots, \Sm_{i,j_{K(i)}}\big),\big(\Sm'_{i,j_1},\dots, \Sm'_{i,j_{K(i)}}\big)\big). \nonumber
\end{equation}
This measure is bounded in the interval $[0,1]$, with $m_\text{2nd-cos}^k(\R, \Rp) = 1$ indicating equivalence of $\R$ and $\Rp$.

Rather than considering the union of the neighborhood sets, \citet{chen_graph-based_2021} considered the intersection of the top-$k$ neighborhoods.
Another similar approach was presented by \citet{moschella_relative_2023}, who used a random fixed set of reference instances instead of neighbors.
Further, \emph{Pointwise Normalized Kernel Alignment} (PNKA) \cite{kolling_pointwise_2023} can be seen as a variant of second-order cosine similarity with $k=N$, but different similarity function $s$ for the RSM.

\subheader{Rank Similarity.}
The $k$-NN Jaccard similarity captures the extent to which two neighborhood sets overlap, but not the order of the common neighbors within those sets.
To increase the importance of close neighbors, \citet{wang_towards_2020} determined distance-based ranks $r_{\R_i}(j)$ to all $j\in\mathcal{N}^k_{\R}(i)$, where $r_{\R_i}(j) = n$ if $\R_j$ is the $n$-th closest neighbor of $\R_i$ with respect to a given similarity function $s$.
Based on these ranks, they defined the instance-based similarities as
\begin{equation}\label{eq:ranksim}
    \big(\bm{v}^k_{\operatorname{ranksim}}(\R, \Rp)\big)_i = \tfrac{1}{(\bm{v}_{\max})_i}
    \cdot \textstyle\sum_{j \in \mathcal{N}^k_{\R}(i) \cap \mathcal{N}^k_{\Rp}(i)} \tfrac{2}{(1 + |r_{\R_i}(j) - r_{\Rp_i}(j)|) (r_{\R_i}(j) + r_{\Rp_i}(j))},
\end{equation}
where $(\bm{v}_{\max})_i=\textstyle\sum_{k=1}^{K}\tfrac{1}{k}$, with $K=|\mathcal{N}^k_{\R}(i) \cap \mathcal{N}^k_{\Rp}(i)|$, is a normalization factor that limits the maximum of the ranking similarity to one, which is achieved for completely identical rankings.
Intuitively, the first factor of the denominator in \Cref{eq:ranksim} measures the similarity of the ranks of an instance, whereas the second factor assigns rank-based weights to this similarity, with lower-ranked instances gaining less influence.

\subheader{Joint Rank and k-NN Jaccard Similarity.}
Rank similarity has the issue that it is only calculated on the intersection of the $k$-nearest neighbor sets in different representations.
That means rank similarity might be high, even if the $k$-NN sets have almost no overlap.
Similarly, Jaccard similarity might be high, but the order of the nearest neighbors might be completely different.
Therefore, \citet{wang_towards_2020} combined these two approaches to calculate the \emph{Embedding Stability}, by considering the product of Jaccard and rank similarity.
Thus, using the instance vectors defined in \Cref{eq:jaccard} and  \Cref{eq:ranksim}, we can define the vector of instance-wise similarities as 
\begin{equation}
    \big(\bm{v}^k_{\operatorname{Jac-Rank}}(\R, \Rp)\big)_i = \big(\bm{v}_\text{Jac}^k\left(\R, \Rp\right)\big)_i \cdot \big(\bm{v}^k_{\operatorname{ranksim}}(\R, \Rp)\big)_i.
\end{equation}
Scores are bounded in the interval $[0,1]$, with $m_{\operatorname{Jac-Rank}}^k(\R, \Rp)=1$ indicating perfect similarity.

\subsection{Topology-Based Measures}\label{sec:topological}

The measures in this category are motivated by the \emph{manifold hypothesis} \cite[Sec. 5.11.3]{goodfellow_deep_2016}, which states that high-dimensional representations are expected to be concentrated in the vicinity of a low-dimensional data manifold $\mathcal{M}$.
Following this assumption, these measures then aim to approximate the manifolds in terms of discrete topological structures such as graphs, or, more generally, (abstract) simplicial complexes \cite{hatcher_algebraic_2005}, based on which the representations can then be compared.
A simplicial complex can be seen as a generalization of graphs, in which vertices may not only be paired by edges, but can also form higher-dimensional simplices.
In both cases, each instance $i$ typically corresponds to a vertex $v_i\in\mathcal{V}$, and edges/simplices are formed from instances that are close together in the representation space.

\subheader{Geometry Score.}
The \emph{Geometry Score} (GS) \cite{khrulkov_geometry_2018} characterizes representations by the number of \emph{one-dimensional holes} in their data manifolds.
To obtain this number of holes, the manifold is approximated in terms of simplicial complexes $\mathcal{S}_\alpha$, $\alpha>0$, in which vertices $v_i$ form a simplex, if the $\alpha$-neighborhoods of their representations $\R_i$ overlap with each other.
On this simplicial complex, the number of one-dimensional holes corresponds to the number of specific cycles in the complex and can be efficiently computed as the rank of its first homology group $H_1$.

Given that the number of holes may differ dependent on $\alpha$, and that there is no ground-truth regarding which value of $\alpha$ yields the most accurate approximation of the manifold, \citet{khrulkov_geometry_2018} suggest varying the value $\alpha$ between 0 and $\alpha_{\max}\propto \max_{i,j}\|\R_i-\R_j\|_2$.
For each number of holes $k$, they collect the longest intervals $(\alpha_1, \alpha_2)$, in which the number of holes is constant at $k$, into sets $\mathcal{B}_k$.
Then, the \emph{relative living time} of $k$ holes defined as
$\operatorname{RLT}(k, \R) = \frac{1}{\alpha_{\max}}\sum_{(\alpha_1, \alpha_2)\in\mathcal{B}_k} (\alpha_2-\alpha_1)$ can be considered as the probability that $k$ holes exist in the manifold.
Since building simplicial complexes from large data is computationally challenging, \citet{khrulkov_geometry_2018} suggested sampling numerous subsets  $\mathcal{I}$ of $n<N$ instances to built multiple so-called witness complexes with much lower number of simplices.
Finally, one then considers the mean relative living times (MRLT) resulting from these complexes:
\begin{equation}
    m_{\text{GS}}(\R,\Rp)= \textstyle\sum_{k=0}^{k_{\max} - 1}(\operatorname{MRLT}(k, \R) - \operatorname{MRLT}(k, \Rp))^2,
\end{equation}
where $k_{\max}$ denotes the maximum number of holes that is considered.
The authors suggested using $k_{\max}=100$,
aggregating the RLTs from 10,000 complexes of $n=64$ vertices each,
and setting $\alpha_{\max} = \tfrac{1}{128} / \tfrac{N}{5000} \cdot  \max_{i,j\in\mathcal{I}}\|\R_i-\R_j\|_2$ for each sample $\mathcal{I}$.
This measure is bounded in the interval $[0,k_{\max}]$, with $m_{\text{GS}}(\R,\Rp)=0$ indicating equivalent representations. Further, it is invariant to orthogonal transformations, isotropic scaling, and translations.

\subheader{Multi-Scale Intrinsic Distance.}
The \emph{Multi-Scale Intrinsic Distance} (IMD) \cite{tsitsulin_shape_2020} applies $k$-NN graphs $\mathcal{G}(\R)$ as a proxy to characterize and compare the manifold of the representations.
Specifically, \citet{tsitsulin_shape_2020} utilize the \emph{heat kernel trace} on $\mathcal{G}(\R)$ to compare representations, which is defined as $\operatorname{hkt}_{\mathcal{G}(\R)}(t)=\sum_i e^{-t\lambda_i}$, with $\lambda_i$ as the eigenvalues of the normalized graph Laplacian of $\mathcal{G}(\R)$.
Similarity between the manifolds is then computed as a lower bound of the Gromov-Wasserstein distance, which can be expressed in terms of the heat kernel trace:
\begin{equation}\label{eq:imd}
    m_{\text{IMD}}(\R, \Rp) = \sup_{t>0} e^{-2(t+t^{-1})} | \operatorname{hkt}_{\mathcal{G}(\R)}(t) - \operatorname{hkt}_{\mathcal{G}(\Rp)}(t)|.
\end{equation}
Practically, \citet{tsitsulin_shape_2020} approximate $\operatorname{hkt}_{\mathcal{G}(\R)}(t)$ using the \emph{Stochastic Lanczos Quadrature} \cite{ubaru_fast_2017}, and obtain the supremum by sampling $t$ from a parameter grid.
They built the graph using the $k=5$ nearest neighbors with respect to Euclidean distance.
The IMD has no upper bound; the minimum, which indicates maximal similarity, is zero.
It is invariant to orthogonal transformations, isotropic scaling and translations.

\subheader{Representation Topology Divergence.}\label{sec:rtd}
Similar to the geometry score, \emph{Representation Topology Divergence} (RTD) \cite{barannikov_representation_2022} also considers persistence intervals of topological features of representations.
However, in this approach graphs are applied for simplicial approximation of the representations, and the number of their connected components are the topological feature of interest.
Specifically, \citet{barannikov_representation_2022} first compute RSMs with Euclidean distance, and normalize these by the 90th percentile of their values.
Then, for a given distance threshold $\alpha>0$, they construct a graph $\mathcal{G}^\alpha(\R)$ with its adjacency matrix $\bm{A}$ defined as $\bm{A}_{i,j} = \Sm_{i,j}\cdot \bbone\{\Sm_{i,j}<\alpha\}$, and a union graph $\mathcal{G}^\alpha(\R,\R')$ with its adjacency matrix $\bm{A}$ defined as $\bm{A}_{i,j} = \min(\Sm_{i,j},\Smp_{i,j})\cdot \bbone\{\min(\Sm_{i,j},\Smp_{i,j})<\alpha\}$.
If $\mathcal{G}^\alpha(\R)$ and $\mathcal{G}^\alpha(\R,\R')$ differ in the number of their connected components, this is considered a topological discrepancy.
For each specific discrepancy that occurs for varying values of $\alpha$, the longest corresponding interval 
$(\alpha_1, \alpha_2)$, for which this discrepancy persists, is collected in a set $\mathcal{B}(\R, \Rp)$.
The total length of these intervals, denoted as $ b(\R, \Rp)=\textstyle\sum_{(\alpha_1,\alpha_2) \in \mathcal{B}(\R, \Rp)} \alpha_2-\alpha_1$,
then quantifies similarity between two representations.
The final RTD measure is constructed by subsampling $K$ subsets $\mathcal{I}^{(k)}$ of $n<N$ instances each, and collecting the values $b\big(\R^{(k)}, \R'^{(k)}\big)$ derived from the representations $\R^{(k)} = (\R_i)_{i\in\mathcal{I}^{(k)}}\in\Real^{n\times D}$ to form a measure
$RTD(\R, \Rp) = \tfrac{1}{K}\textstyle\sum_{i=1}^K b\big(\R^{(k)}, \R'^{(k)}\big).$
Because RTD is asymmetric, the authors proposed to use
\begin{equation}\label{eq:rtd}
    m_{\text{RTD}}(\R, \Rp) = \tfrac{1}{2} (RTD(\R, \Rp) + RTD(\Rp, \R)).
\end{equation}
For hyperparameters, they suggested using $K=10$ subsets of $n=500$ representations each as default values.
An RTD of zero indicates equivalent representations, with higher values indicating less similarity.
By construction of the RSMs, RTD is invariant to orthogonal transformations, isotropic scaling, and translations.

\subsection{Descriptive Statistics}

Measures of this category deviate from all previous measures in a way that they describe statistical properties of either (i) individual representations $\R$, or (ii) measures of variance in the instance representations $\R_i$ over sets $\mathcal{R}$ of more than two representations.
In case of (i), the statistics can be directly compared over pairs or sets of representations.
For case (ii), one could aggregate or analyze the distribution of the instance-wise variations.
While there are numerous statistics that could be used to compare representations, in the following we specifically outline statistics that have already been used to characterize representations in existing literature.

\subheader{Intrinsic Dimension.}
The \emph{intrinsic dimension} of a representation $\R$ corresponds to the minimal number of variables that are necessary to describe its data points.
It can be defined as the lowest value $M\in\Nat, M<N$, for which the representation $\R$ lies in a $M$-dimensional manifold of $\Real^N$ \cite{camastra_intrinsic_2016}.
This statistic has its roots in social sciences \cite{shepard_analysis_1962} and information theory \cite{bennet_intrinsic_1969}, and has since been applied in countless other fields, resulting in different variants, and numerous methods to estimate its exact values---for more details, we point the interested reader to the survey by \citet{camastra_intrinsic_2016}.
In the context of neural network analysis, different variants of the intrinsic dimension have been used as a tool to analyze the amount of information that is contained and processed within high-dimensional layers \cite{ma_dimensionality_2018, ansuini_2019_intrinsic,basile2024intrinsic}.
This statistic is invariant to affine transformations.

\subheader{Magnitude.}
\citet{wang_towards_2020} characterized \emph{magnitude} as the Euclidean length of instance representations $\R_i$.
Consequently, they considered the length of the mean instance representation as a statistic for a representation $\R$:
\begin{equation}
    m_{\operatorname{Mag}}(\R) := \|\tfrac{1}{N}\textstyle\sum_{i=1}^N\R_i\|_2.
\end{equation}
Aside from aggregating magnitude over all instances, they further proposed a measure to quantify the variance of the magnitude of instance-wise representations over multiple models.
More precisely, given a set of representations $\mathcal{R}$, \citet{wang_towards_2020} measured the variance in the magnitudes of individual instances $i$ as
\begin{equation}
    m_{\operatorname{Var-Mag}}(\mathcal{R},i)=\tfrac{1}{\max_{\R\in\mathcal{R}}\|\R_i\|_2 - \min_{\R\in\mathcal{R}}\|\R_i\|_2}
    \cdot \sqrt{\tfrac{1}{|\mathcal{R}|}\textstyle\sum_{\R \in \mathcal{R}}(\|\R_i\|_2 - \ols{d_i}(\mathcal{R}))^2},
\end{equation}
where $\ols{d_i}(\mathcal{R})=\tfrac{1}{|\mathcal{R}|}\textstyle\sum_{\R \in \mathcal{R}}\|\R_i\|_2$ is the average magnitude of the representations of instance $i$ in $\mathcal{R}$.
As magnitude is unaffected by transformations that preserve vector length, this statistic is invariant to orthogonal transformations.

\subheader{Concentricity.}
\citet{wang_towards_2020} proposed \emph{concentricity} as a measure of the density of representations.
It is based on measuring the cosine similarities of each instance representation $\R_i$ to the average representation, which we denote as $\alpha_i(\R)=\operatorname{cos-sim}(\R_i, \tfrac{1}{N}\sum_{j=1}^N{\R_j})$. 
Similar to magnitude, \citet{wang_towards_2020} then considered the mean concentricity
\begin{equation}
    m_{\operatorname{mConc}}(\R) := \tfrac{1}{N} \textstyle\sum_{i=1}^N  \alpha_i(\R)
\end{equation}
as a statistic for a single model,
and measured the instance-wise variance of concentricity via
\begin{equation}
    m_{\operatorname{Var-Conc}}(\mathcal{R},i)=\tfrac{1}{\max_{\R\in\mathcal{R}}\alpha_i(\R)\ - \min_{\R\in\mathcal{R}}\alpha_i(\R)}
    \cdot \sqrt{\tfrac{1}{|\mathcal{R}|}\textstyle\sum_{\R \in \mathcal{R}}(\alpha_i(\R) - \ols{\alpha}_i(\mathcal{R}))^2},
\end{equation}
where $\ols{\alpha_i}(\mathcal{R})=\tfrac{1}{|\mathcal{R}|}\textstyle\sum_{\R \in \mathcal{R}}\alpha_i(\R)$ is the average concentricity of instance $i$ in $\mathcal{R}$.
Concentricity inherits from cosine similarity the invariances to orthogonal transformations and isotropic scaling.

\subheader{Uniformity.}
\emph{Uniformity} \cite{wang_understanding_2020,gwilliam_beyond_2022} quantifies density of representations by measuring how close the distribution of instance representations is to a uniform distribution on the unit hypersphere. This measure is defined as
\begin{equation}
m_{\operatorname{uniformity}}(\R)=\log \left( \tfrac{1}{N^2}\textstyle\sum_{i=1}^N\textstyle\sum_{j=1}^N e^{-t \|\R_i - \R_j\|_2^2} \right),
\end{equation}
where $t$ is a hyperparameter that was set to $t=2$ by \citet{wang_understanding_2022} and \citet{gwilliam_beyond_2022}.
The statistic is bounded in the interval $[0,1]$, with $m_{\operatorname{uniformity}}(\R)=1$ indicating perfectly uniform representations. Uniformity is invariant to orthogonal transformations and translation, as these transformations preserve distances.

\subheader{Tolerance.}
This statistic considers the proximity of representations of semantically similar inputs \cite{wang_understanding_2021}.
In contrast to the previous statistics, it requires a vector of ground-truth labels $\bm{y}\in\Real^N$.
Further, it is assumed that all instance representations have unit norm.
\emph{Tolerance} is computed as the mean similarity of inputs with the same class:
\begin{equation}
    m_{\operatorname{tol}}(\R)=\tfrac{1}{N^2}\textstyle\sum_{i=1}^N\textstyle\sum_{j=1}^N (\R_i^\transp \R_j) \cdot \bbone\{\bm{y}_i=\bm{y}_j\}.
\end{equation}
Tolerance is bounded in the interval $[-1,1]$, with  $m_{\operatorname{tol}}(\R)=0$ indicating that representations that share the same label are always uncorrelated. This statistic is invariant to orthogonal transformations and isotropic scaling.

\subheader{Instance-Graph Modularity.}
Similar to the topology-based measures (see \Cref{sec:topological}), \citet{saini_subspace_2021} and \citet{lu_understanding_2022} proposed measures based on building a graph to model representations, though this was not motivated from a topological perspective.
Specifically, they used \emph{modularity} \cite{newman_finding_2004} to identify whether semantically similar inputs are close together in the graph, and consequently, the representation space.
In both cases, a sparse graph was constructed.
\citet{saini_subspace_2021} used the affinity matrix resulting from sparse subspace clustering \cite{elhamifar_sparse_2013} as the adjacency matrix $\bm{A}\in\Real^{N\times N}$,
whereas \citet{lu_understanding_2022} determined the adjacency matrix element-wise via $\bm{A}_{i,j} = \Sm_{i,j} \cdot \bbone\{j \in \mathcal{N}_{\R}^{k}(i)\}$,
considering a cosine similarity-based RSM $\Sm$.
The modularity of the network, and in consequence the statistic for $\R$, is then defined as
\begin{equation}\label{eq:modularity}
    m_\text{Mod}(\R) = \tfrac{1}{2W} \textstyle\sum_{i,j}\left(\bm{A}_{i,j} - \tfrac{d_i d_j}{W}\right)\cdot \bbone\{\bm{y}_i=\bm{y}_j\},
\end{equation}
where $d_i = \textstyle\sum_j \bm{A}_{i,j}$ denotes the effective degree of node $v_i$,  $W=\textstyle\sum_{i,j} \bm{A}_{i,j}$ is a normalization factor, and $\bm{y}$ is the vector of ground-truth labels.
The maximum modularity is given by 1, and high modularity implies that nodes of the same label are highly connected with each other, with only few connections to nodes of another label.
Both variants are invariant to orthogonal transformations.
The variant by \citet{lu_understanding_2022} is additionally invariant to isotropic scaling.

\subheader{Neuron-Graph Modularity.}
\citet{lange_clustering_2022} also considered modularity as a statistic to characterize representations.
However, in their approach, the nodes $v_j$ represented neurons $\R_{-,j}$ instead of instance representations $\R_i$.
To model the similarity of neurons that is needed to construct the graphs, they proposed four different variants of RSMs that either consider pure neuron activations or also gradients with respect to neuron activations.
In that latter case, one may consider the modularity based on such RSMs as a hybrid measure of representational and functional characteristics.

Once an RSM $\Sm \in \Real^{D\times D}$ has been computed, \citet{lange_clustering_2022} constructed the adjacency matrix $\bm{A}$ of $\mathcal{G}(\R)$ via
$\bm{A}_{i,j} = \Sm_{i,j} \cdot (1-\bbone\{i=j\}).$
Unlike \citet{lu_understanding_2022}, they did not allocate nodes to clusters based on ground-truth labels, but determined an optimal soft assignment of $n$ clusters that maximizes modularity.
Specifically, they tried to find an optimal cluster assignment matrix $\bm{C}\in\Real^{D\times n}$, where each entry $\bm{C}_{j,k} \in [0,1]$ determines the assignment of neuron $j$ to cluster $k$.
The number of clusters $n\leq D$ of neuron activations is a parameter that is to be optimized as well.
Given a definition of clustering from \citet{girvan_community_2002}, neuron modularity is then defined as
\begin{equation}\label{eq:neurmod}
    m_{\operatorname{nMod}}(\R) = \max_{\bm{C}} \tr(\bm{C}^\transp \widetilde{\bm{A}} \bm{C}) - \tr (\bm{C}^\transp  \bm{1}_D^\transp\bm{1}_D\widetilde{\bm{A}} \bm{C}),
\end{equation}
where $\widetilde{\bm{A}} = \tfrac{1}{\bm{1}_D^\transp\bm{A}\bm{1}_D}\bm{A}$ is the normalized adjacency matrix.
To determine the cluster assignment $\bm{C}$, they provided an approximation method based on Newman's modularity maximization algorithm \cite{newman_modularity_2006}.
Generally, $m_{\operatorname{nMod}}$ is invariant to permutations, since these effectively only relabel the nodes in the resulting graph.

\begin{figure}[t]
    \centering
    \includegraphics[width=\textwidth]{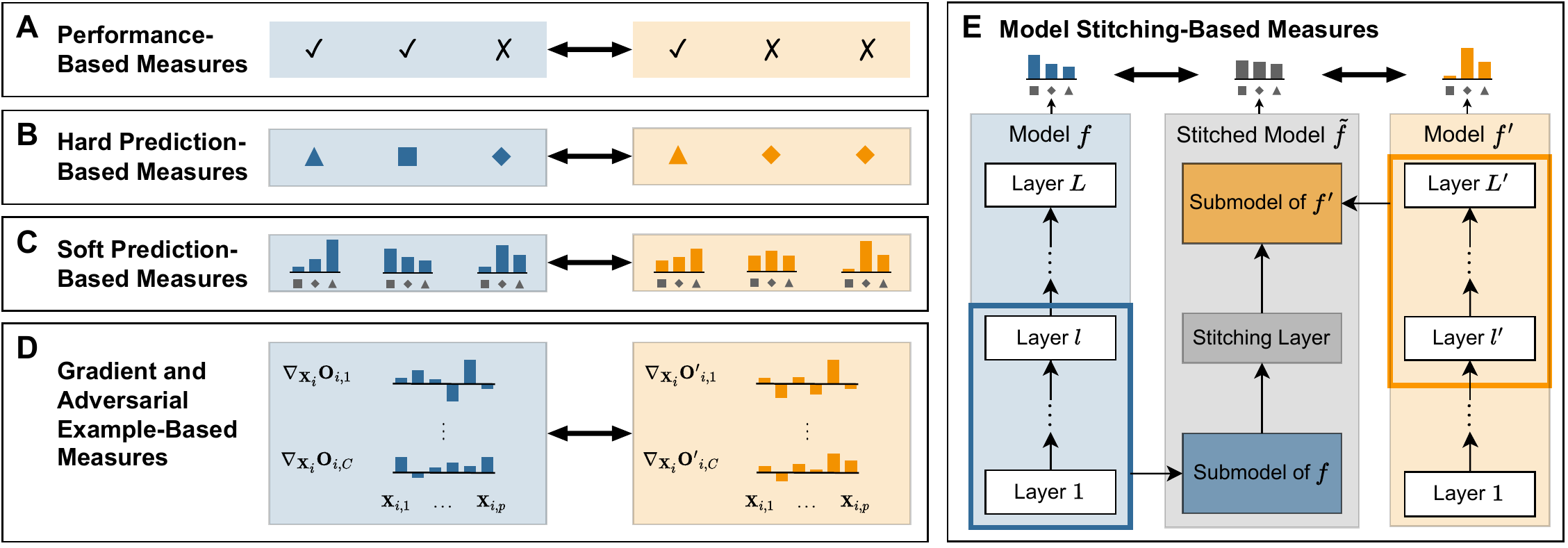}
    \caption{Types of functional similarity measures, illustrated in the context of classifying inputs with respect to their shape ($\diamond,\square, \triangle$). Performance-based (\textbf{A}), hard prediction-based (\textbf{B}), soft prediction-based (\textbf{C}), and gradient and adversarial example-based measures (\textbf{D}) compare
    outputs of different granularity. 
    Model stitching (\textbf{E}) combines parts of two models and measures functional similarity between the resulting model and the original models.
    }

    \label{fig:funcsim_illustr}
\end{figure}

\section{Functional Similarity Measures}
\label{sec:funcsim_methods}
Next, we present functional similarity measures.
As mentioned in \Cref{subsec:funcsim_intro}, these measures compare outputs $\Out, \Outp \in \Real^{N\times C}$, where each element $\Out_{i,c}$ denotes the probabilities or scores of class $c$ for input $\bm{X}_i$, and $\arg\max_c \Out_{i,c}=\hat{c}$ indicates that class $\hat{c}$ is the prediction for input $\bm{X}_i$.
We mainly categorize measures based on the granularity of the model outputs that they require, as illustrated in \Cref{fig:funcsim_illustr}.
An overview of all measures is given in \Cref{tab:funcsim}.

\subsection{Performance-Based Measures}

A popular view on functional similarity is that models are similar if they reach similar performance on some downstream task (e.g., \cite{ding_grounding_2021,lenc_understanding_2015, csiszarik_similarity_2021, bansal_revisiting_2021}).
This approach is easy to implement, as the comparison of models is reduced to comparing two scalar performance scores, such as accuracy.
However, this simplification also obfuscates more nuanced differences in functional behavior, which cannot directly be captured with a single number per model.

Most commonly, given some quality function $q$ that evaluates the performance of a model with respect to ground-truth labels, the (absolute) difference in performance is used for similarity:
\begin{equation}
    m_{\operatorname{Perf}}(\Out, \Outp)= |q(\Out) - q(\Outp)|.
\end{equation}
Although accuracy is an often used quality function in the literature \cite{ding_grounding_2021,lenc_understanding_2015, csiszarik_similarity_2021, bansal_revisiting_2021}, other performance metrics such as F1 score may be used \cite{yadav_survey_2019}.
However, choosing performance metrics that capture relevant aspects of functional behavior requires careful consideration \cite{reinke_understanding_2023}.

\begin{table*}[t]
    \centering
    \small
    \resizebox{0.8\linewidth}{!}{%
    \rowcolors{2}{white}{gray!25}
    \begin{tabular}{llcccc} \toprule
    Type                                                & Measure                                                                                                           & Groupwise & Blackbox Access   & Labels Required   & Similarity $\uparrow$  \\
    \midrule
    \cellcolor{white}Performance                        & Performance Difference                                                                                            &\xmark     & \cmark            & \cmark            &\xmark        \\
    \midrule
    \cellcolor{white}                                   & Disagreement \cite{madani_co-validation_2004,milani_fard_launch_2016,liu_model_2022,shamir_anti-distillation_2020}&\xmark     & \cmark            & \xmark            &\xmark        \\
    \cellcolor{white}                                   & Error-Corrected Disagreement \cite{fort_deep_2020}                                                                &\xmark     & \cmark            & \cmark            &\xmark        \\
    \cellcolor{white}                                   & Min-Max-normalized Disagreement \cite{klabunde_prediction_2023}                                                    &\xmark     & \cmark            & \cmark            &\xmark        \\
    \cellcolor{white}                                   & Kappa Statistic \cite{cohen_coefficient_1960}                                                                     &\xmark$^*$     & \cmark            & \xmark            & \cmark          \\
    \cellcolor{white}                                   & Ambiguity \cite{schumacher_effects_2021,marx_predictive_2020}                                                     &\cmark     & \cmark            & \xmark            &\xmark        \\
    \cellcolor{white}                                   & Discrepancy \cite{marx_predictive_2020}                                                                           &\cmark     & \cmark            & \xmark            &\xmark        \\
    \cellcolor{white}\multirow{-7}{*}{Hard Prediction}  & Label Entropy \cite{datta_measuring_2023}                                                                         &\cmark     & \cmark            & \xmark            &\xmark        \\
    \midrule
    \cellcolor{white}                                   & Norm of Soft Prediction Difference \cite{zhang_efficient_2021,ba_deep_2014}                                                          &\xmark     & \cmark            & \xmark            &\xmark        \\
    \cellcolor{white}                                   & Surrogate Churn \cite{bhojanapalli_reproducibility_2021}                                                          &\xmark     & \cmark            & \xmark            &\xmark        \\
    \cellcolor{white}                                   & Jensen-Shannon Divergence \cite{lin_divergence_1991}                                                              &\xmark     & \cmark            & \xmark            &\xmark        \\
    \cellcolor{white}                                   & Prediction Difference \cite{shamir_anti-distillation_2020}                                                        &\cmark     & \cmark            & \xmark            &\xmark        \\
    \cellcolor{white}\multirow{-5}{*}{Soft Prediction}  & Rashomon Capacity \cite{hsu_rashomon_2022}                                                                        &\cmark     & \cmark            & \xmark            &\xmark        \\
    \midrule
    \cellcolor{white}                                   & ModelDiff \cite{li_modeldiff_2021}                                                                                &\xmark     & \xmark$^\ddagger$ & \cmark            & \cmark          \\
    \cellcolor{white}                                   & Adversarial Transferability \cite{hwang_similarity_2023}                                                          &\xmark     & \xmark$^\ddagger$ & \cmark            & \cmark          \\
    \cellcolor{white}\multirow{-3}{*}{\shortstack{Gradient \& \\Adversarial Ex.}}         & Saliency Map Similarity \cite{jones_if_2022}                                                                      &\xmark     & \xmark            & \xmark            & \cmark          \\
    \midrule
    \cellcolor{white}Stitching                          & Performance Difference \cite{bansal_revisiting_2021,csiszarik_similarity_2021,lenc_understanding_2015}            &\xmark     & \xmark            & \cmark            &\xmark$^\dagger$        \\ \bottomrule
    \end{tabular}
    }
    {\par \tiny \raggedright $^{*}$: Groupwise variants available. $^\dagger$: Depends on comparison. $^\ddagger$: Depends on adversarial example generation.
		\par}
    \caption{
    Overview of functional similarity measures.
    We indicate whether measure enable groupwise comparison of models, whether they can be applied with blackbox access to the models, and if they require ground-truth labels.
    \emph{Similarity} $\uparrow$ indicates whether increasing scores imply increasing similarity of models. 
    }
    \label{tab:funcsim}
\end{table*}

\subsection{Hard Prediction-Based Measures}

The measures in this section quantify functional similarity by comparing hard predictions.
Thus, each measure of this category will report high similarity if the hard predictions agree for most inputs, regardless of correctness or confidence.
These measures are related to literature on ensemble diversity \cite{kuncheva_measures_2003, tang_analysis_2006} and inter-rater agreement \cite{banerjee_beyond_1999, tinsley_interrater_1975}.

\subheader{Disagreement.}\label{sec:disagr}
\emph{Disagreement}, also known as \emph{churn} \cite{milani_fard_launch_2016}, \emph{jitter} \cite{liu_model_2022}, or \emph{Hamming prediction differences} \cite{shamir_anti-distillation_2020}, is the expected rate of conflicting hard predictions over inputs and models \cite{skalak_sources_1996, madani_co-validation_2004}.
Due to its simplicity and interpretability, it is a particularly popular measure for functional similarity.
Formally, disagreement between two models is defined as
\begin{equation}\label{eq:disagreement}
    m_{\operatorname{Dis}}(\Out, \Outp) = \tfrac{1}{N} \textstyle\sum_{i=1}^{N} \bbone\{ \arg\max_j \Out_{i,j} \neq \arg\max_j \Outp_{i,j} \} .
\end{equation}
The measure is bounded in the interval $[0,1]$, with a score of zero indicating perfect agreement, and a score of one indicating completely distinct functional behavior.
Practically, this range is bounded by model quality, with high disagreement being impossible if the compared models are both very accurate.
Further, there are bounds on disagreement that depend on the soft predictions of the compared models \cite{bhojanapalli_reproducibility_2021}.

\subheader{Error-Corrected Disagreement.}\label{subsubsec:err_disagreement}
As the range of possible disagreement values depends on the accuracy of the compared models, \citet{fort_deep_2020} proposed to correct for this influence by dividing the disagreement by the error rate $q_{\operatorname{Err}}(\Out) := q_{\operatorname{Err}}(\Out,\bm{y}) := \tfrac{1}{N} \textstyle\sum_{i=1}^{N} \bbone\{ \arg\max_j \Out_{i,j} \neq \bm{y}_i  \}$ 
of one of the models:
\begin{equation}
    m_{\operatorname{ErrCorrDis}}(\Out, \Outp) = \tfrac{m_{\operatorname{Dis}}(\Out, \Outp)}{q_{\operatorname{Err}}(\Out)}.
\end{equation}
By design, this measure is not symmetric since the error rates of the outputs $\Out, \Outp$ may vary.
A normalized disagreement of zero indicates perfect agreement, whereas the upper limit is dependent on the error rate---exact limits are provided by \citet{fort_deep_2020}, which help to contextualize the similarity scores that are obtained.

A normalized and symmetric variant of this measure was used by  \citet{klabunde_prediction_2023}.
Their \emph{Min-Max-normalized disagreement} measure relates the observed disagreement $m_{\operatorname{Dis}}(\Out, \Outp)$ to the minimum and maximum possible disagreement, given error rates $q_{\operatorname{Err}}(\cdot)$ of the models.
The minimum is computed as $m_{\operatorname{Dis}}^{(\min)}(\Out, \Outp) = |q_{\operatorname{Err}}(\Out) -  q_{\operatorname{Err}}(\Outp)|$, and the maximum possible disagreement as $m_{\operatorname{Dis}}^{(\max)}(\Out, \Outp) = \min( q_{\operatorname{Err}}(\Out) + q_{\operatorname{Err}}(\Outp), 1)$, leading to the measure
\begin{equation}
    m_{\operatorname{MinMaxNormDis}}(\Out, \Outp) =
    \tfrac{m_{\operatorname{Dis}}(\Out, \Outp) - m_{\operatorname{Dis}}^{(\min)}(\Out, \Outp)}{m_{\operatorname{Dis}}^{(\max)}(\Out, \Outp) - m_{\operatorname{Dis}}^{(\min)}(\Out, \Outp)}.
\end{equation}
This measure is bounded in the interval $[0,1]$, with $m_{\operatorname{MinMaxNormDis}}(\Out, \Outp) = 0$ indicating perfect agreement. 

\subheader{Chance-Corrected Disagreement.}
Rather than correcting for accuracy of models, one can correct for the rate of agreement that two or more classification models are expected to have by chance.
The probably most prominent measure that follows this rationale is \emph{Cohen's kappa} \cite{cohen_coefficient_1960}, which was proposed as a measure for inter-rater agreement, but has also been used in machine learning \cite{celikyilmaz_evaluation_2021,geirhos_beyond_2020}.
Assuming that the compared outputs $\Out, \Outp$ are statistically independent, and letting $k_c$ denote the absolute amount of times that class $c$ is predicted in the output $\Out$, the expected agreement rate of such models is given by $p_e=\tfrac{1}{N^2}\textstyle\sum_{c=1}^C k_c k_c'$.
Based on these values, Cohen's Kappa is defined as
\begin{equation}
    m_{\text{Cohen}}(\Out, \Outp) = 1 - \tfrac{m_{\text{Dis}}(\Out, \Outp)}{1-p_e}=\tfrac{p_o-p_e}{1-p_e},
\end{equation}
where $p_o=1-m_{\text{Dis}}(\Out, \Outp)$ denotes the observed agreement.
When $m_{\text{Cohen}}(\Out, \Outp) = 1$, perfect agreement of the models is indicated; a value $m_{\text{Cohen}}(\Out, \Outp) < 0$ indicates less agreement than expected by chance.

To measure similarity between bigger sets of outputs, Fleiss's kappa \cite{fleiss_measuring_1971} can be used as a more general variant \cite{du_measuring_2023}.
The literature on inter-rater agreement \cite{feng_mistakes_2014} lists related measures that are more general or have weaker assumptions.

\subheader{Groupwise Disagreement.}
Disagreement cannot identify commonalities across a whole set of models, as pairwise similarity of models does not imply groupwise similarity.
The two following measures extend disagreement to identify functional similarity across sets of models.

First, \emph{ambiguity} \cite{marx_predictive_2020}, also called \emph{linear prediction overlap} \cite{gwilliam_beyond_2022}, is the share of instances that receive conflicting predictions by \emph{any} pair of models out of a given set of models.
Ambiguity is defined as
\begin{equation}
    m_{\text{Ambiguity}}(\mathcal{O}) =
    \tfrac{1}{N}\textstyle\sum_{i=1}^N \max_{\substack{\Out,\Outp \in \mathcal{O} \\ s.t.~\Out \neq \Outp}} \bbone\{ \arg\max_j \Out_{i,j} \neq \arg\max_j \Outp_{i,j} \}.
\end{equation}
The counterpart to ambiguity is the \emph{stable core} measure proposed by \citet{schumacher_effects_2021}, which counts the share of instances with consistent predictions.
They also considered a relaxation of this consistency, in which an instance is only required to obtain the same prediction by a fixed proportion of models (e.g., 90\% of all models) to be considered stable.

Second, \emph{discrepancy} \cite{marx_predictive_2020} gives 
the maximum disagreement between two classifiers from a set of multiple models:
\begin{equation}
    m_{\text{Discrepancy}}(\mathcal{O}) =
    \max_{\substack{\Out,\Outp \in \mathcal{O} \\ s.t.~\Out \neq \Outp}} \tfrac{1}{N}\textstyle\sum_{i=1}^N \bbone\{ \arg\max_j \Out_{i,j} \neq \arg\max_j \Outp_{i,j} \}.
\end{equation}
Both ambiguity and discrepancy are bounded in the interval $[0,1]$, with a value of zero indicating perfect agreement.

\subheader{Label Entropy.}
\citet{datta_measuring_2023} measured the variance in individual predictions over a group of outputs in terms of entropy.
Letting $k_c^{(i)}$ denote the number of times that instance $i$ is predicted as class $c$, \emph{Label Entropy} (LE) is defined as 
\begin{equation}\label{eq:label_entropy}
    m_{\text{LE}}(\mathcal{O}, i) = \textstyle\sum_{c=1}^C -\tfrac{k_c^{(i)}}{|\mathcal{O}|} \log\left(\tfrac{k_c^{(i)}}{|\mathcal{O}|}\right).
\end{equation}
Label Entropy is bounded in the interval $[0,\log(C)]$, with $m_{\text{LE}}(\mathcal{O}, i) = 0$ indicating identical predictions.

\subsection{Soft Prediction-Based Measures}

This group of measures compares soft predictions, such as class-wise probabilities or scores from decision functions.
Intuitively, this provides more nuance to the notion of similarity in outputs, since we can consider differences in confidence of individual predictions.
The impact of confidence is specifically exemplified by cases where scores are close to the decision boundary.
Even a minimal change in scores may cause a different classification in one case, whereas scores would need to change drastically for a different classification in another case.

\subheader{Norm of Soft Prediction Difference.}
A direct way to generalize disagreement to soft predictions is to apply a norm $\|\cdot\|$ on instance-wise differences in soft predictions, and average this over all inputs, which yields a measure
\begin{equation}\label{eq:pred_norm_diff}
    m_{\text{PredNormDiff}}(\Out, \Outp)=
    \tfrac{1}{2N}\textstyle\sum_{i=1}^N
        \left\|
            \Out_i - \Outp_i
        \right\|
\end{equation}
that assigns a score of zero when outputs are equal.
\citet{ba_deep_2014} and \citet{zhang_efficient_2021} applied this measure using the Euclidean norm to compare logits and probabilities, respectively.

\subheader{Surrogate Churn.}
\citet{bhojanapalli_reproducibility_2021} proposed \emph{surrogate churn} (SChurn) as a relaxed version of disagreement, that takes into account the distribution of the soft predictions.
For $\alpha > 0$, it is defined as
\begin{equation}
    m_{\text{SChurn}}^\alpha(\Out, \Outp)=
    \tfrac{1}{2N}\textstyle\sum_{i=1}^N
        \left\|
            \left(\tfrac{\Out_i}{\max_c \Out_{i,c}}\right)^\alpha - \left(\tfrac{\Outp_i}{\max_c \Outp_{i,c}}\right)^\alpha
        \right\|_1.
\end{equation}
A value $m_{\text{SChurn}}^\alpha(\Out, \Outp)=0$ indicates perfect agreement of outputs.
The authors showed that when $\alpha \rightarrow \infty$, this measure is equivalent to standard disagreement (cf. Sec. \ref{sec:disagr}), and use $\alpha=1$ as the default value.

\subheader{Divergence-Based Measures.}
When soft predictions represent class probabilities, divergence measures for probability distributions can be used to evaluate the similarity of instance-level predictions.
For example, Kullback-Leibler divergence is commonly used when training similar models in knowledge distillation \cite{gou2021knowledge}.
When focusing on pure similarity assessment, a common choice is to apply the symmetric \emph{Jensen-Shannon Divergence} (JSD) by averaging over all instances \cite{du_measuring_2023,fort_deep_2020, wu_similarity_2020}.
Letting $\operatorname{KL}(\cdot \| \cdot)$ denote the Kullback-Leibler divergence, this measure is defined as
\begin{equation}
    m_{\text{JSD}}(\Out, \Outp) =  \tfrac{1}{2N}\textstyle\sum_{i=1}^N
         \operatorname{KL}(\Out_i \| \ols{\Out}_i) +  \operatorname{KL}(\Outp_i \| \ols{\Out}_i),
\end{equation}
with $\ols{\Out} = \frac{\Out + \Outp}{2}$ denoting the average output.
Equality of outputs is given when  $m_{\text{JSD}}(\Out, \Outp) = 0$, and higher values indicate dissimilarity.
A similar approach to compare probabilistic outputs is given as \emph{Graph Explanation Faithfulness} (GEF) \cite{agarwal_evaluating_2023}.
An overview of divergence measures that could be used has been given by \citet{cha_comprehensive_2007}.

\subheader{Prediction Difference.}
\citet{shamir_anti-distillation_2020} specifically considered differences in predictions over more than two models.
Their \emph{prediction difference} (PD) intuitively quantifies the variance in model predictions.
Letting $\bm{\ols{\Out}}=\tfrac{1}{|\mathcal{O}|}\textstyle\sum_{\Out\in \mathcal{O}} \Out$ denote the average output matrix, their standard prediction difference measure aggregates instance-wise deviations from the average output in terms of a $p$-norm:
\begin{equation}
    m_{\text{PD}}^p(\mathcal{O}) = \tfrac{1}{N}\textstyle\sum_{i=1}^N \tfrac{1}{|\mathcal{O}|}\textstyle\sum_{\Out \in \mathcal{O}} \|\Out_{i} - \bm{\ols{\Out}}_i\|_p.
\end{equation}
\citet{shamir_anti-distillation_2020} used $p=1$ for interpretable differences of probability distributions.
$m_{\text{PD}}^p(\mathcal{O}) = 0$ indicates identical outputs of all models. Higher PD indicates higher dissimilarity between the compared models.

Next to norm-based prediction difference, \citet{shamir_anti-distillation_2020} further proposed a variant of the PD that relates the variance in the outputs to their average magnitude, and a variant that considers class labels $\bm{y}$ if these are given.

\subheader{Rashomon Capacity.}
Similar to label entropy \myeqref{eq:label_entropy}, \emph{Rashomon Capacity} (RC) \cite{hsu_rashomon_2022} also applies concepts from information theory to measure multiplicity in predictions on individual instances.
Formally, letting $P_{\bm{O}}$ denote a probability distribution over the set of outputs $\mathcal{O}$, and $\Delta_C= \big\{\bm{p}\in[0,1]^C : \textstyle\sum_{i=1}^C \bm{p}_i=1 \big\}$ the probability simplex, it considers the output spread
$\inf_{\bm{p}\in\Delta_C} \E_{\Out \sim P_{\bm{O}}} \operatorname{KL}(\Out_i\| \bm{p})$,
where $\bm{p}\in\Delta_C$ is a reference distribution that is optimized to minimize distances to all outputs.
The Rashomon Capacity is then defined via the \emph{channel capacity}, which maximizes the output spread over all probability distributions over the outputs:
\begin{equation}
    m_{\text{RC}}(\mathcal{O}, i) = 2^{\operatorname{Capacity}(\mathcal{O}, i)}, \quad\text{with}\quad \operatorname{Capacity}(\mathcal{O}, i) = \textstyle\sup_{P_{\bm{O}}}\textstyle\inf_{\bm{p}\in\Delta_C} \E_{\Out \sim P_{\bm{O}}} \operatorname{KL}(\Out_i\| \bm{p}).
\end{equation}
To approximate the Rashomon Capacity of an instance, \citet{hsu_rashomon_2022} suggested using the Blahut–Arimoto algorithm \cite{arimoto_algorithm_1972,blahut_computation_1972}.
A similarity measure over all instances can be obtained by aggregation, e.g., via the mean value.

It holds that $m_{\text{RC}}(\mathcal{O}, i)\in[1,C]$ with $m_{\text{RC}}(\mathcal{O}, i) = 1$ if and only if all outputs are identical, and $m_{\text{RC}}(\mathcal{O}, i) = C$ if and only if every class is predicted once with perfect confidence.
Further, the measure is monotonous, i.e., it holds that $m_{\text{RC}}(\mathcal{O}', i) \leq m_{\text{RC}}(\mathcal{O}, i)$ for all $\mathcal{O}'\subseteq \mathcal{O}$.

\subsection{Gradient and Adversarial Example-Based Measures}
The measures in this section use model gradients to characterize similarity either directly or indirectly via adversarial examples.
A core assumption of these measures is that similar models have similar gradients.
This assumption also leads to transferability of adversarial attacks, i.e., the behavior of the compared models changes similarly when given an adversarial example computed for only one of the models.

\subheader{ModelDiff.}
In their \emph{ModelDiff} measure, \citet{li_modeldiff_2021} used adversarial examples from perturbation attacks to characterize decision regions, which can then be compared across two models.
Given a model $f$, they first created adversarial examples $\widetilde{\bm{X}}_i$ for every input $\bm{X}_i$ by adding noise to these inputs that steer the model away from a correct prediction.
Such examples can be determined by methods such as projected gradient descent \cite{madry_towards_2019}.
The difference between the instance-wise original soft predictions $\Out_i = f(\bm{X}_i)$ and the predictions for the corresponding adversarial example $\widetilde{\Out}_i = f(\widetilde{\bm{X}}_i)$ is then collected in a \emph{decision distance vector} defined as $\big(\bm{v}_{\operatorname{DDV}}(\Out,\widetilde{\Out})\big)_i= \operatorname{cos-sim}(\Out_i, \widetilde{\Out}_i).$
Finally, they quantified the difference between models via the difference in the DDVs, measured with cosine similarity:
\begin{equation}
    m_{\text{ModelDiff}}(\Out, \Outp ) = \operatorname{cos-sim}(\bm{v}_{\operatorname{DDV}}(\Out,\widetilde{\Out}), \bm{v}_{\operatorname{DDV}}(\Outp,\widetilde{\Out}')).
\end{equation}
The outputs $\widetilde{\Out}'_i = f'(\widetilde{\bm{X}}_i)$ are computed from the same adversarial examples $\widetilde{\bm{X}}_i$.
A similarity score of one indicates equivalence of outputs.
Since this measure uses adversarial examples of only one of the models, it is not symmetric.

\subheader{Adversarial Transferability.}
Similar to ModelDiff, \citet{hwang_similarity_2023} measured the similarity of networks in terms of the transferability of adversarial attacks.
Given two networks $f,f'$, for each input $\bm{X}_i$ that is predicted correctly by both networks, a pair of corresponding adversarial examples $\widetilde{\bm{X}}_i, \widetilde{\bm{X}}_i'$ is generated with projected gradient descent \cite{madry_towards_2019}.
These adversarial examples are then fed into the opposite model, yielding outputs $\widetilde{\Out}_i = f(\widetilde{\bm{X}}'_i)$ and $\Tilde{\Out}'_i = f'(\widetilde{\bm{X}}_i)$, for which it is then determined how often both are incorrect.
Thus, given the vector of ground-truth labels $\bm{y}$, and letting $\mathcal{X}_{\operatorname{true}}$ denote the set of instances that were predicted correctly by both models, \citet{hwang_similarity_2023} defined the measure
\begin{equation}
    m_{\text{AdvTrans}}(\widetilde{\Out}, \widetilde{\Out}') = \log \big[
        \max \big\{
            \varepsilon,
            \tfrac{100}{2\left|\mathcal{X}_{\operatorname{true}}\right|} \textstyle\sum_{i \in \mathcal{X}_{\operatorname{true}}}
                \big(\bbone(\arg \max_j \widetilde{\Out}_{i,j} \neq \bm{y}_i)
                + \bbone(\arg \max_j \widetilde{\Out}'_{i,j} \neq \bm{y}_i)
        \big)
        \big\}
    \big],
\end{equation}
where $\varepsilon>0$ is introduced to avoid $\log(0)$.
A value of $m_{\text{AdvTrans}}(\widetilde{\Out}, \widetilde{\Out}') = \log(100)$ indicates perfect model similarity, whereas $m_{\text{AdvTrans}}(\widetilde{\Out}, \widetilde{\Out}') = \log(\varepsilon)$ indicates complete disagreement.

\subheader{Cosine Similarity of Saliency Maps.}
\citet{jones_if_2022} used a direct approach to compare models in terms of their gradients.
They computed the cosine similarity between (vectorized) saliency maps \cite{simonyan_deep_2013}, which model the impact of input features on individual predictions.
Practically, this impact is quantified using instance-wise gradients $\nabla_{\bm{X}_i}\Out_{i,c}$, and the instance-wise similarities then aggregated to yield the following measure:
\begin{equation}
    m_{\text{SaliencyMap}}(\Out, \Outp) =
    \tfrac{1}{nC}\textstyle\sum_{i=1}^N \textstyle\sum_{c=1}^C \operatorname{cos-sim} \big( 
    \big|\nabla_{\bm{X}_i}\Out_{i,c}\big|,
    \big|\nabla_{\bm{X}_i}\Outp_{i,c}\big|
    \big),
\end{equation}
where the absolute value $|\cdot|$ is applied element-wise (for inputs with a single channel).
A value $m_{\text{SaliencyMap}}(\Out, \Outp) = 1$ indicates perfect similarity, with lower values indicating stronger differences between models.

\subsection{Stitching-Based Measures} \label{subsec:stitching}

The intuition behind \emph{stitching} is that similar models should be similar in their internal processes and, thus, swapping layers between such models should not result in big differences in the outputs if a layer that converts representations is introduced \cite{lenc_understanding_2015, csiszarik_similarity_2021, bansal_revisiting_2021}.
Given two models $f, f'$, stitching consists of training a \emph{stitching layer} (or network) $g$ to convert representations from $f$ at layer $l$ into representations of $f'$ at layer $l'$.
One then considers the composed model
$\widetilde{f} := f'^{(L')} \circ \dots \circ f'^{(l'+1)} \circ f'^{(l')} \circ g \circ f^{(l)} \circ f^{(l-1)} \circ \dots \circ f^{(1)}$,
which uses the bottom-most layers of $f$ and the top-most layers of $f'$, and compares its output with the original models.
Most commonly they are compared in terms of a quality function $q$ such as accuracy \cite{csiszarik_similarity_2021, bansal_revisiting_2021}.
This yields a measure
\begin{equation}
    m_{\text{stitch}}(\widetilde{\Out}, \Outp) = q(\widetilde{\Out}) - q(\Outp),
\end{equation}
where $\widetilde{\Out}=\widetilde{f}(\bm{X})$ is the output of the stitched model.
However, other functional similarity measures can also be used.

Both design and placement of stitching layers affects assessments of model similarity, and several types of stitching layers were studied \cite{csiszarik_similarity_2021, godfrey_symmetries_2022}.
\citet{bansal_revisiting_2021} chose stitching layers such that the architecture of the stitched model is consistent with the original models.
For instance, they use a token-wise linear function to stitch transformer blocks.
For CNNs, $1\times 1$ convolutions are generally used in stitching layers \cite{lenc_understanding_2015, csiszarik_similarity_2021, bansal_revisiting_2021}.

Compared to other measures, model stitching requires training an additional layer and thus might be more costly to implement.
Further, (non-deterministic) training of the stitching layer presents a source of instability of the final results.
To train the stitching layers, one typically freezes parameters of the original models and only optimizes the weights of the stitching layer via gradient descent, using ground truth labels or the output of $f'$ as soft labels \cite{csiszarik_similarity_2021, bansal_revisiting_2021, lenc_understanding_2015}.
Additional tweaks such as normalization or regularization may be beneficial in certain contexts \cite{bansal_revisiting_2021, csiszarik_similarity_2021}.
For simple linear stitching layers $\bm{T}$, the weights can be computed by solving the least squares problem $\|\R^{(l)}\bm{T}-\bm{R'}^{(l^\prime)}\|_F$.

\section{Properties and Application of Similarity Measures}\label{sec:application}
In this section, we discuss practical aspects regarding the application of similarity measures.
We begin by outlining the current state of research that analyzes properties of existing measures and their relationship.
Afterwards, we summarize applications in existing literature, and then discuss the choice of measures in more detail, before providing additional considerations for comparing neural networks.

\begin{table}
\centering
\resizebox{\linewidth}{!}{%
{
\begin{tabular}{l|r|r|r|r|r|r|r|r||r|r|r|r|r|r|r|r|r|r|r|r|r|r|r}
\toprule
& \multicolumn{8}{c||}{Correlation (\ref{sec:corr_repfuncsim})} & \multicolumn{14}{c}{Discriminative Abilities (\ref{subsec:discr_repsim})}\\ \midrule
Test &  \multicolumn{4}{c|}{\begin{sideways}\hyperlink{test:ding}{Accuracy}\end{sideways} } &  \multicolumn{2}{c|}{\begin{sideways}\hyperlink{test:disagreement}{Disagreement}\end{sideways}}  & \begin{sideways} \hyperlink{test:jsd}{JSD} \end{sideways} & \begin{sideways} \hyperlink{test:squared_diff}{Squared Difference} \end{sideways} & \begin{sideways} \hyperlink{test:snr}{Noise Addition} \end{sideways} & \multicolumn{4}{c|}{\begin{sideways} \hyperlink{test:layer_match}{Layer Matching}\end{sideways}}  & \begin{sideways}\hyperlink{test:dim_subsample}{Dimension Subsample} \end{sideways} & \begin{sideways} \hyperlink{test:cluster_count}{Cluster Count} \end{sideways} & \begin{sideways} \hyperlink{test:arch_clustering}{Architecture Clustering} \end{sideways} & \begin{sideways} \hyperlink{test:contrasim}{Multilingual} \end{sideways} & \begin{sideways} \hyperlink{test:contrasim}{Image Caption} \end{sideways} & \begin{sideways} \hyperlink{test:resi_design}{Shortcut Affinity} \end{sideways} & \begin{sideways} \hyperlink{test:resi_design}{Augmentation} \end{sideways} & \begin{sideways} \hyperlink{test:resi_design}{Label Randomization} \end{sideways} & \begin{sideways} \hyperlink{test:resi_design}{Layer Monotonicity}\end{sideways} \\
Reference & \begin{sideways} \cite{boix-adsera_gulp_2022} \end{sideways} & \begin{sideways} \cite{ding_grounding_2021} \end{sideways} & \begin{sideways} \cite{hayne2024does} \end{sideways} & \begin{sideways} \cite{resi_benchmark_2024} \end{sideways} & \begin{sideways} \cite{barannikov_representation_2022} \end{sideways} & \begin{sideways} \cite{resi_benchmark_2024} \end{sideways} & \begin{sideways} \cite{resi_benchmark_2024} \end{sideways} & \begin{sideways} \cite{boix-adsera_gulp_2022} \end{sideways} & \begin{sideways} \cite{morcos_insights_2018} \end{sideways} & \begin{sideways} \cite{chen_graph-based_2021} \end{sideways} & \begin{sideways} \cite{kornblith_similarity_2019} \end{sideways} & \begin{sideways} \cite{rahamim_contrasim_2024} \end{sideways} & \begin{sideways} \cite{shahbazi_using_2021} \end{sideways} & \begin{sideways} \cite{shahbazi_using_2021} \end{sideways} & \begin{sideways} \cite{barannikov_representation_2022} \end{sideways} & \begin{sideways} \cite{boix-adsera_gulp_2022} \end{sideways} & \begin{sideways} \cite{rahamim_contrasim_2024} \end{sideways} & \begin{sideways} \cite{rahamim_contrasim_2024} \end{sideways} & \begin{sideways} \cite{resi_benchmark_2024} \end{sideways} & \begin{sideways} \cite{resi_benchmark_2024} \end{sideways} & \begin{sideways} \cite{resi_benchmark_2024} \end{sideways} & \begin{sideways} \cite{resi_benchmark_2024} \end{sideways} \\
\midrule
\rowcolor{Gray}Mean Canonical Correlation & 4 & & 4 & & & & & 5 & 3 & & 5 & & & & & & & & & & & \\
Mean Canonical Correlation$^2$ & & & 2 & & & & & & & & 3 & & & & & & & & & & & \\
\rowcolor{Gray}Singular Vector Canonical Correlation Analysis (SVCCA) & & & & 19 & & 16 & 14 & & 2 & & 6 & & & & 4 & & & & 21 & 23 & 10 & 5 \\
Projection-Weighted Canoncial Correlation Analysis (PWCCA) & 4 & 3 & 3 & 22 & & 14 & 17 & 4 & \textbf{1} & & 2 & 3 & & & & 4 & & & 17 & 20 & 20 & \textbf{1} \\
\rowcolor{Gray}Orthogonal Procrustes [CC] & & & & 5 & & 9 & 8 & & & & & & & & & & & & 8 & 8 & 9 & 5 \\
Orthogonal Procrustes [CC, MN] & 2 & \textbf{1} & \textbf{1} & 3 & & 5 & 3 & 2 & & & & & & & & 3 & & & 4 & 10 & 6 & 5 \\
\rowcolor{Gray}Permutation Procrustes & & & & 8 & & 11 & 15 & & & & & & & & & & & & 19 & 24 & 16 & 5 \\
Angular Shape Metric & & & & 3 & & 4 & 2 & & & & & & & & & & & & 3 & 10 & 7 & 5 \\
\rowcolor{Gray}Linear Regression & & & & 14 & & 12 & 11 & & & & 6 & & & & & & & & 12 & 9 & 22 & 5 \\
Aligned Cosine Similarity & & & & 12 & & 7 & 6 & & & & & & & & & & & & 6 & 6 & 11 & 5 \\
\rowcolor{Gray}Correlation Match [Relaxed] & & & & \textbf{1} & & 10 & 13 & & & & & & & & & & & & 13 & 18 & 19 & 5 \\
Correlation Match [Strict] & & & & 2 & & 14 & 16 & & & & & & & & & & & & 15 & 21 & 18 & 5 \\
\rowcolor{Gray}ContraSim & & & & & & & & & & & & \textbf{1} & & & & & \textbf{1} & \textbf{1} & & & & \\
Norm of Representational Similarity Matrix Difference & & & & 11 & & 21 & 21 & & & & & & & 3 & & & & & 18 & 13 & \textbf{1} & 5 \\
\rowcolor{Gray}Representational Similarity Analysis (RSA) & & & & 9 & & 2 & 10 & & & & & & & 4 & & & & & 10 & 12 & 17 & 5 \\
Centered Kernel Alignment (CKA) [Linear] & 3 & 2 & & 7 & 2 & 6 & 5 & 2 & & 2 & 4 & 2 & 2 & 2 & 3 & 2 & 2 & 2 & 9 & 7 & 4 & 5 \\
\rowcolor{Gray}Centered Kernel Alignment (CKA) [RBF 0.8] & & & & & & & & & & & \textbf{1} & & & & & & & & & & & \\
Distance Correlation (dCor) & & & & 6 & & \textbf{1} & 3 & & & & & & 3 & & & & & & 7 & 2 & 3 & 5 \\
\rowcolor{Gray}Eigenspace Overlap Score (EOS) & & & & 15 & & 20 & 18 & & & & & & & & & & & & 15 & 15 & 24 & 5 \\
Unified Linear Probing (GULP) [$\lambda=0$] & 6 & & & 23 & & 18 & 12 & 6 & & & & & & & & & & & 14 & 14 & 21 & 3 \\
\rowcolor{Gray}Unified Linear Probing (GULP) [tuned $\lambda$] & \textbf{1} & & & & & & & \textbf{1} & & & & & & & & \textbf{1} & & & & & & \\
Riemannian Distance & & & & & & & & & & & & & \textbf{1} & \textbf{1} & & & & & & & & \\
\rowcolor{Gray}Jaccard & & & & 21 & & 3 & \textbf{1} & & & & & & & & & & & & 2 & \textbf{1} & 15 & 5 \\
Second-Order Cosine Similarity & & & & 20 & & 8 & 9 & & & \textbf{1} & & & & & & & & & \textbf{1} & 3 & 5 & 5 \\
\rowcolor{Gray}Rank Similarity & & & & 10 & & 13 & 7 & & & & & & & & & & & & 5 & 4 & 12 & 5 \\
Multi-Scale Intrinsic Distance (IMD) & & & & 17 & & 19 & 20 & & & & & & & & 2 & & & & 20 & 16 & 8 & 3 \\
\rowcolor{Gray}Representation Topology Divergence (RTD) & & & & 13 & \textbf{1} & 17 & 19 & & & & & & & & \textbf{1} & & & & 11 & 5 & 2 & 2 \\
Magnitude Difference & & & & 17 & & 24 & 23 & & & & & & & & & & & & 24 & 19 & 13 & 5 \\
\rowcolor{Gray}Concentricity Difference & & & & 16 & & 23 & 22 & & & & & & & & & & & & 23 & 17 & 14 & 5 \\
Uniformity Difference & & & & 24 & & 22 & 23 & & & & & & & & & & & & 22 & 22 & 23 & 24 \\
\bottomrule
\end{tabular}
}
}
\caption{
\emph{Ranked performance of representational similarity measures in existing tests.} 
Rank 1 indicates best performance.
Empty cells indicate that a measure was not considered in the corresponding test.
For the Orthogonal Procrustes measure, different normalization strategies were used---either centering (CC), or both centering and normalization to unit norm (MN). 
Measures are ranked by their average performance across all variations of a single test; variance and quantitative differences in performance are not shown.
More details on the tests and rank aggregation are given in \Cref{ap:evaluation}.
Overall, it can be seen that most tests considered only a few measures, indicating a gap in existing research.
Further, no measure generally stands out.
}
\label{tab:evaluations_ranks}
\end{table}

\subheader{Properties and Evaluation of Similarity Measures.}
Most research on properties of similarity measures in the deep learning literature focuses on representational similarity.
We give a detailed review of existing analyses of representational similarity measures in \Cref{ap:analyses}.
We further provide an overview of existing comparative tests of these measures in \Cref{tab:evaluations_ranks}, which highlights that except for the recent \texttt{ReSi} benchmark \cite{resi_benchmark_2024}, most analyses only considered very limited sets of measures.
By contrast, functional similarity measures have been broadly analyzed in various contexts, including inter-rater agreement \cite{sim_kappa_2005,stemler_comparison_2019,maclure_misinterpretation_1987}, model fingerprinting \cite{sun_deep_2023}, and ensemble learning \cite{kuncheva_measures_2003}.

\subheader{Resources.}
There are only few resources that enable easy use of similarity measures.
Most notably, the recent \texttt{ReSi} benchmark\footnote{\url{https://github.com/mklabunde/resi}}
\cite{resi_benchmark_2024} provides implementations of 24 representational similarity measures, and also allows for testing of new measures on representations from a broad range of neural network models and datasets, spanning the graph, language, and vision domains.
Similarly, \citet{ding_grounding_2021} provide code\footnote{\url{https://github.com/js-d/sim_metric}}
to replicate their experiments and test new measures, albeit being smaller in scope.
Finally, \citet{cloos2024a} have collected implementations of similarity measures in an online repository\footnote{\url{https://github.com/nacloos/similarity-repository}},
aiming to provide a standardized interface for application of existing measures.

\subheader{Applications in the Literature.}
Similarity measures have been used in a wide array of contexts with two main objectives: to \emph{understand} aspects of deep learning, and to \emph{improve} deep learning systems.
In this section, we give an overview and examples of such applications; for a wider overview we refer to \citet[Section 4]{sucholutsky_getting_2023}.

Focal points of work aiming at \emph{understanding} deep learning include the effects of model architecture and objective function on what neural networks learn, as well as studies on model universality, i.e., the extent models converge to similar behavior under different training setups.
For example, \citet{raghu_vision_2021} and \citet{park_how_2022} studied the differences between vision transformers and CNNs by comparing their representations and analyzing changes in classification performance after modifying the architectures.
Similarly, the effect of width and depth \cite{nguyen_wide_2021} and the importance of specific layers \cite{sridhar_undivided_2020} has been investigated.
The impact of differences in the objective functions has, for instance, been studied by \citet{kornblith_why_2021}, who analyzed layer-wise differences in representations between models that vary only in their loss function, and further evaluated transferability of these models via functional similarity measures.
\citet{grigg_self-supervised_2021} took a similar approach when comparing supervised to self-supervised models.
Some studies have also investigated the impact of training adversarially robust models, by considering how intra- \cite{cianfarani_understanding_2022} and inter-architecture \cite{jones_if_2022} similarities of representations from robust and non-robust models differ, or how stitching these kinds of models affects performance \cite{balogh_functional_2023a}.
Finally, studies on model universality have found that neural networks trained under different training setups are often at least partially similar in their representations \cite{wu_similarity_2020,mcneely_exploring_2020,mehrer_individual_2020,summers_nondeterminism_2021,mehrer_beware_2018}, even if stemming from different modalities \cite{maniparambil_vision_2024}.
This is, however, contrasted by substantial differences in functional similarity that result from varying training seeds \cite{liu_model_2022,summers_nondeterminism_2021,klabunde_prediction_2023,bhojanapalli_reproducibility_2021,milani_fard_launch_2016} or the training data by a single instance \cite{black_leave-one-out_2021}.

Yet, there are other directions, including analyses on the impact of input features \cite{hermann_shapes_2020} and finetuning \cite{merchant_happens_2020}, or studies comparing representations of visual information from CNNs to those from mice \cite{shi_comparison_2019} and human brains \cite{xu_limits_2021}.

Work that used similarity measures to \emph{improve} systems is comparatively rarer, but includes studies on optimizing ensembles and knowledge distillation, i.e., the problem of transferring knowledge of a typically large teacher model into a smaller student model.
Works on improving ensembles have applied similarity measures when encouraging representational diversity of models \cite{wald_exploring_2023,do_simsmoe_2024,yang_trs_2021}, or penalizing similarity of soft predictions of ensemble parts \cite{zhang_efficient_2021}.
Similarly, in knowledge distillation, the student models have been trained by maximizing similarity with the teacher model in its representations \cite{saha_cka_distillation_2022,zhou_rethinking_2024, park_relational_2019} as well as in functional outputs \cite{zong_better_2023, gou2021knowledge}.

\subheader{Similarity Measure Selection.}
Using both representational and functional similarity measures allows for assessing similarity of neural networks in a holistic manner.
While deciding suitability of a measure requires a case-by-case evaluation, we can make some high-level recommendations.

For functional similarity measures, there are generally no measures that are fundamentally incorrect for a given application.
\Cref{tab:funcsim} provides all information necessary to narrow down the most suitable measures.
Using multiple measures can give nuanced insights.
Most prediction- and performance-based measures, except Rashomon capacity, can be computed in linear time, keeping computational costs low when using several measures from these categories.
For a robust analysis, we recommend using measures that control for confounding factors such as random agreement and error rate.
If white-box access to the models is available and computational constraints allow for it, one could further consider more granular gradient-based measures or stitching.

In contrast, selecting appropriate measures for representational similarity is challenging due to the opacity of neural representations.
It is often unclear which representations can be considered equivalent, and what kinds of differences in models or representations measures are sensitive to.
Despite the big number of existing measures, research evaluating their applicability with respect to these aspects is surprisingly limited.
Therefore, we can only give a few general recommendations.
First, if it is known which groups of transformations the given representations are equivalent under, measures should be filtered accordingly (see \Cref{tab:categorization}).
Second, one can check if some of the existing analyses referenced in \Cref{tab:evaluations_ranks} are relevant for the given scenario to narrow down the number of measures.
Third, some insights may come from the objective functions of the models to be compared.
For instance, if similarity of instance representations is modeled in terms of angles, similarity measures based on Euclidean distance may not be suitable.

Further, one can consider advantages and disadvantages of different categories of measures.
For instance, alignment-based measures are less flexible in their invariances than RSM- and neighborhood-based measures, which can easily be adapted in their inner similarity functions.
Topology-based measures, which also compute pairwise distance matrices in addition to estimate persistence intervals of topological features, face similar computational challenges, likely making them unsuitable for a large number of comparisons.
Hence, if the number of inputs is large, other categories of measures may be preferred---though for individual measures, faster variants such as a batched computation of CKA were proposed \cite{nguyen_origins_2022}.
Another key difference between representational similarity measures is their flexibility in weighting local versus global similarity of representations.
Neighborhood-based measures are easiest to adjust based on the number of nearest neighbors considered.
RSM-based measures can generally also address this issue by creating RSMs based on similarity functions that take distance between instances into account, e.g., the RBF kernel.
However, measures from other categories generally lack this flexibility.
Finally, the simple and interpretable nature of many descriptive statistics and neighborhood-based measures may be of interest in some applications.

\subheader{Additional Practical Considerations.}
Apart from measure selection, a few other aspects need to be considered when comparing neural networks.
First, as discussed in \Cref{sec:input_confounding}, the input data influences the similarity estimates.
When generalizability of results is desired, input data needs to be diverse \cite{sucholutsky_getting_2023}.
A larger number of inputs will, however, increase computational costs.
When focussing on representational similarity, the choice of layers for comparison yields a similar trade-off.
While pairwise comparisons of all layers provide the most detailed results, limiting comparisons to selected layers, such as the penultimate layer \cite{klabunde_towards_2023,jones_if_2022,nanda_measuring_2022}, may balance cost and detail effectively.

Finally, as discussed in \Cref{sec:reppreproc}, preprocessing of representations might be necessary to meet the requirements of some measures.
Understanding of the given representation space can, however, inform additional prepocessing.
For example,  language model representations like those from BERT \cite{devlin_bert_2019} often have a few dimensions with high mean and variance compared to the remaining dimensions \cite{timkey_rogue_2021}, skewing measures like cosine similarity.
\citet{timkey_rogue_2021} addressed this by standardizing the representations to zero mean and unit variance, thereby improving the alignment of cosine similarity between words with human similarity judgements.
Therefore, this normalization may be advisable when comparing such language representations via similarity measures that use cosine similarity, e.g., for RSMs or nearest neighbors.
This example also illustrates how normalization can extend the invariances of a given measure, which, in this case, would otherwise not be invariant to translation and anisotropic scaling.
However, normalizing representations without such insights can be counterproductive.
As can be seen in \Cref{tab:evaluations_ranks}, the performance of Orthogonal Procrustes is strongly affected by differences in normalization.

\section{Discussion and Open Research Challenges}
In this survey, we describe more than 50 similarity measures.
This yields a stark contrast to the rather small amount of research dedicated to systematically analyzing and comparing the existing measures that is highlighted in \Cref{ap:analyses}.
In particular, representational similarity measures pose many open questions of high practical relevance.
We argue that this constitutes a significant gap in research, as deeper understanding of the properties of measures is crucial to properly measure similarity and correctly interpret their scores.

In this section, we discuss challenges in the application of similarity measures, and connect these to open research questions that we argue require more attention in the future.
A discussion of notions of similarity and corresponding measures beyond the scope of this survey is provided in \Cref{sec:alt_similarity}.

\subheader{Applicability of Representational Similarity Measures.}
Applying different representational similarity measures to the same pair of models can yield materially different results \cite{klabunde_towards_2023}.
Given this potential for disagreement, it is crucial to have an understanding which measures are able to capture those differences in representations that are relevant for a given application scenario.
As discussed in \Cref{sec:application}, there is, however, only very limited research that has investigated the applicability of representational similarity measures in a broad manner.
Further, there is also only limited research aimed at understanding the geometry of neural representations, which could additionally inform about the compatibility of similarity measures with specific representations, or preprocessing approaches that could be utilized to create such compatibility.
The recent \texttt{ReSi} benchmark \cite{resi_benchmark_2024}, which builds on this survey, can be seen as a first effort toward enabling such systematic analyses, and we believe additional research in this direction is required to enable more informed decisions when choosing representational similarity measures.

\subheader{Interpretability.} 
Unless a similarity score indicates perfect (dis-)similarity through a bounded minimum or maximum value, one typically cannot directly infer an intuitively interpretable degree of similarity from the score itself.
One reason for this is that, due to non-linearities in a measure, the resulting scores may be misleading.
For example, the widely-used cosine similarity \myeqref{eq:cos-sim} changes non-linearly with the angle between two compared vectors, to the degree that a seemingly high similarity of 0.95 still corresponds to an 18° angle.
Another issue is that the similarity scores that one can obtain may strongly depend on the context.
For instance, a prediction disagreement \myeqref{eq:disagreement} of $0.05$ can be considered low in a difficult classification problem with many classes and high in an easy binary classification problem where one expects near-perfect accuracy.
Such contextualization may be easy to establish for measures as intuitive as disagreement, however, for more opaque measures, interpretation is typically much more difficult.
Generally, properties of the inputs can influence the obtained similarity scores (see \Cref{sec:input_confounding}), and further, factors such as dimensionality of representations might also affect the range that a similarity measure can produce.
The latter issue is exemplified in \Cref{ap:procdim}, where we show that the Orthogonal Procrustes scores of two random representations increase with increasing dimension, which proves how dissimilar representations may receive different scores based on such underlying factors.
Therefore, we argue that more research is required to improve the interpretability of measures, e.g., via expected values or boundaries of similarity scores in terms of input similarity or dimensionality.

\subheader{Robustness of Representational Similarity Measures.}
Specifically for the CKA \myeqref{eq:cka}, it has been shown that perturbations of single instance representations can strongly affect the resulting similarity scores (see \Cref{sec:corr_repfuncsim}).
Such sensitivities can be particularly harmful in applications where reliability of similarity measures is a prerequisite.
For instance, similarity measures could be used to identify model reuse in the legal context of intellectual property protection \cite{sun_deep_2023}.
Therefore, we argue that more research on the robustness of similarity measures is required to understand and improve their reliability.

\section{Conclusion} \label{sec:conclusion}
Representational similarity and functional similarity represent two complementing perspectives on analyzing and comparing neural networks.
In this work, we provide a comprehensive overview of existing measures for both representational and functional similarity.
We provide formal definitions for 53 similarity measures, along with a systematic categorization into different types of measures and pedagogical illustrations.

In addition, we survey the literature to shed light on some of their salient properties,
and provide guidance for the practical application of similarity measures.
We specifically identify a lack of research that analyzes properties and applicability of representational similarity measures for specific contexts in a unified manner.
This gap in the literature also affects the quality of the recommendations that one can make about their practical applicability.
We argue that additional research is necessary to enable the informed application of similarity measures and better understand similarity of neural network models.
Moreover, assessing similarity of neural networks is an important aspect in several deep learning-related problems, including knowledge distillation, pruning, model updating, continual learning, model merging, and contrastive learning.
Despite this importance, only limited consideration has been put into the choice of measures within such applications, which may also be due to a lack of awareness about the available measures and their properties.
In that sense, we hope that our work lays a foundation for more systematic research on the properties of similarity measures and their applicability across deep learning.
Further, with our categorization and analysis, we believe that our work can assist researchers and practitioners in choosing appropriate similarity measures.

\section*{Acknowledgements}
This work is supported by the Deutsche Forschungsgemeinschaft (DFG, German Research Foundation) under Grant No.: 453349072.


\printbibliography[filter=onlymain]

\newpage
\appendix
\begin{refsegment}
\section{Overview of Notations and Basic Definitions}
\label{ap:notations}
\renewcommand{\theequation}{\thesection.\arabic{equation}}
\setcounter{equation}{0}

\subsection{Notations}

Within the notations in this survey, we use a few conventions.
Sets are usually denoted with uppercase calligraphic letters, such as $\mathcal{N},\mathcal{O},\mathcal{P}$.
Matrices $\bm{M}\in\Real^{n_1\times n_2}$, $n_1,n_2\in\mathbb{N}$ are always denoted with bold uppercase letters, whereas vectors $\bm{v}\in\Real^n, n\in\mathbb{N}$ are denoted with bold lowercase letters.
General scalar variables $a\in\Real$ are usually denoted with regular lower-case letters, whereas specific constants, such as the number of classes $C$ in a classification task, or the dimension of representations $D$, are denoted with upper-case letters.
Specific lower-case variables are reserved, such as $m$ for model similarity measures, or $f$ for layer functions of neural networks.
All of these fixed variables are given in Table \ref{tab:notations}, all other variables are excluded there.

\begin{table}[t]
\begin{center}
\caption{Overview of Notations
}
\label{tab:notations}
\rowcolors{2}{white}{gray!25}
\begin{tabular}{|c|l|}
\bottomrule
     $f,f'$& Neural networks\\
     $f^{(l)},f'^{(l')}$& Layer $l$/$l'$ of neural networks $f, f'$\\
     $L, L'$& Total number of layers in $f,f'$\\
     $D, D'$& Number of neurons of a layer\\
     $N$& Number of inputs\\
     $C$& Number of classes in a classification task\\
     $m$& Similarity measure\\
     $q$& Quality function\\
     $\bm{y}$ & Vector of ground-truth labels \\
     $\bm{X}$& $N\times p$ matrix of $N$ inputs\\
     $\R, \Rp$& $N\times D$/$N\times D'$ representation matrices \\
     $\Out, \Outp$& $N\times C$ output matrices\\
     $\Sm, \Smp$& Representational Similarity Matrices (RSMs)\\
     $\mathcal{R}, \mathcal{O}$& Sets of representation/output matrices\\
     $\mathcal{T}$& Group of linear transformations\\
     $\sim_{\mathcal{T}}$& Equivalence up to transformations from $\mathcal{T}$\\
     $\operatorname{O}(D)$& Group of orthogonal transformations\\
     $\operatorname{GL}(D,\Real)$& Group of invertible matrices in $\Real^{D\times D}$\\
     $\mathcal{N}_{\R}^k(i)$& Set of $k$ nearest neighbors of $i$ in $\R$ \\
     $\transp$& Transpose of a matrix/vector\\
     $\bm{1}_n$& Vector of $n$ ones\\
     $\bm{I}_n$& Identity matrix of size $n\times n$\\
     $\bm{H}_n$& Centering matrix of size $n \times n$\\
     $\bbone$& Indicator function\\
\toprule
\end{tabular}
\end{center}
\end{table}

\subsection{Norms and Inner Products for Matrices}
Here, we briefly describe the Frobenius and nuclear norm that are used in some representational similarity measures.

\subheader{Frobenius Norm}. On the vector space of all matrices in $\Real^{n \times d}$, $n,d\in\Nat$, the \emph{Frobenius inner product} is defined as
\begin{equation}\label{eq:frobenius_prod}
    \langle \bm{A}, \bm{B} \rangle_F
    = \textstyle\sum_{i=1}^n \textstyle\sum_{j=1}^d \bm{A}_{i,j} \bm{B}_{i,j} = \operatorname{tr}(A^\transp B).
\end{equation}
This inner product induces the \emph{Frobenius norm}, which for $A\in \Real^{n \times d}$ is defined as
\begin{equation}\label{eq:frobenius_norm}
    \| \bm{A} \|_F = \sqrt{\langle \bm{A}, \bm{A} \rangle_F}
    = \sqrt{\textstyle\sum_{i=1}^n \textstyle\sum_{j=1}^d |\bm{A}_{i,j}|^2 } = \sqrt{\operatorname{tr}(A^\transp A)}
    = \sqrt{\textstyle\sum_{i=1}^{\min(n,d)} \sigma_i^2},
\end{equation}
with $\sigma_i$ denoting the $i$-th singular value of $A$. The Frobenius norm is invariant to orthogonal transformations.

\subheader{Nuclear Norm.}
Similar to the Frobenius norm, one can define the nuclear norm in terms of the singular values of a matrix
\begin{equation}\label{eq:nuclear_norm}
   \| \bm{A} \|_* = \textstyle\sum_{i=1}^{\min(n,d)} \sigma_i.
\end{equation}

\subsection{Similarity Functions for RSMs}\label{sec:simfct}
When analyzing representational similarity, instance-wise similarity functions $s: \Real^n\times \Real^n\longrightarrow \Real$, $n\in\Nat$ are often needed, in particular for RSM-based measures (here $n=D$).
As noted in \Cref{sec:rsm}, they further strongly impact which groups of transformations these measures are invariant to.
In the following, we provide a brief overview of common similarity functions, where we always assume two input vectors $v,v'\in\Real^n$ to be given.
We also provide an overview of the invariances that they induce on RSM-based measures in \Cref{tab:simfct}.
\begin{itemize}
    \item \textbf{Euclidean Distance.}
    This well-known distance function is defined as
    \begin{equation}
        \|v - v'\|_2 = \sqrt{\textstyle\sum_{i=1}^n (v_i - v_i')^2}
    \end{equation}
    This function satisfies the properties of a distance metric.
    \item \textbf{Cosine Similarity.}
    The cosine similarity between two vectors is defined as
    \begin{equation}\label{eq:cos-sim}
        \operatorname{cos-sim}(\bm{v}, \bm{v}') = \tfrac{ \bm{v}^\transp \bm{v}'}{\|\bm{v}\|_2 \|\bm{v}'\|_2}.
    \end{equation}
    It is bounded in the interval $[-1,1]$, with $\operatorname{cos-sim}(\bm{v}, \bm{v}') = 1$ indicating that both vectors point in the exact same direction, and $\operatorname{cos-sim}(\bm{v}, \bm{v}') = 0$ indicating orthogonality.
    \item \textbf{Linear Kernel.}
    This kernel is defined as
    \begin{equation}\label{eq:linkernel}
        K(\bm{v}, \bm{v}') = \bm{v}^\transp \bm{v}'.
    \end{equation}
    When $\bm{v}, \bm{v}$ have unit norm, this measure is equivalent to cosine similarity.
    $K(\bm{v}, \bm{v}') = 0$ indicates that vectors are orthogonal to each other.
    The linear kernel is not bounded.
    \item \textbf{Radial Basis Function Kernel.}
    The radial basis function (RBF) kernel is defined as
    \begin{equation}\label{eq:rbfkernel}
        K_\sigma(\bm{v}, \bm{v}') = \exp\Big(-\tfrac{\|v - v'\|^2_2}{2\sigma^2} \Big),
    \end{equation}
    where $\sigma\in\Real$ is a free parameter.
    With a range of $[0, 1]$, $K_\sigma(\bm{v}, \bm{v}') = 1$ indicates maximum similarity, a value of zero indicates minimal similarity.
    \item \textbf{Pearson Correlation.} Letting $\ols{\bm{v}}=\frac{1}{n}\sum_{i=1}^n \bm{v}_i$ denote the average value of the vector $\bm{v}$, the Pearson correlation coefficient is defined as
    \begin{equation}\label{eq:pearson}
        r(\bm{v}, \bm{v}') =
        \tfrac{\sum_{i=1}^n (\bm{v}_i - \ols{\bm{v}}_i)(\bm{v}'_i - \ols{\bm{v}}_i')}
        {\sqrt{\sum_{i=1}^n (\bm{v}_i - \ols{\bm{v}}_i)^2}\sqrt{\sum_{i=1}^n (\bm{v}'_i - \ols{\bm{v}}_i')^2}}.
    \end{equation}
    This function is bounded in the interval $[-1,1]$, with $r(\bm{v}, \bm{v}') = 1$ indicating perfect correlation, and $r(\bm{v}, \bm{v}') = 0$ no correlation at all.
    When $\bm{v},\bm{v}'$ are mean-centered, i.e. $\ols{\bm{v}}=\ols{\bm{v}}'=0$, this function is equivalent to cosine similarity.
\end{itemize}

\begin{table*}[t]
\centering

\rowcolors{2}{white}{gray!25}
\caption{Overview of Instance-wise similarity functions}\label{tab:simfct}
\begin{tabular}{lccccccc}
\bottomrule  &\multicolumn{6}{c}{\cellcolor{white}Induced Invariances} & \\
\cline{2-7}
Function & PT 			& OT &\cellcolor{white} IS 			& ILT 			& TR 			& AT & Metric \\
\bottomrule
Euclidean distance & \cmark 			& \cmark & \xmark & \xmark & \cmark 			& \xmark & \cmark \\
Cosine similarity & \cmark & \cmark & \cmark & \xmark & \xmark & \xmark & \xmark \\
Linear kernel & \cmark & \cmark & \xmark & \xmark & \xmark & \xmark & \xmark \\
RBF kernel & \cmark 			& \cmark & \xmark & \xmark & \cmark 			& \xmark & \xmark \\
Pearson correlation & \cmark & \cmark & \cmark & \xmark & \cmark & \xmark & \xmark \\\toprule
\end{tabular}
\end{table*}

\subsection{Intertwiner Groups}\label{ap:intertwiner}
\citet{godfrey_symmetries_2022} introduced the concept of \emph{intertwiner groups}, which they applied to analyze symmetries in neural network models.
Formally, given an invertible activation function $\sigma: \Real \longrightarrow \Real$, its corresponding intertwiner group is defined as:
\begin{equation}\label{eq:intertwiner_group}
    \operatorname{G}_{\sigma} := \operatorname{G}_{\sigma, D}= \{ \bm{A} \in \operatorname{GL}(D, \Real) : \exists \bm{B} \in \operatorname{GL}(D, \Real) \text{ s.t. } \sigma \circ \bm{A} = \bm{B} \circ \sigma\},
\end{equation}
where $\operatorname{GL}(D, \Real)$ denotes the general linear group of invertible matrices in $\Real^{D\times D}$.
The corresponding group of transformations of neural representations is then defined as
\begin{equation}\label{eq:intertwiner_transformers}
    \mathcal{T}_\sigma = \{\R\mapsto \R \bm{M} : \bm{M}\in\operatorname{G}_{\sigma}\}.
\end{equation}
We highlight the case where $\sigma = \operatorname{ReLU}$, which yields the group $\mathcal{T}_{\operatorname{ReLU}}$, because \citet{godfrey_symmetries_2022} detail similarity measures that are invariant to transformations from
 $\mathcal{T}_{\operatorname{ReLU}}$.
$G_{\operatorname{ReLU}}$ consists of matrices of the form $\bm{PD}$, where $\bm{P} \in \mathcal{P}$ is a permutation matrix and $\bm{D}$ is a diagonal matrix with positive elements.
Thus, invariance to $\mathcal{T}_{\operatorname{ReLU}}$ implies invariance to permutations, and when assuming representations with normalized columns, one can, for instance, constrain the Orthogonal Procrustes measure to be invariant to this group \cite{godfrey_symmetries_2022}.

\section{Preprocessing of Representations}\label{ap:preprocessing}

Next, we discuss techniques for normalization, adjusting dimensionality, and flattening of representations.

\subheader{Normalization.}
Some similarity measures assume that the representations are normalized.
For instance, it is commonly assumed that representations are mean-centered in the columns \cite{kornblith_similarity_2019,williams_generalized_2022,morcos_insights_2018}.
Mean-centering effectively constitutes a translation of the representations, which imposes the assumption that representations are equivalent under translations.
In consequence, the corresponding measures are invariant towards translations.
For such reasons, normalization methods should be used with caution, as they require the compared representations to be compatible with such assumptions.

In the following, we briefly discuss some commonly used normalization methods for representations or RSMs.
To keep the broader scope, we consider normalization of matrices $\bm{M}\in\Real^{n\times d}, n,d\in\Nat$.
Then, a normalization can be considered as a mapping $\psi:\Real^{n\times d} \longrightarrow \Real^{n\times d}$.
To simplify notation, in this context we apply the \emph{centering matrix} $\bm{H}_n$, $n\in\Nat$, which is defined as $\bm{H}_n = \bm{I}_n-\tfrac{1}{n}\bm{1_n}\bm{1_n}^\transp,$

\begin{itemize}
    \item \textbf{Rescaling of Instances to Unit Norm.} Letting $\bm{D} = \operatorname{diag}(\|\bm{M}_1\|_2,\dots,\|\bm{M}_n\|_2)$ denote the diagonal matrix of row lengths, this rescaling can be written as a transformation
    \begin{equation}\label{eq:unit_rescale}
     \bm{M} \mapsto \bm{D}^{-1}\bm{M}.
    \end{equation}
    This transformation preserves angles but alters Euclidean distances between vectors.

    \item \textbf{Rescaling of Columns to Unit Norm.} Letting $\bm{D} = \operatorname{diag}(\|\bm{M}_{-,1}\|_2,\dots,\|\bm{M}_{-,d}\|_2)$ denote the diagonal matrix of column lengths, this rescaling can be written as a transformation
    \begin{equation}\label{eq:unit_col_rescale}
     \bm{M} \mapsto \bm{M}\bm{D}^{-1}.
    \end{equation}
    This transformation preserves neither angles nor distances.

    \item \textbf{Rescaling of Matrix to Unit Norm.} This preprocessing rescales the whole matrix to unit norm:
    \begin{equation}\label{eq:matrix_norm_rescale}
        \bm{M} \mapsto \tfrac{\bm{M}}{\|\bm{M}\|_F}.
    \end{equation}
    Like the previous rescaling, angles are preserved, but Euclidean distances are not.

    \item \textbf{Mean-Centering of Columns}. This normalization sets the column means to zero, while preserving their variance.
    It can be written as a transformation
    \begin{equation}\label{eq:mean_center}
        \bm{M} \mapsto \bm{H}_n \bm{M},
    \end{equation}
    which effectively constitutes a translation of the representations.
    Thus, it alters angles but preserves Euclidean distance between representations. 

    \item \textbf{Double Mean-Centering.} This approach translates both rows and columns such that both row and column means equal zero. For any matrix $\bm{M}\in\Real^{n\times d}, n,d\in\Nat$, double mean-centering in rows and columns can be defined as a transformation
    \begin{equation}\label{eq:double_center}
     \bm{M} \mapsto \bm{H}_n \bm{M} \bm{H}_d
    \end{equation}
    This normalization is typically not applied directly to representations, as it would translate individual rows differently, and alter both Euclidean distance and angles between the row vectors.
\end{itemize}

\subheader{Adjusting Dimensionality.}
Many of the representational similarity measures presented in Section \ref{sec:repsim_methods} implicitly assume that the representations $\R, \Rp$ have the same dimensionality, i.e., $D = D'$.
Thus, if $D<D'$, some preprocessing technique must be applied to match the dimensionality.
Two techniques have been recommended for preprocessing: zero-padding and dimensionality reduction, such as principal component analysis (PCA) \cite{gower_generalized_1975, williams_generalized_2022}.
When zero-padding, the dimension $D$ of representation $\R$ is inflated by appending $D'-D$ columns of zeros to $\R$.
PCA conversely reduces the dimension of the representation $\R'$ by removing the $D'-D$ lowest-information components from the representation.

\subheader{Flattening.}
Representational similarity measures assume matrices $\R \in \Real^{N \times D}$ as input.
However, some models such as convolutional neural networks (CNNs) produce representations of more than two dimensions, making them incompatible with these measures.
In such a case, representations have to be flattened, taking into account model-specific properties of representations.
For example, representations from CNNs usually have the form $\R\in\Real^{N \times h\times w\times c}$, where $h,w$ denote height and width of the feature maps, and $c$ the number of channels.
Directly flattening these representations into matrices $\R\in\Real^{N\times hwc}$ would yield a format in which permuting the features would disregard the spatial information in the original feature map, which may be undesirable.
To avoid this issue, flattening CNN representations into matrices $\R\in\Real^{Nhw \times c}$ yields representations where permutations only affect the channels \cite{williams_generalized_2022}.
However, when comparing two models $f, f'$, their flattened representations are only compatible if the height and width of both models match or a feature map is upsampled, as the number of rows in the resulting matrices must match.
Further, computational cost of a similarity measure may be affected by the new effective numbers of features and inputs in the flattened representation.

\section{Analyses of Similarity Measures} \label{ap:analyses}
This section gives an overview of analyses of similarity measures that study the relation between representational and functional similarity, what kind of representations similarity measures can distinguish, and how the scores are influenced by the given inputs.
A summary of the comparative evaluations is shown in \Cref{tab:evaluations_ranks}.

\subsection{Correlation between Functional and Representational Measures}\label{sec:corr_repfuncsim}

\hypertarget{test:ding}{}
There has only been little work that investigates the relationship between representational and functional similarity.
Most prominently, \citet{ding_grounding_2021} studied on BERT \cite{devlin_bert_2019} and ResNet \cite{he_deep_2015} models whether diverging functional behavior correlates with diverging representational similarity.
To that end, they induced functional changes on the given models, such as varying training seeds, removing principal components of representations at certain layers, or applying out-of-distribution inputs, and investigated whether observed changes in accuracy on classification tasks correlate with changes in representational similarity as measured by CKA \myeqref{eq:cka}, PWCCA \myeqref{eq:pwcca}, and Orthogonal Procrustes \myeqref{eq:procrustes}.
They observed that Orthogonal Procrustes generally correlates with changes in functional behavior to a higher degree than CKA and PWCCA.
Further, CKA appeared much less sensitive to removal of principal components of representations than Orthogonal Procrustes and PWCCA---it still indicated high similarity between the original and the modified representation when the accuracy of the model has already dropped by over 15 percent.
GULP \myeqref{eq:gulp} was later benchmarked using the same protocol, and found to perform similarly to CKA and Procrustes, although it relied on good selection of regularization strength \cite{boix-adsera_gulp_2022}.

A similar analysis was conducted by \citet{hayne2024does}, who induced functional changes by deleting neurons in the linear layers of CNNs that were trained on ImageNet \cite{deng_imagenet_2009}.
\hypertarget{test:disagreement}{}
They reported that Orthogonal Procrustes and CKA correlate more with functional similarity than CCA measures.
\hypertarget{test:squared_diff}{}
\citet{barannikov_representation_2022} further compared disagreement~\myeqref{eq:disagreement} of models with CKA and RTD~\myeqref{eq:rtd} scores. CKA correlated to a lower degree than RTD.
\hypertarget{test:jsd}{}
\citet{boix-adsera_gulp_2022} correlated representational similarity with mean squared difference between outputs of regression models that were trained on the representations with random labels.
They found that GULP correlated better than CCA-based measures and CKA.
The recent \texttt{ReSi} benchmark \cite{resi_benchmark_2024} correlated over 20 representational similarity measures with accuracy, disagreement, and Jensen-Shannon Divergence. They used representations from vision, text, and graph models across multiple datasets. No measure consistently outperformed others across these tests.

\citet{davari_inadequacy_2022} pointed out how CKA is sensitive to manipulations of representations that would not affect the functional similarity of the underlying models.
For instance, they showed that one can alter the CKA of two identical representations to almost zero by translating the representation of a single instance in one of the copies, without affecting the separability of the representations with respect to their class.
Further, they could modify representations of multiple layers to obtain prespecified CKA scores between them, while leaving functional similarity almost unaffected.
Similar results were also reported by \citet{csiszarik_similarity_2021}.

\subsection{Discriminative Abilities of Representational Similarity Measures} \label{subsec:discr_repsim}
Invariances of representational similarity measures indicate which representations are considered equivalent.
However, measures have practical differences in distinguishing representations, which have been assessed in numerous works.

\hypertarget{test:snr}{}
\citet{morcos_insights_2018} tested the robustness of CCA-based measures (see \Cref{sec:cca}) to noise in representations.
They argued that measures should identify two representations as similar if
they share an identical subset of columns, next to a number of random noise dimensions.
In their experiments, they found that PWCCA is most robust in indicating high similarity, even if half of the dimensions are noise. By comparison, mean CCA was the least robust.

\hypertarget{test:layer_match}{}
A number of works \cite{kornblith_similarity_2019,chen_graph-based_2021,shahbazi_using_2021,rahamim_contrasim_2024} have explored the ability of representational similarity measures to match corresponding layers in pairs of models that only differ in their training seed: for instance, given two model instantiations and comparing layer five in one instantiation with all layers from the other model, the similarity with layer five from the other instantiation should be the highest.
No measure clearly outperformed other measures consistently.

\hypertarget{test:dim_subsample}{}
\citet{shahbazi_using_2021} tested whether representations obtained by sampling a low number of dimensions from a baseline representation yield high similarity with the baseline or other low-dimensional samples.
They compared CKA \myeqref{eq:cka}, Riemannian distance \myeqref{eq:riemann}, RSA \myeqref{eq:rsa}, and RSM norm difference \myeqref{eq:rsm_norm} on a neuroscience dataset, with sampled dimensions varying between 10 and 50.
For higher dimensions, all measures assigned high similarity between the samples and the baseline.
For low dimensions, only Riemannian distance consistently assigned high similarity between the sample and its original representation.
Other measures yielded lower similarities, yet CKA gave better results than RSA and the norm-based measure.

\hypertarget{test:cluster_count}{}
\citet{barannikov_representation_2022} used synthetic data patterns to test the ability of RTD \myeqref{eq:rtd} to discriminate between topologically different data.
They generated data consisting of increasing amounts of clusters, which were arranged circularly in two-dimensional space, and argued that the similarity between the dataset of one cluster and datasets with more clusters should decrease with increasing number of clusters.
The rank correlation between similarity score of a measure and number of clusters in the data was perfect for RTD, whereas CKA, SVCCA, and IMD \myeqref{eq:imd} had relatively low correlations.

\hypertarget{test:arch_clustering}{}
\citet{boix-adsera_gulp_2022} assumed that models of similar architecture have similar representations. Hence, they clustered ImageNet-trained models based on pairwise representational similarity and measured the quality of the resulting clusters.
CKA, Orthogonal Procrustes \myeqref{eq:procrustes}, and GULP \myeqref{eq:gulp} all allowed for good clustering in general; CCA-based measures tended to perform worse in comparison.
With optimized regularization strength $\lambda$, GULP overall yielded the best clustering among these measures.
Further, GULP clustered well even for inputs from other datasets.

\citet{tang_similarity_2020} argued that models trained from two similar datasets, such as CIFAR-10 and CIFAR-100 \cite{krizhevsky_learning_2009}, should be more similar compared to models trained on dissimilar datasets, that for instance do not contain natural images.
In their experiments, they compared CKA and NBS with respect to this desideratum, but results were inconclusive.

\hypertarget{test:contrasim}{}
\citet{rahamim_contrasim_2024} tested whether representations of text in different languages for a fixed model are more similar than representations of two random texts.
Similarly, they evaluated whether the representation of an image is most similar to the representation of the true caption compared to captions of other images.
In both cases, ContraSim \myeqref{eq:contrasim}, which was specifically trained for the respective task, outperformed CKA.

\hypertarget{test:resi_design}{}
Finally, the \texttt{ReSi} benchmark \cite{resi_benchmark_2024} proposed four tests that evaluate the discriminative abilities of measures.
In three of these tests final-layer representations of models with different behavior need to be distinguished.
These behavior differences stem from training with varying amount of random labels, shortcut features, or augmentation.
The fourth test correlated similarity in layer depth with representational similarity.
All tests are implemented over models from the vision, language, and graph domains across multiple datasets.
Initial results from the benchmark indicate that there is no similarity measure that performs well over all tests and domains.

\subsection{Influence of Inputs}\label{sec:input_confounding}

Another issue studied in literature is the impact of the inputs $\bm{X}$ on similarity scores.
For popular functional similarity measures, it is well-known that similarity of outputs is confounded by the accuracy of the models, the number of classes and the class distribution \cite{fort_deep_2020,klabunde_prediction_2023,bhojanapalli_reproducibility_2021,byrt_bias_1993,bakeman_detecting_1997}.
Similar confounding effects also exist with respect to representational similarity measures. In the following, we discuss corresponding results.

First, \citet{cui_deconfounded_2022} argued that similarity between input instances leads to similarity of their representations in early layers, as the extracted low-level features---even if they are different overall---cannot clearly distinguish between instances. Thus, RSMs mirror the pairwise similarities of the inputs, which leads to high similarity estimates between models that may actually be dissimilar.
This was demonstrated by showing that two random neural networks can obtain higher RSA \myeqref{eq:rsa} and CKA \myeqref{eq:cka} scores than a pair of networks trained for the same task.
To alleviate this problem, they proposed a regression-based approach to de-confound the RSMs.

Second, it was shown that representational similarity measures can be confounded by specific input features.
\citet{dujmovic_pitfalls_2022} compared a model trained on standard image data to models trained on modified images.
The modified images contained a class-leaking pixel to allow models to learn a shortcut for classification.
The locations of the leaking pixels affected representational similarity between the models, measured by RSA.
Similarly, \citet{jones_if_2022} found that feature co-occurrence in inputs may lead to overestimation of model similarity by CKA.
Different input features may co-occur in the data used to compute representations, but models may use these features to different extents.
For example, on a high level, the features ``hair" and ``eyes" co-occur in images of human faces, but one model may only use the hair to compute its representations, whereas the other model may only use the eyes feature.
They showed that CKA scores ignore the difference in feature use with an image inversion approach: using data synthetically generated to produce the same representations in one model, similarity to the other model dropped drastically as feature co-occurrences were eliminated.

Third, the number of input instances $N$ may influence similarity scores.
\citet{williams_generalized_2022} compared two CNNs trained on CIFAR-10.
The representations were computed from the test data with varying sample size $N$.
The similarity between the two CNNs in terms of the Angular Shape Metric \myeqref{eq:angshapemetric} generally decreased with increased ratio $N/D$, before a stable score was reached that did not change with more inputs.
When constraining the measure to permutation invariance, a lower ratio $N/D$ was sufficient to achieve a stable similarity score.

Finally, the effect of the choice of inputs was studied by \citet{brown_wild_2024}.
Using inputs from different data distributions significantly affected similarity scores of CKA and Orthogonal Procrustes.
However, similarity between models when giving in-distribution data was significantly correlated with similarity when given out-of-distribution data.
The extent of correlation heavily depends on the specific dataset.

\section{Details on Evaluation of Representational Similarity Measures}
\label{ap:evaluation}

In \Cref{tab:evaluations_ranks}, we show the rankings of representational similarity measures from different tests in literature, which we describe in \Cref{ap:analyses}.
Most of the tests, however, considered multiple variants where, for instance, datasets and models have been varied.
To obtain the single rank that is presented in \Cref{tab:evaluations_ranks}, we first created rankings for each test variant.
We then averaged these ranks for each measure and finally assigned the ranks as depicted in \Cref{tab:evaluations_ranks} based on these averages.
We note that these aggregated results do not highlight the considerable variance in performance that often occurred across test variants.
For example, $k$-NN Jaccard similarity is the best measure on average for the JSD correlation test in the \texttt{ReSi} benchmark, but across all model and data variants its rank varies between 1 and 18.
Further, these ranks do not indicate statistically significant differences.
The ranks should only be interpreted as a general direction of performance.
Thus, we generally recommend looking into the study that a test originated from to obtain more nuanced insights regarding the applicability of a measure for specific application scenarios.

In the following, we provide more detailed descriptions regarding how we determined and (if necessary) aggregated ranks for each of the listed tests.
To provide some further orientation, we give an overview of the models and datasets that were considered in each test in \Cref{tab:eval_tests}.
\begin{description}
    \item[Accuracy Correlation.] From the experiments conducted by \citet{boix-adsera_gulp_2022}, we aggregated the results depicted in Figures 22-24 in their appendix, where we computed individual rankings based on the correlation measured by Spearman's rho. As for the rank of GULP with optimized $\lambda$, we always chose the best result across all values of $\lambda$ in each individual test.
    For the analysis by \citet{ding_grounding_2021}, we aggregated ranks over all tests with respect to Spearman correlation, as they depicted in Table 1. Since PWCCA was not applicable in the vision tests, we ranked it last in the tests from these domains.
    From the experiments by \citet{hayne2024does}, we used the data published in their code repository.
    We first averaged the correlation values over all layers, then created separate rankings per model.
    Regarding the \texttt{ReSi} benchmark \cite{resi_benchmark_2024}, we considered and aggregated all results for test 1 (correlation to accuracy difference) as presented in Appendix B, where we ranked based on the reported Spearman correlation.
    \item[Disagreement Correlation.] From the experiments by \citet{barannikov_representation_2022}, we considered the results reported in Tables 1, 2 and 4, where RTD always outperformed CKA.
    Regarding \texttt{ReSi} \cite{resi_benchmark_2024}, we considered and aggregated all results for test 2 (correlation to output difference) as presented in Appendix B, where we ranked based on the reported Spearman correlation of representational similarity measures with disagreement.
    \item[JSD Correlation.] Again, we considered and aggregated all results for test 2 (correlation to output difference) of \texttt{ReSi} \cite{resi_benchmark_2024} as presented in Appendix B, where we ranked based on the reported Spearman correlation of representational similarity measures with Jensen-Shannon divergence.
    \item[Squared Error Correlation.] We allocated ranks based on the Spearman correlations reported in \cite[Figure 4]{boix-adsera_gulp_2022} which was averaged over the two given regularization strengths of the given linear predictors. Regarding the rank of GULP with optimized $\lambda$, we chose the best result across all values of $\lambda$ at each regularization strength.
    \item[Noise Addition.] To construct the ranks for the experiments by \citet{morcos_insights_2018}, we considered the areas under the curves as presented in Figure 2. They only considered one model and dataset, so we did not aggregate ranks.
    \item[Layer Matching.] \citet{chen_graph-based_2021} present the effect of hyperparameters on matching accuracy in Figure 3. We rank the measures based on the accuracy with respect to the optimal hyperparameters mentioned in the text (degree 5 and graph size 50) for the three used architectures.
    For \citet{kornblith_similarity_2019}, we rank measures based on the matching accuracies with respect to both CNNs (Table 2) and Transformers (Table F.1).
    From the experiments by \citet{rahamim_contrasim_2024}, we consider the results depicted in Table 1, where we
    rank each combination of encoder training set with representation dataset as a variant (separately for each domain).
    For \citet{shahbazi_using_2021}, we rank measures based on the mean matching accuracy as reported in their Figure 13.
    \item[Dimension Subsampling.] We considered and aggregated the results from \citet{shahbazi_using_2021} as depicted in Figures 5 and 6. For these individual experiments, ranks were, again, determined by the depicted areas under the curves. For RSA, we used the Spearman curve, RSM Norm Difference corresponds to the Euclidean curve.
    \item[Cluster Count.] We considered and aggregated results from both experiments with synthetic clusters and rings, where we ranked measures according to the Kendall's $\tau$ rank correlation with the number of clusters and rings, respectively, as is reported in \cite[Section 3.1]{barannikov_representation_2022}.
    \item[Architecture Clustering.] From the experiments by \citet{boix-adsera_gulp_2022}, we consider the results depicted in Figure 16, where we consider the pretrained and untrained models as different variants and rank measures by their average standard deviation ratio.
    \item[Multilingual.] From the multilingual benchmark by \citet{rahamim_contrasim_2024}, we considered the results depicted in Tables 2, 4, 5, where, in every test variant and for each probing layer, ContraSim had higher accuracy than CKA.
    \item[Image Caption.] From the image caption benchmark by \citet{rahamim_contrasim_2024}, we considered the results depicted in Figure 5 and Table 3 and 6, where, again, ContraSim had higher accuracy than CKA in all test variants.
    \item[Shortcut Affinity.] We considered and aggregated all results for test 4 (shortcut affinity) of \texttt{ReSi} \cite{resi_benchmark_2024} as presented in Appendix B, where we ranked all measures based on the reported AUPRC scores.
    \item[Augmentation.] We considered and aggregated all results for test 5 (augmentation) of the \texttt{ReSi} benchmark \cite{resi_benchmark_2024} as presented in Appendix B, where we ranked all measures based on the reported AUPRC scores.
    \item[Label Randomization.] We considered and aggregated all results for test 3 (label randomization) of the \texttt{ReSi} benchmark \cite{resi_benchmark_2024} as presented in Appendix B, where we ranked all measures based on the reported AUPRC scores.
    \item[Layer Monotonicity.] We considered and aggregated all results for test 6 (layer monotonicity) of the \texttt{ReSi} benchmark \cite{resi_benchmark_2024} as presented in Appendix B, where we ranked all measures based on the reported Spearman correlation.
\end{description}

\section{Neural Network Similarity Beyond This Survey}\label{sec:alt_similarity}
In this survey, we reviewed representational similarity measures that can compare representations from two different models that use the same inputs, and functional similarity measures that compare models in (multi-class) classification contexts.
Beyond the scope of this survey, there are other views on neural network similarity and application contexts, which we briefly discuss here.

\subheader{Functional Similarity for Non-Classification Tasks.}
Although we focus on functional similarity with respect to classification, many of the functional similarity measures can be used for or directly transferred to other downstream tasks.
In particular, if a suitable performance measure is given, performance-based measures can be used in any other context.
This is also the case for gradient-based and stitching measures if white-box access to the models is given, and, in case of gradient-based measures, adversarial examples can be constructed for the given context.
Soft and hard prediction-based measures, conversely, are limited to tasks where outputs are assigned discrete labels.
For regression, one could consider binning outputs to obtain discrete labels.
Further, there are specialized measures of agreement of continuous outputs \cite{tinsley_interrater_1975,gisev_interrater_2013}.
Finally, if output is structured, e.g., text or image generation, functional similarity becomes more difficult as outputs do not share universally identical semantics as in classification tasks.
For example, generated images may have differences that are not perceivable to the human eye, and thus could be considered equivalent.
This equivalence could lead to considering the invariances of functional similarity measures.
The evaluation of these kinds of models, including comparison of outputs to a human reference, was studied in prior surveys \cite{borji_pros_2018,celikyilmaz_evaluation_2021}.

\subheader{Alternative Notions of Neural Network Similarity.}
Aside from representational and functional similarity measures, there are several other notions of similarity that have been used to compare neural networks.
Some of these approaches are applicable for specific types of neural networks.
For instance, \emph{visualizations} have emerged as a popular tool to analyze CNN similarity, although not limited to them.
Approaches include the visualization of decision regions \cite{somepalli_can_2022}, neuron activations \cite{bau_network_2017}, or reconstructed images \cite{mahendran_understanding_2014}.
\citet{chen_inner-product_2023} proposed a method to compare \emph{weights} of convolutional layers.
For language models, \emph{probing} \cite{belinkov_probing_2021} has become a popular approach.
The idea behind probing is to compare the extent to which representations of models trained for a specific task such as sentiment analysis can also be used to predict related concepts such as part of speech.

There are also more universal approaches.
For instance, \citet{wang_understanding_2022} and \citet{guth_universality_2024} proposed methods to compare the \emph{weight matrices} of neural networks.
Further, one could also consider the impact of inputs, as done by \citet{shah_modeldiff_2022}.
They utilize the concept of \emph{datamodels} \cite{ilyas_datamodels_2022}, which aim to explain predictions in terms of which data samples were used in training.
Using that approach, they measure similarity in neural networks by comparing the influence that data points have on individual predictions.
\citet{salle_thinkalike_2024} considered the extent to which differences in meta-features, such as part of speech or tense for text models, predict differences in instance representations, and compared different models based on the importance of such features for the prediction.
Finally, measures that compare representations which are derived from different sets of inputs, but mapped into the same vector space, e.g., by coming from the same model, were proposed in the context of evaluating generative adversarial networks \cite{kynkaanniemi_improved_2019,barannikov_manifold_2021} and metric learning \cite{kulis_metriclearning_survey_2013,kaya_deepmetriclearning_survey_2019}.

\subheader{Representational Similarity for Training Neural Networks.}
Optimizing representations for high or low similarity during model training is a reoccurring theme across deep learning, e.g., in knowledge distillation \cite{gou2021knowledge} or fields that use contrastive representation learning \cite{manzoor2023multimodality, wang_SurveyContinualLearning_2024,uelwer2023survey,jaiswal_contrastive_survey_2020,gan_vision_survey_2022}.

The approaches to assessing similarity in this context are different from the similarity measures in this survey.
First, differentiability and computational efficiency become important properties to enable gradient descent-based optimization.
Second, and more importantly, in these processes it is often assumed that representations lie in the same representation space \cite{kim2021vilt,kaya_deepmetriclearning_survey_2019,kulis_metriclearning_survey_2013}, and similarity is often evaluated on the instance level rather than on full representation matrices \cite{reimers-gurevych-2019-sentence,wang2023all,kim_vilt_2021}.
Alternatively, if the representations come from different models such as in multi-modal representation learning, their mapping into the joint space can be trained together with the rest of the system \cite{wang2023all}.
Hence, unless invariances of the similarity measure should be used to make optimization more flexible, invariances are not important.
While these approaches are useful for training and aggregating representations at fixed layers of models, they do not generalize to post-training analysis of neural networks.

\subheader{Distinction to \emph{Functional Representations}.}
This survey covers representational and functional similarity measures.
The terms \emph{representational} and \emph{functional} should not be confused with \emph{functional representations}.
In contrast to our work, which is about comparing neural networks, work on functional representations is about training neural networks to represent continuous functions that are only known via samples at discrete points \cite{Mehta_2021_ICCV,sitzmann_implicit_2020,klocek_hypernetwork_2019}.

\section{Orthogonal Procrustes and Dimensionality}
\label{ap:procdim}
\begin{figure}
    \centering
    \includegraphics[width=0.5\textwidth]{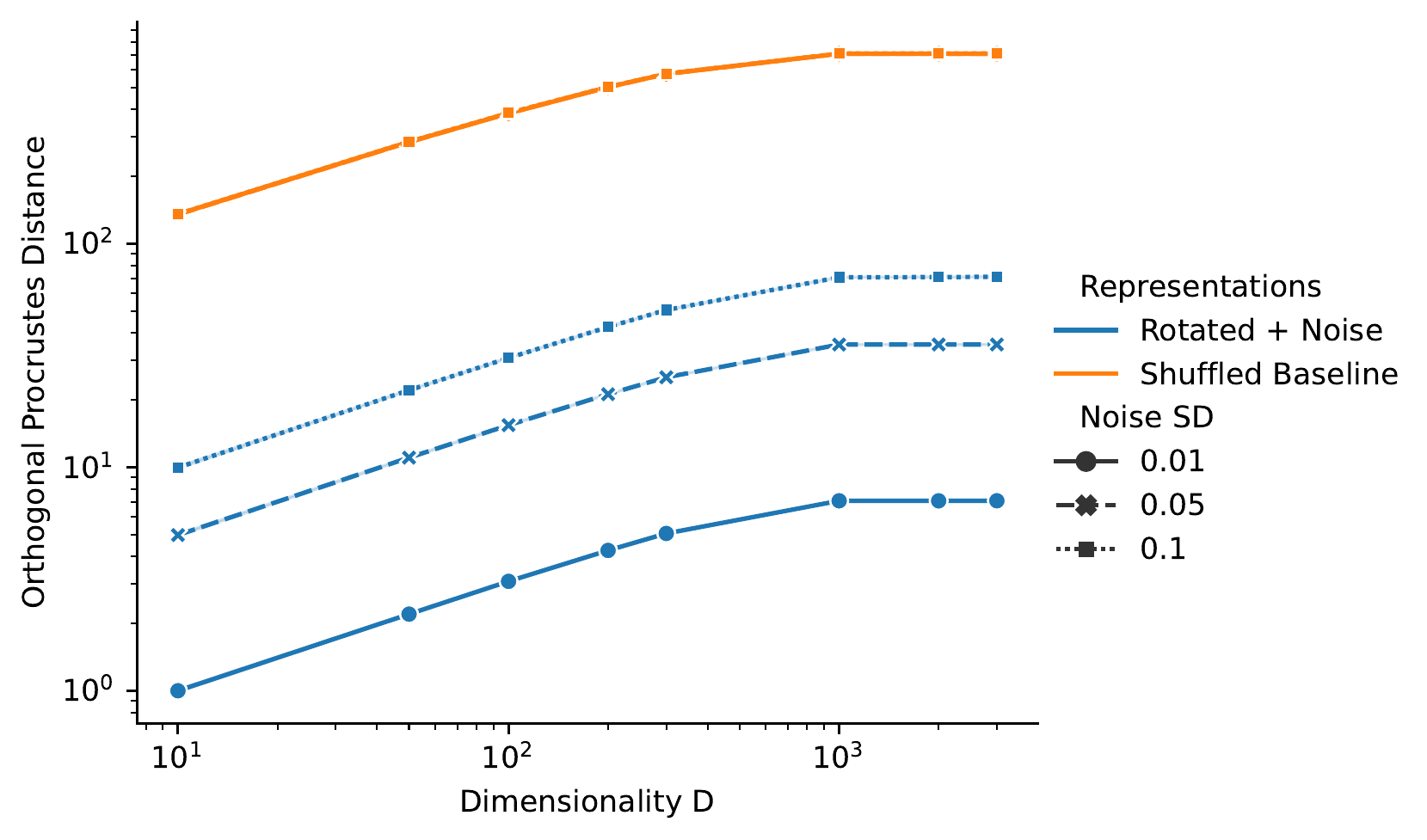}
    \caption{
        Mean Orthogonal Procrustes scores between two matrices over increasing dimensionality with varying noise level.
        The matrices have $N=1000$ rows.
        Shuffled Baseline refers to the score between two effectively unrelated matrices, a row-wise shuffled copy of the representation matrix and the original, similar to \citet{kriegeskorte_representational_2008}. The baseline is unrelated to the noise level.
        Scores increase until the number of dimensions matches the number of inputs ($N=D$), then stays flat.
        While $N>D$, the relation between the similarity score and the dimensionality follows a power law, as shown by the linear relation in the log-log plot.
        The standard deviation is too small to be visible.
        The same trend can be observed with other $N$ (not shown).
    }
    \label{fig:orthoproc}
\end{figure}

To demonstrate how the similarity scores of a measure may be influenced by external factors such as dimensionality, we plot values of the Orthogonal Procrustes measure over varying dimension in Figure \ref{fig:orthoproc}.

We compare two synthetic representation matrices:
the first matrix is a random matrix with entries drawn from a standard normal distribution, and the second matrix is generated by multiplying the first matrix with an orthogonal matrix that was randomly drawn from the Haar distribution as implemented by \texttt{scipy}\footnote{\url{https://docs.scipy.org/doc/scipy/reference/generated/scipy.stats.ortho_group.html}} \cite{mezzadri_how_2007}, with added noise, that is again drawn from a normal distribution.
These matrices have $N=1000$ rows and varying dimension $D \in \{10, 50, 100, 200, 300, 1000, 2000, 3000\}$.
This matrix generation process is repeated ten times for each value $D$, and we report the mean orthogonal Procrustes distance resulting from these matrix pairs.
In addition, we create a baseline similarity score by permuting the rows of a copy of the original representation matrix, and comparing it to the original representation matrix, similar to the technique proposed by \citet{kriegeskorte_representational_2008}.
We compute the baseline scores by shuffling the rows ten times for each representation pair, again reporting the mean.

The code to this experiment is available on GitHub\footnote{ \url{https://github.com/mklabunde/survey_measures}}.

\begin{landscape}
	\begin{table}[t]
	\caption{\emph{Models and Datasets that were considered in each test.} We separate models and datasets by domain (language, vision, or graphs).  Checkmarks indicate that a model or dataset has been used in the corresponding test, otherwise cells are empty. For the \emph{dimension subsample}, the \emph{signal to noise matching},  and \emph{cluster count} tests, no models have been used. Similarity measures were directly applied on the raw neuroimage data for the dimension subsample test, the other two used synthetic data.}
	   \label{tab:eval_tests}
	   \centering
				\resizebox{\linewidth}{!}{%
	   \begin{tabular}{lc||c|c|c|c|c|c||c|c|c|c|c|c|c|c|c|c|c||c|c|c|c||c|c|c|c|c|c|c||c|c|c|c|c|c||c|c|c||c|c|c}\bottomrule
																   &											& \multicolumn{21}{c||}{Models}																																														& \multicolumn{19}{c}{Datasets}																																																							\\
																   &											& \multicolumn{6}{c||}{Language}					& \multicolumn{11}{c||}{Vision}																										& \multicolumn{4}{c||}{Graphs}	& \multicolumn{7}{c||}{Language}											& \multicolumn{6}{c||}{Vision}									& \multicolumn{3}{c||}{Graphs}		& \multicolumn{3}{c}{Other}							 		\\\cmidrule{3-39}
		   Test													& Ref.										& \begin{sideways} Transformer \cite{vaswani_transformer_2017}	\end{sideways} & \begin{sideways} BERT \cite{devlin_bert_2019}	\end{sideways} & \begin{sideways} ALBERT \cite{Lan2020ALBERT}	\end{sideways} & \begin{sideways} SmolLM2 \cite{allal2024SmolLM2}	\end{sideways} & \begin{sideways} GPT-2	\cite{radford_gpt2_2019}\end{sideways} & \begin{sideways} XLM-R \cite{conneau_xlmr_2020} \end{sideways} & \begin{sideways} CNN/All-CNN-C \cite{springenberg_allcnnc_2014}	\end{sideways} & \begin{sideways} AlexNet \cite{krizhevsky_alexnet_2012}	\end{sideways} & \begin{sideways} Inception \cite{szegedy_inception_2015} \end{sideways} & \begin{sideways} VGG \cite{simonyan_vggnet_2014}	\end{sideways} & \begin{sideways} ResNet	\end{sideways} & \begin{sideways} MobileNet \cite{howard_mobilenets_2017}	\end{sideways} & \begin{sideways} MnasNet \cite{Tan_mnasnet_2019} \end{sideways} & \begin{sideways} RegNet \cite{Radosavovic_regnet_2020}	\end{sideways} & \begin{sideways} EfficientNet \cite{tan_efficientnet_2019}	\end{sideways} & \begin{sideways} ConvNeXt \cite{Liu_convnext_2022}	\end{sideways} & \begin{sideways} ViT \cite{dosovitskiy_vit_2021}	\end{sideways} & \begin{sideways} GCN \cite{kipf_gcn_2017} 	\end{sideways} & \begin{sideways} GraphSAGE \cite{hamilton_graphsage_2018}	\end{sideways} & \begin{sideways} GAT \cite{velickovic_gat_2018} 	\end{sideways} & \begin{sideways} P-GNN \cite{you_p-gnn_2019} 	\end{sideways} & \begin{sideways} MNLI \cite{williams_broad-coverage_2018}	\end{sideways} & \begin{sideways} SST-2 \cite{socher-etal-2013-recursive}	\end{sideways} & \begin{sideways} QNLI \cite{wang-etal-2018-glue} \end{sideways} & \begin{sideways} Penn TreeBank \cite{marcus_ptb_1993}	\end{sideways} & \begin{sideways} WikiText2 \cite{merity2017pointer}\end{sideways} & \begin{sideways} HANS \cite{mccoy-etal-2019-right} \end{sideways} & \begin{sideways} WMT 2014 \cite{bojar-etal-2014-findings}	\end{sideways} & \begin{sideways} ImageNet \cite{deng_imagenet_2009}	\end{sideways} & \begin{sideways} CIFAR-10 \cite{krizhevsky_learning_2009}	\end{sideways} & \begin{sideways} CIFAR-100 \cite{krizhevsky_learning_2009}\end{sideways} & \begin{sideways} SVHN \cite{netzer2011reading}	\end{sideways} & \begin{sideways} XNLI \cite{conneau-etal-2018-xnli}	\end{sideways} & \begin{sideways} UTKFace \cite{Zhang_2017_CVPR} 	\end{sideways} & \begin{sideways} Cora \cite{yang_revisiting_2016} 	\end{sideways} & \begin{sideways} Flickr \cite{zeng_graphsaint_2020}	\end{sideways} & \begin{sideways} OGBN-Arxiv \cite{hu_open_2021}	\end{sideways} & \begin{sideways} Conceptual Captions \cite{sharma-etal-2018-conceptual} 	\end{sideways} & \begin{sideways} Neuroimages \cite{visual_object_recognition_dataset}	\end{sideways} & \begin{sideways} Synthetic Data	\end{sideways}	\\\midrule \rowcolor{Gray}
		   \cellcolor{white}	 									& \cite{boix-adsera_gulp_2022} 				& 		&\cmark & &&	&	& 		&				& 		& 		& 		&  		&  		&  		& 		&  		& 		& 		& 		& 		& 		&\cmark	&\cmark &\cmark	&  		&   	&\cmark &   	&      	&   	& 		&     	&		&		& 		& 		& 		& 		& 		&		\\
																   & \cite{ding_grounding_2021}				& 		&\cmark &&&		& 		&				& 		& 		& 	 	&\cmark &  &		&  		& 		&  		& 		& 		& 		& 		& 		&\cmark	&\cmark &\cmark &  		&   	&   	&   	&       &\cmark &\cmark	&\cmark &		&		& 		& 		& 		& 		& 		&		\\ \rowcolor{Gray}
		   \cellcolor{white}   									& \cite{hayne2024does}				& 	&&	&  		&		& 		&				&\cmark	& 		& 	 	&\cmark	&\cmark &  &		& 		&  		& 		& 		& 		& 		& 		& 		&  		&	    &  		&   	&   	&   	&\cmark	&   	& 		&     	&		&		& 		& 		& 		& 		& 		&		\\
		   \multirow{-4}{*}{Accuracy Correlation} 					& \cite{resi_benchmark_2024}				& 		&\cmark	& \cmark& \cmark &		& 		&				& 		& 		&\cmark &\cmark &  		&  		& 		&  		& 		&\cmark	&\cmark&\cmark	&\cmark	&\cmark	&\cmark	&\cmark &   	&  		&   	&   	&   	&\cmark	&   	& \cmark		&     	&		&		&\cmark	&\cmark	&\cmark	& 		& 		&		\\\rowcolor{Gray}
		   \cellcolor{white}										& \cite{barannikov_representation_2022}		& 		&  		&&&		& 		&				& 		& 		&\cmark	&\cmark & & 		&  		& 		&  		& 		& 		& 		& 		& 		& 		&  		&   	&  		&   	&   	&   	&    	&\cmark &\cmark	&     	&		&		& 		& 		& 		& 		& 		&		\\
		   \multirow{-2}{*}{Disagreement Correlation} 				& \cite{resi_benchmark_2024}				& 		&\cmark&\cmark&\cmark	&		& 		&				& 		& 		&\cmark	&\cmark &  		&  		& 		&  		& 		&\cmark	&\cmark	&\cmark	&\cmark	&\cmark	&\cmark&\cmark &   	&  		&   	&   	&   	&\cmark	&   	& \cmark		&     	&		&		&\cmark	&\cmark	&\cmark	& 		& 		&		\\\rowcolor{Gray}
		   \cellcolor{white}JSD Correlation						& \cite{resi_benchmark_2024}				& 		&\cmark&\cmark&\cmark	&		& 		&				& 		& 		&\cmark	&\cmark &  		&  		& 		&  		& 		&\cmark	&\cmark	&\cmark	&\cmark	&\cmark	&\cmark&\cmark &   	&  		&   	&   	&   	&\cmark	&   	& \cmark		&     	&		&		&\cmark	&\cmark	&\cmark	& 		& 		&		\\
		   Squared Error Correlation								& \cite{boix-adsera_gulp_2022}				& 		&  		&	&&	& 		&				&\cmark	&\cmark	&\cmark &\cmark &\cmark &\cmark &\cmark	&\cmark	&\cmark	& &		& 		& 		& 		& 		&  		&   	&  		&   	&   	&   	&    	&   	& 		&     	&		&\cmark	& 		& 		& 		& 		& 		&		\\\midrule\rowcolor{Gray}
		   \cellcolor{white}Noise Addition							& \cite{morcos_insights_2018}  				& 		& 	&&	&&		& 		&				& 		& 		& 	 	&  		&  		&  		& 		&  		& 		& 		& 		& 		& 		& 		&  		&   	&		&		&   	&   	&    	& 		& 		&     	&		&		& 		& 		& 		&  		&  		&\cmark	\\
																   & \cite{chen_graph-based_2021}				&&& 		&  		&		& 		&\cmark			& 		& 		&\cmark	&\cmark &  		&  		& 		&  		& 		& 		& 		& 		& 		& 		&  		&   	&  		&   	&   	&   	&		&\cmark &		&     	&		&		& 		& 		& 		& 		& 		&		\\\rowcolor{Gray}
		   \cellcolor{white}										& \cite{kornblith_similarity_2019}			&\cmark	& 	&&	&		& 		&\cmark			& &		& 		& 	 	& 		&  		&  		& 		&  		& 		& 		& 		& 		& 		& 		&  		&   	&  		&   	&   	&\cmark &    	&\cmark &		&     	&		&		& 		& 		& 		& 		& 		&		\\
																   & \cite{rahamim_contrasim_2024} 			& 		&\cmark	&&&		& 		&				& 		& 		& 	 	&  		&  		&  		& 		&  		& 		&\cmark	& 	&	& 		& 		& 		&  		&   	&\cmark	&\cmark	&   	&   	&    	&\cmark &\cmark	&     	&		&		& 		& 		& 		& 		& 		&		\\\rowcolor{Gray}
		   \cellcolor{white}\multirow{-4}{*}{Layer Matching}		& \cite{shahbazi_using_2021}		&&		& 		& 		&		& 		&\cmark		&	& 		& 		& 	 	&  		&  		&  		& 		&  		& 		& 		& 		& 		& 		& 		&  		&   	&  		&   	&   	&   	&    	&\cmark & 		&     	&		&		& 		& 		& 		& 		& 		&		\\
		   Dimension Subsample						 				& \cite{shahbazi_using_2021}				&&& &		& 		&		& 		&				& 		& 		& 	 	&  		&  		&  		& 		&  		& 		& 		& 		& 		& 		& 		&  		&   	&  		&   	&   	&   	&    	&   	& 		&     	&		&		& 		& 		& 		& 		&\cmark	&		\\\rowcolor{Gray}
		   \cellcolor{white}Cluster Count							& \cite{barannikov_representation_2022}		&&& &		& 		&		& 		&				& 		& 		& 	 	&  		&  		&  		& 		&  		& 		& 		& 		& 		& 		& 		&  		&   	&  		&   	&   	&   	&    	&   	& 		&     	&		&		& 		& 		& 		& 		& 		&\cmark	\\
		   Architecture Clustering									& \cite{boix-adsera_gulp_2022}				&&& 		& 		&		& 		&				&\cmark	&\cmark	&\cmark &\cmark &\cmark	&\cmark &\cmark	&\cmark	&\cmark	& &		& 		& 		& 		& 		&  		&   	&  		&   	&   	& 		&\cmark	&   	& 		&     	&		&		& 		& 		& 		&      	&      	&		\\\rowcolor{Gray}
		   \cellcolor{white}Multilingual							& \cite{rahamim_contrasim_2024} 			& 		&\cmark	&	&&	&\cmark	&&				& 		& 		& 		&  		&  		&  		& 		&  		& 		& 		& 		& 		& 		& 		&  		&   	&  		&   	&   	& 		&   	&   	& 		&     	&\cmark	&		& 		& 		& 		& 		& 		&		\\
		   Image Caption											& \cite{rahamim_contrasim_2024} 			& 		&\cmark	&&&\cmark	& 		&				& 		& 		& 		&  		&  		&  		& 		&  		&\cmark	&\cmark	&& 		& 		& 		& 		&  		&   	&  		&   	&   	& 		&   	&   	& 		&     	&		&		& 		& 		& 		&\cmark	& 		&		\\\rowcolor{Gray}
			  \cellcolor{white}Shortcut Affinity						& \cite{resi_benchmark_2024}				& 		&\cmark&\cmark&\cmark	&		& 		&				& 		& 		&\cmark &\cmark &  		&  		& 		&  		& 		&\cmark	&\cmark	&\cmark	&\cmark	&\cmark	&\cmark&\cmark	&   	&  		&   	&   	& 		&\cmark &   	& \cmark		&     	&		&		&\cmark	&\cmark	&\cmark	& 		& 		&		\\
		   Augmentation											& \cite{resi_benchmark_2024}				& 		&\cmark	&\cmark&&		& 		&				& 		& 		&\cmark &\cmark &  		&  		& 		&  		& 		&\cmark	&\cmark	&\cmark	&\cmark	&\cmark	&\cmark&\cmark	&   	&  		&   	&   	& 		&\cmark &   	& \cmark		&     	&		&		&\cmark	&\cmark	&\cmark	& 		& 		&		\\\rowcolor{Gray}
		   \cellcolor{white}Label Randomization 					& \cite{resi_benchmark_2024}				& 		&\cmark&\cmark&\cmark	&		& 		&				& 		& 		&\cmark &\cmark &  		&  		& 		&  		& 		&\cmark	&\cmark	&\cmark	&\cmark	&\cmark	&\cmark	&\cmark&   	&  		&   	&   	& 		&\cmark &   	& 	\cmark	&     	&		&		&\cmark	&\cmark	&\cmark	& 		& 		&		\\
		   Layer Monotonicity										& \cite{resi_benchmark_2024}				& 		&\cmark&\cmark&\cmark	&		& 		&				& 		& 		&\cmark &\cmark &  		&  		& 		&  		& 		&\cmark	&\cmark	&\cmark	&\cmark	&\cmark	&\cmark&\cmark	&   	&  		&   	&   	& 		&\cmark &   	& \cmark		&     	&		&		&\cmark	&\cmark	&\cmark	& 		& 		&		\\\toprule
	   \end{tabular}
	   }

	\end{table}
   \end{landscape}

\section{Transformations for Figure 2}\label{ap:fig2_transforms}
In \Cref{fig:invariances}, the AT, ILT, and TR invariances use
\begin{align*}
  \mathbf{A}=\begin{bmatrix} 0.68 & 0.05\\0.22&0.18 \end{bmatrix} \quad  \mathbf{b} = \begin{bmatrix} 1.2 & -1.6 \end{bmatrix}.
\end{align*}
The illustrations of the OT, PT, and IS invariances use the following parameter values in their respective transformations:
\begin{align*}
  \mathbf{Q} = \begin{bmatrix}-0.87 & 0.5 \\ 0.5 & 0.87\end{bmatrix} \quad  \mathbf{P}=\begin{bmatrix} 0 & 1\\1&0 \end{bmatrix} \quad a = 2.
\end{align*}
The transformation $\bm{Q}$ corresponds to rotating the representation by 120 degrees and reflecting across the 15 degree axis.
The permutation $\bm{P}$ effectively swaps the axes in the coordinate system.

\printbibliography[filter=onlyapx]

\end{refsegment}

\end{document}